\documentclass[journal]{IEEEtran}
\ifCLASSINFOpdf
\else
\fi
\usepackage{graphicx}
\usepackage{amssymb}
\usepackage{amsmath, nccmath}
\usepackage{hyperref}
\usepackage[latin1,utf8]{inputenc}
\usepackage{multirow}% http://ctan.org/pkg/multirow
\usepackage{hhline}% http://ctan.org/pkg/hhline
\usepackage{multicol}
\usepackage{rotating}
\usepackage{float}
\usepackage{titlesec}

\DeclareMathOperator*{\argmin}{argmin} 
\usepackage{xcolor}
\usepackage{mathabx}
\usepackage[linesnumbered,ruled]{algorithm2e}
\usepackage{booktabs}
\usepackage[flushleft]{threeparttable}
\usepackage{accents}
\usepackage{enumitem}

\newcommand{\ubar}[1]{\underaccent{\bar}{#1}}
\newlength\lengthb 

\begin{document}

\newcommand{\ra}[1]{\renewcommand{\arraystretch}{#1}}
\newcommand{\qr}{\check{r}}
\newcommand{\hw}{\hat{w}}
\newcommand{\bvr}{\breve{r}}
\newcommand{\tbx}{\tilde{\mathbf{x}}}
\newcommand{\tbX}{\tilde{\mathbf{X}}}
\newcommand{\tbY}{\tilde{\mathbf{Y}}}
\newcommand{\hbth}{\boldsymbol{\hat{\theta}}}
\newcommand{\bth}{\boldsymbol{\theta}}
\newcommand{\hbt}{\hat{\boldsymbol{\tau}}}
\newcommand{\bt}{\boldsymbol{\tau}}
\newcommand{\hr}{\hat{r}}
\newcommand{\be}{\mathbf{e}}
\newcommand{\bbf}{\mathbf{f}}
\newcommand{\bA}{\mathbf{A}}
\newcommand{\bv}{\mathbf{v}}
\newcommand{\bY}{\mathbf{Y}}
\newcommand{\by}{\mathbf{y}}
\newcommand{\bx}{\mathbf{x}}
\newcommand{\bX}{\mathbf{X}}
\newcommand{\bb}{\boldsymbol{\beta}}
\newcommand{\hbb}{\hat{\boldsymbol{\beta}}}
\newcommand{\bR}{\mathbb{R}}
\newcommand{\bE}{\mathbb{E}}
\newcommand{\hs}{\hat{\sigma}}
\newcommand{\hst}{\hat{\sigma}_{\tau}}
\newcommand{\hsm}{\hat{\sigma}_{M}}
\newcommand{\st}{\sigma_{\tau}}
\newcommand{\br}{\mathbf{r}}
\newcommand{\bs}{\mathbf{s}}
\newcommand{\bOm}{\boldsymbol{\Omega}}
\newcommand{\bom}{\boldsymbol{\omega}}
\newcommand{\cby}{\check{\mathbf{y}}}
\newcommand{\cbY}{\check{\mathbf{Y}}}
\newcommand{\cy}{\check{y}}
\newcommand{\cbx}{\check{\mathbf{x}}}
\newcommand{\cbX}{\check{\mathbf{X}}}
\newcommand{\cbr}{\check{\mathbf{r}}}
\newcommand{\tr}{\tilde{r}}
\newcommand{\om}{\omega}

% no space, limits on side in displays
\def\lf{\left\lfloor}   
\def\rf{\right\rfloor}
\SetKwInput{KwInput}{Input}                % Set the Input
\SetKwInput{KwOutput}{Output}              % set the Output
\newcommand{\RN}[1]{%
  \textup{\uppercase\expandafter{\romannumeral#1}}%
}

\title{Two-Stage Robust and 
Sparse Distributed Statistical Inference for Large-Scale Data}

\author{Emadaldin~Mozafari-Majd,~\IEEEmembership{Student Member,~IEEE,}
        Visa~Koivunen,~\IEEEmembership{Fellow,~IEEE}
        % and~Jane~Doe,~\IEEEmembership{Life~Fellow,~IEEE}% <-this % stops a space
\thanks{The authors are with the Department
of Signal Processing and Acoustics, Aalto University, FI-00076 Aalto, Finland (email: emadaldin.mozafarimajd@aalto.fi; visa.koivunen@aalto.fi).}% <-this % stops a space
}

\maketitle

\begin{abstract}
In this paper, we address the problem of conducting statistical inference in settings involving large-scale data that may be high-dimensional and contaminated by outliers. The high volume and dimensionality of the data require distributed processing and storage solutions. We propose a two-stage distributed and robust statistical inference procedures coping with high-dimensional models by promoting sparsity. In the first stage, known as model selection, relevant predictors are locally selected by applying robust Lasso estimators to the distinct subsets of data. The variable selections from each computation node are then fused by a voting scheme to find the sparse basis for the complete data set. It identifies the relevant variables in a robust manner. In the second stage, the developed statistically robust and computationally efficient bootstrap methods are employed. The actual inference constructs confidence intervals, finds parameter estimates and quantifies standard deviation. Similar to stage 1, the results of local inference are communicated to the fusion center and combined there. By using analytical methods, we establish the favorable statistical properties of the robust and computationally efficient bootstrap methods including consistency for a fixed number of predictors, and robustness. The proposed two-stage robust and distributed inference procedures demonstrate reliable performance and robustness in variable selection, finding confidence intervals and bootstrap approximations of standard deviations even when data is high-dimensional and contaminated by outliers.
\end{abstract}


% \fi
\IEEEpeerreviewmaketitle

\section{Introduction}
\IEEEPARstart{M}{assive} quantities of ubiquitous and heterogeneous data  are generated by social media, smart phones, IoT, environmental monitoring, astronomical imaging devices and financial markets. Harnessing information from such large-scale data provides enterprises with meaningful insights into their performance and offers tremendous business opportunities. However, these benefits come with formidable challenges in handling storage, processing, acquisition and privacy concerns of high-speed and high volume data \cite{slavakis2014modeling}. In order to remedy the storage and processing issues, distributed computation and storage solutions are preferred. In a variety of statistical inference applications, one is dealing with high-dimensional data where the number of explaining variables $p$ may be comparable or much larger than the number of observations $n$. Often, high-dimensional problems exhibit a lower dimensional structure such as sparsity or low-rank. Regularization may be necessary to address the ill-posed high-dimensional problems and capture the parsimonious representation. In the context of linear regression, regularization by $\ell_1$-norm is known as the celebrated Lasso and performs simultaneous parameter estimation and model selection \cite{tibshirani1996regression}. It improves the prediction accuracy by introducing some bias while reducing the variance. There are certain scenarios where regularization by $\ell_1$-norm may cause significant bias in estimated coefficients. Moreover, the exact characterization of the limiting distribution for the Lasso estimator is a challenging task. In recent years, several valid statistical inference procedures have been introduced to characterize the construction of confidence intervals and hypothesis testing for high-dimensional problems \cite{chatterjee2011bootstrapping}, \cite{javanmard2014confidence}, \cite{liu2013asymptotic}. In general, the state-of-art procedures to address the uncertainty associated with parameter estimates belong to three main categories. First category concerns inference procedures based on bootstrapping. However, the conventional bootstrap methods fail to provide a reliable approximation to the distribution of Lasso estimator. In order to address this issue, Chatterjee and Lahiri \cite{chatterjee2011bootstrapping,chatterjee2013rates} proposed two alternative solutions,  modified residual bootstrap Lasso and  residual bootstrap adaptive Lasso. These solutions consistently estimate their limiting distributions and provide valid approximation of confidence intervals. The second category includes post-Lasso inference methods where the first stage involves model selection using Lasso and the actual inference is made in the second stage using the selected variables. This category includes sample splitting \cite{wasserman2009high}, bootstrap Lasso-OLS \cite{liu2013asymptotic}, bootstrap Lasso-partial Ridge \cite{liu2017bootstrap} and post-selection inference \cite{lee2016exact,berk2013valid} methods. Herein, we extend the concept of Post-Lasso estimator to conduct statistical inference for large-scale data sets using distributed computation and storage. The third category is based on the debiased-Lasso method \cite{javanmard2014confidence}, \cite{zhang2014confidence} where the key idea is to remove the bias  introduced by regularization from Lasso solution. These methods offer a concrete and general framework to quantify the uncertainty associated with parameters in high-dimensional settings, e.g. hypothesis testing and construction of confidence intervals. Other alternatives are covariance test \cite{lockhart2014significance}, knockoff filter \cite{barber2015controlling}, the ridge projection and bias correction \cite{buhlmann2013statistical}.

Another challenge in dealing with large-scale data is that the probability of observing outliers may increase as the dimensionality ($p$) and sample size ($n$) grows larger. Outliers may become masked and hence difficult to detect. Outliers may severely deteriorate the performance of ordinary least square and regularized least square estimators. Robust multivariate statistical procedures are required to cope with outlying observations and ensure the veracity of estimation, classification and decision making. In the context of linear regression, robust regularized estimators are  employed to ensure robustness and find sparse solutions simultaneously \cite{martinez2016regularized,smucler2017robust,freue2017pense, alfons2013sparse,zheng2017robust,zhang2014confidence}.

Modern statistical inference procedures need to accommodate distributed storage and parallel computations to deal with high volume and high-dimensional data. Bootstrap is a powerful tool for quantifying the uncertainty of estimates, i.e., confidence intervals and hypothesis tests. However, the applicability of conventional bootstrap techniques in large-scale settings may not be feasible because of computational constraints. The bootstrap resamples may have the same size as the original large-scale data, and repeating estimation for each bootstrap replicate becomes prohibitively expensive. The Bag of Little Bootstraps (BLB) \cite{kleiner2014scalable} offers a scalable and computationally efficient inference procedure for quantification of the uncertainty associated with estimates that accommodates distributed and parallel computing architectures. It subdivides the complete large-scale data into smaller distinct subsets. It can be considered as sampling without replacement from the complete data set. Then,  bootstrap is applied to each subset and combines the inference results from each subset. However, the BLB procedure is highly sensitive to outliers. In order to overcome this issue, a statistically robust BLFRB approach that extends the idea of fast fixed-point computations of FRB method in \cite{salibian2002bootrapping} to MM-estimators was introduced in \cite{basiri2015robust}. The modified bootstrap replicates are calculated by applying a linear correction factor to the one-step approximation of bootstrap replicates. Despite its robustness and computational efficiency, BLFRB performs unreliably in high-dimensional settings. 
%We will overcome the problem by using variable selection before fixed point estimation equations. As we explain later in the text, one may address this issue by initially conducting model selection and then using BLFRB on the selected model.

This paper focuses on statistical inference in large-scale settings with distributed architecture where data may be high-dimensional and contaminated by outliers. This problem has not been thoroughly studied in open literature. In particular, we propose two-stage statistical inference procedures that are robust to outlying observations and allow for using distributed storage and processing architectures for scalability. 
In the first stage, the relevant predictors are selected in two steps. First robust Lasso estimator is applied to distinct subsets of data in order to perform local variable selection. 
The variables for the whole data set are then found by applying a fusion rule to the selections from individual nodes at the fusion center or cloud.
In the second stage, we conduct inference by constructing confidence intervals, finding point estimates of the selected parameters and their standard deviations based on the large-scale data. The developed distributed and low-complexity inference procedures that use linearly corrected one-step robust estimators are employed. In special cases with very high dimensionality ($p/n \approx 1$ or $p \gg b$, $p$ number of predictors, $n$ sample size and $b$ subsample size), one may accelerate the model selection by a preprocessing stage excluding the majority of irrelevant variables via a robust variable screening procedure on the distinct subsets of data stored at each node. We address this issue in \textbf{Supplemental Materials}. The methods proposed in this paper extend our previous work on two-stage robust and  distributed inference procedures \cite{mozafari2019robust}, \cite{mozafari2021robust} by deriving computationally more efficient estimation methods and establishing statistical properties of the inference methods using analytical tools. We emphasize that our asymptotic analysis is restricted to the classical fixed $p$ setting. In \cite{liu2013asymptotic}, the asymptotic properties were established for a special case of post-Lasso estimator Lasso+mLS and the valid bootstrap approximation while allowing $p$ to grow at an exponential rate in $n$. However, their analysis does not cover the distributed and robust settings.

The main new contributions of the paper are summarized as follows:

%\begin{enumerate}[label={\Large\textbullet}]
\begin{itemize}
  \item \textbf{T}wo-\textbf{S}tage \textbf{R}obust and \textbf{D}istributed inference employing the class of $\tau$-estimators called TSRD-$\tau$ is introduced. A robust $\tau$-Lasso sparsity promoting estimator is employed in the first stage.
  
  \item The proposed TSRD-$\tau$ employs a novel \textbf{R}obust and \textbf{S}calable linearly corrected \textbf{O}ne-step \textbf{B}ootstrap procedure using $\boldsymbol{\tau}$-estimator (RSOB-$\tau$) for performing the actual inference. Its computational complexity is reduced by efficient estimation of bootstrap replicates. 
  
  \item  \textbf{T}wo-\textbf{S}tage \textbf{R}obust and \textbf{D}istributed inference employing the class of MM-estimators called TSRD-MM is developed. The sparsity promoting estimator in the first stage is robust MM-Lasso. It extends the BLFRB procedure \cite{basiri2015robust} by a robust variable selection stage promoting sparsity. 
  
%  \item In very high-dimensional scenarios, we apply a robust variable screening preprocessing step prior to variable selection for each distinct subset of data. It reduces the computational complexity by identifying irrelevant predictors and omitting them in subsequent processing. 
  %Application of robust variable screening prior to model selection in data subsets of very high-dimensionality reduces computational burden associated with high dimensionality of the data for the subsequent procedures by omitting large number of irrelevant predictors.
  
  \item Analytical results proving the robustness and consistency of the TSRD-$\tau$ method are derived for fixed $p$ and diverging $n$. In order to formally show the quantitative robustness of $\tau$-Lasso, its finite-sample breakdown is characterized.
  \item Extensive simulations are conducted to assess the performance of robust and distributed two-stage procedures in variable selection, bootstrap estimation of standard deviation, robustness of confidence intervals and computational complexity of the RSOB-$\tau$ procedure.
  \item Analytical results on the consistency and asymptotic normality of the RSOB-$\tau$ procedure are verified through computer simulations. Furthermore, the favorable theoretical findings on robustness of bootstrap replications are confirmed by extensive simulation studies and comparing them to \cite{basiri2015robust}.
 \end{itemize}
%\end{enumerate}

The two-stage inference solutions presented in this work achieve scalability by performing inference over smaller distinct subsets of data in parallel, using multinomial weighting and discarding irrelevant predictors. Considerable computational gains are achieved by using the low-complexity procedures to calculate bootstrap replications while retaining the consistency. The model selection stage facilitates preventing the impact of undesirable bias introduced by regularization and allows for the inference free of regularization parameter tuning.

This paper is organized as follows: In section 2, we describe the proposed two-stage inference methods and the employed data model. In section 3, the robust model selection methods and the distributed inference procedures are explained in more detail. In section 4, theoretical characterization of finite-sample breakdown point of robust $\tau$-Lasso is provided. Moreover, the details of RSOB-$\tau$ are explained and its consistency and robustness properties are established. Section 5 studies the performance of the two-stage robust inference procedures in simulations. Their consistency, robustness of confidence intervals, computational complexity, variable selection and standard deviation are considered. Section 6 concludes the paper. Detailed derivations and explanations of the theorems and their proofs can be found in the \textbf{Appendix} and \textbf{Supplemental Material}. Moreover, we present a robust variable screening procedure \cite{ghosh2020robust} prior to model selection, aiming to reduce the computational complexity in very high-dimensional data sets. We refer the interested readers to \textbf{Supplemental Materials}.

\section{Overview}
In this section, we will define the employed data model and briefly describe the proposed two-stage inference procedures.

\subsection{Data Model}
Consider a large-scale data with $n$ independent and identically distributed (i.i.d.) observations following a linear regression model
\begin{equation} \label{eq:linear_model}
    \by=\bX\bb_0+\bv
\end{equation}
where $\bX=(\bx_{[1]},\bx_{[2]},\hdots,\bx_{[n]})^T \in \bR^{n \times p}$ denotes a regression matrix, $\by=(y_1,y_2,\hdots,y_n)^T \in \bR^n$ is a response vector, $\bv=(v_1,v_2,\hdots,v_n)^T \in \bR^n$ is a measurement noise and the errors $v_l$ are assumed to be independent of predictors $\bx_{[l]}$. $\bb_0 \in \bR^p$ denotes a sparse parameter vector with $k_s=|\mathcal{S}|$ non-zero entries and $\mathcal{S}=\{j: \mathbf{1}([\beta_0]_{(j)} \neq 0)\}$. In order to deal with sheer volume of data, $\bY=\big( \by, \bX \big)$ is split into $s$ smaller distinct subsets of data $\cbY^{(i)}=\big( \cby^{(i)}, \cbX^{(i)} \big) \in \bR^{b \times (p+1)}, i=1,\cdots,s$ that can be stored and processed separately. The subsets may be formed by resampling without replacement from rows of the complete data set where $b=\{\lfloor n^{\gamma} \rfloor | \gamma \in [0.6, 1) \}$. The same situation would occur if subsets of data are stored on $s$ storage and computing nodes and each node contains $b$ observations.

\subsection{Assumptions}
 Suppose the measurement noise or errors $v_l$ follow some distribution $F_0$ and the distribution of the observed predictors $\bx_{[l]}$ is $G_0$. Then, the joint distribution $H_0$ of $(y_l,\bx_{[l]})$ is

\begin{equation}
\label{eq:dist_joint}
    H_0(\bx_{[l]},y_l)=G_0(\bx_{[l]})F_0(y_l-\bx_{[l]}^T \bb_0).
\end{equation}
We make the following assumptions on the distribution of errors and predictors.
\begin{itemize}
    \item The probability density $f_0(u)$ associated with probability distribution $F_0$ of the errors $v_l$ has the following properties: even, monotone decreasing in $|v|$ and strictly decreasing in a neighborhood of $0$.
    \item \label{B4} $\mathbb{P}(\bx_{[l]}^T \bb =\mathbf{0}) < 1-\delta$ for all non-zero $\bb$ and $\delta$ as defined by equation (\ref{eq:Mscale}).
     \item \label{B5} $G_0$ has a finite second moment and $\bE_{G_0}[\bx_{[l]} \bx_{[l]}^T]$ is non-singular.
\end{itemize}

The condition 1 generalizes the result established in this work to extremely heavy-tailed errors by imposing no moment conditions on the residual distribution $F_0$. The condition 2 guarantees the probability that observed values of explanatory variable are concentrated on a hyperplane does not get too large. The condition 3 concerns the second moment of explanatory variables and very common in the asymptotic analysis of regression estimators.
\subsection{Proposed Two-Stage Inference Methods}
In order to perform inference on potentially high-dimensional models in the presence of sparsity and outlying observations, we propose two-stage robust and distributed statistical inference methods where robust variable selection is performed in the first stage and then selected variables from each distinct subset of data are combined by using a fusion rule in the fusion center or cloud. The actual inference is done in the second stage by using robust and low-complexity bootstrapping procedures on the variables selected in the first stage.

The robust variable selection incorporates features of split-and-conquer approach \cite{chen2014split}, Bolasso \cite{bach2008bolasso} and robust lasso estimators. The Bolasso discusses that Lasso fails to produce consistent variable selection results under certain decays of regularization parameter. More specifically, the probability of selecting irrelevant variables is strictly positive. In order to overcome this issue, Bolasso recommends to generate sufficient number of bootstrap samples of the original data set and intersect the support of the parameter vector estimated based on each bootstrap. Thus, the irrelevant variables would randomly be selected by Lasso and could be eliminated from the support during intersection. The majority voting scheme is often preferred over intersection because if few of the relevant variables are erroneously not selected by one bootstrap replication, the AND-rule excludes those variables from the model. In the split-and-conquer approach, the data set is split into $s$ subsets and distinct subsets are processed separately. In large-scale settings with distributed storage and processing architecture, multiple distinct subsets of i.i.d. data are stored on nodes. One can estimate the support of the parameter vector based on distinct subsets of data stored at each node and combine them in the fusion center by using a voting scheme.  Due to page limitations, we chose to discuss only TSRD-$\tau$ and its related equations. The TSRD-MM method basically replaces the class of $\tau$-estimators in TSRD-$\tau$ with MM-estimators.\\

\subsection{TSRD-$\tau$}
We describe the core components of two-stage robust and distributed inference procedure employing class of $\tau$-estimators. In the first stage, the robust variable selection exploits the $\tau$-Lasso method \cite{martinez2015new}, \cite{martinez2016regularized} to select the sparse basis for each of the $s$ distinct data sets of $b$ observations. The selection results from each node are fused in a cloud or fusion center by using a percentage-based voting rule to choose the variables for the complete large data set. The chosen basis is communicated to each distributed computing and storage node and used in the second stage of the inference. A new robust and scalable technique using one-step bootstrap approximation of robust $\tau$-estimators is proposed to find parameter estimates for the selected basis and their confidence intervals in the presence of outliers. Bootstrap replicates are computed by using a robust low-dimensional $\tau$-estimator of regression \cite{yohai1988high}. The bootstrap percentile method is used to estimate confidence intervals associated with selected variables. The estimated confidence intervals from each node are communicated to the cloud or fusion center for the inference on complete large scale data. The confidence intervals estimated for each subset of data are combined at the fusion center by coordinate-wise averaging over the lower and upper bound of confidence intervals. 

%When dealing with large scale data with very high dimensionality ($p/n \approx 1$ or $p \gg b$), one may reduce the computational burden associated with variable selection by employing a preprocessing step. A robust sure independence screening procedure based on the concept of \textit{minimum density power divergence estimator} (MDPDE) \cite{ghosh2020robust} is employed to discard irrelevant variables from the subsets of data and speed up the computations in the model selection stage. 

\begin{figure*}
	\centering
	\includegraphics[width=13cm]{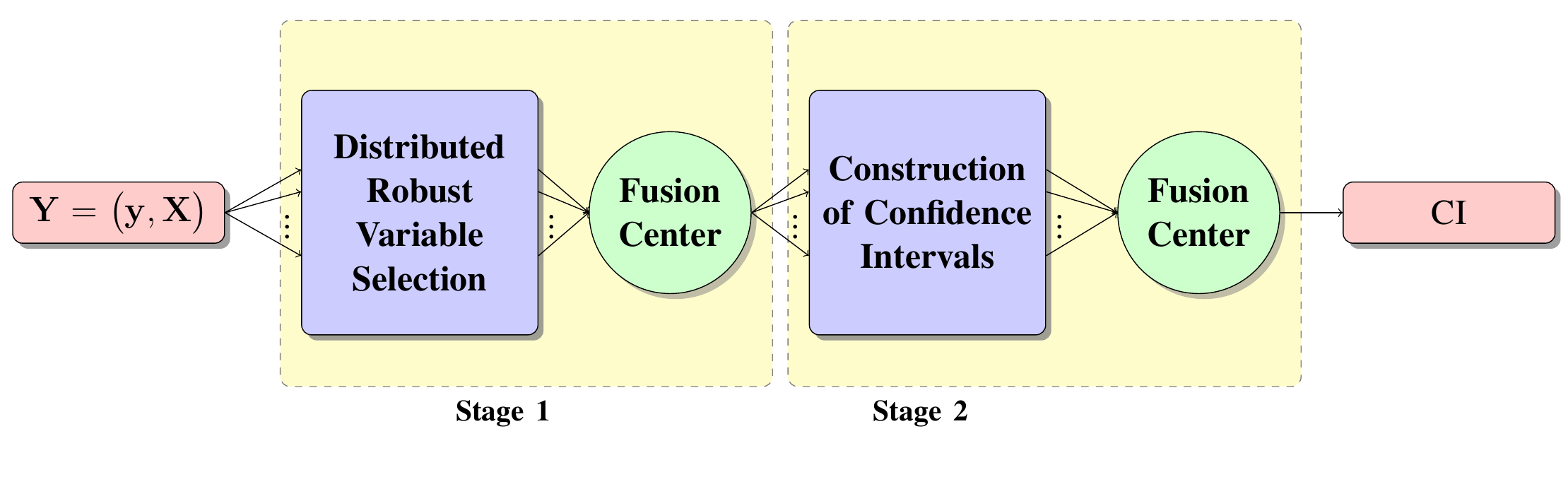}
	\vspace{-15pt}
	\caption{A simple schematic of the proposed two-stage robust and distributed inference.}
	   \label{fig:tws}
\end{figure*}

In addition, we show that the distribution of the proposed RSOB-$\tau$ asymptotically converges to the same limiting distribution as the sampling distribution of $\tau$-estimator for distinct subsets of data. Combining this with the assumption of $\mathbb{P}(\hat{\mathcal{S}}=\mathcal{S})=1$ as $n \rightarrow \infty$ implies that the asymptotic distribution of bootstrap  $\tau$-estimates for non-zero variables would converge to the same limiting normal distribution of $\tau$-estimator as if the true non-zero variables were known a priori. The robustness properties of $\tau$-Lasso and RSOB-$\tau$ are established by using analytical methods and computer simulations. A simple schemtaic of the proposed two-stage inference method is presented in \textbf{Fig. \ref{fig:tws}}. The details of the proposed TSDR-$\tau$ method are presented in the following sections.
\setlength\belowcaptionskip{-40pt}

\section{The Proposed Robust Inference Procedure}
\label{sec:TSRD}
\subsection{Stage \RN{1}: Model Selection}
\begin{figure*}
	\centering
	\includegraphics[width=14.1cm]{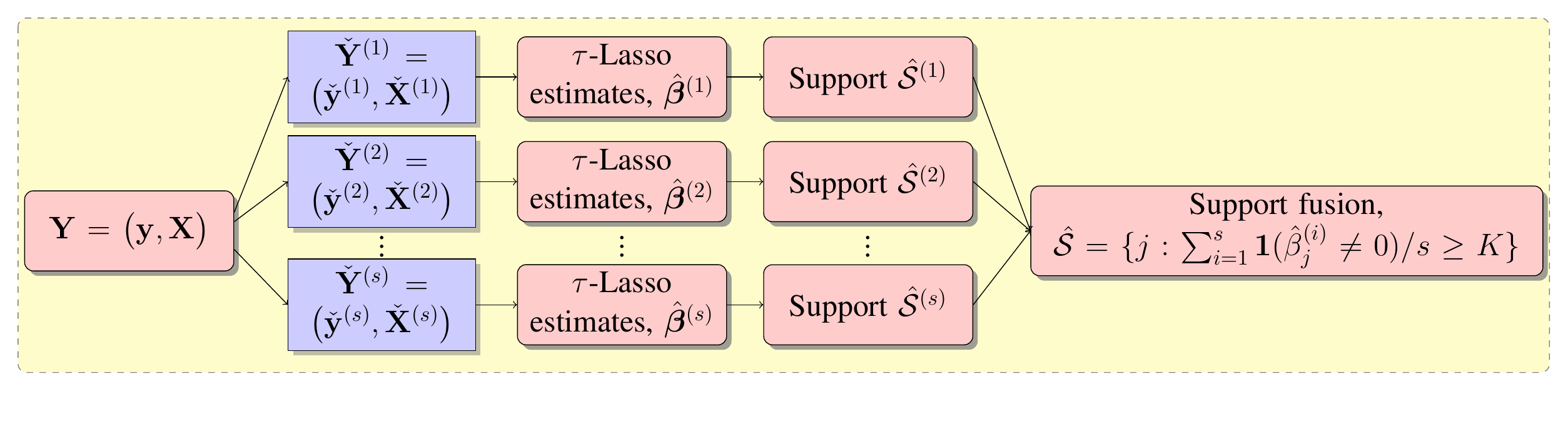}
	\vspace{-10pt}
	\caption{The model selection is carried out in two steps: performing model selection at each node storing a distinct subset of data and combining the selected models from each node in the fusion center through a voting scheme.}
	\label{fig:var_sel}
\end{figure*}

The proposed distributed and robust variable selection algorithm employed in the first stage of the inference procedure is described in detail. The main steps of the algorithm are summarized in \textbf{Fig. \ref{fig:var_sel}}.
\setlength\belowcaptionskip{-30pt}

The large-scale data is first split into $s$ distinct subsets via resampling without replacement. Then, the robust model selection is carried out in two steps in a distributed manner as follows:\\

\subsubsection{Variable Selection Using $\tau$-Lasso}
At each node, the relevant variables are selected by applying robust $\tau$-Lasso estimator \cite{martinez2015new}, \cite{martinez2016regularized} to each distinct subset of observations $\cbY^{(i)}=(\cby^{(i)},\cbX^{(i)}), i=1,...,s$. Hence, the robust variable selection is performed by solving a set of $s$ estimation sub-problems defined as follows:

\begin{equation}\label{eq:TAU-Lasso}
\hbb^{(i)} =\argmin_{\bb} \mathcal{L}(\bb)=\argmin_{\bb} \Big([\hs_{\tau}^{(i)}]^2+\lambda\|\bb\|_{\ell_1} \Big)
\end{equation}
where $\lambda$ controls the level of sparsity imposed by the ${\ell_1}$-norm penalty term. $\hs_{\tau}^{(i)}$ is a shorthand for $\hs_{\tau}(\cbr^{(i)}(\bb))$ an efficient estimate of scale defined as follows:  
 \begin{equation}
      [\hs_{\tau}^{(i)}]^2=\frac{[\hs_b^{(i)}]^2}{b}\sum_{l=1}^{b} \rho_1\Big(\frac{\check{r}_l^{(i)}(\bb)}{\hs_b^{(i)}}\Big)
 \end{equation}
where $\cbr^{(i)}(\bb)=\cby^{(i)}-\cbX^{(i)} \bb$ and $\hs_b^{(i)}$ denotes a shorthand for $\hs_{\text{M}}(\cbr^{(i)}(\bb))$ an M-scale estimate of residuals $\cbr^{(i)}(\bb)$ defined as the solution to

\begin{equation} \label{eq:Mscale}
   \frac{1}{b}\sum_{l=1}^{b} \rho_0 \Big(\frac{\check{r}_l^{(i)}(\bb)}{\hs_b^{(i)}}\Big)=\delta_1 
\end{equation}
where $\delta_1$ is a tuning constant controlling the asymptotic breakdown point of the estimator. $\rho_0(\cdot)$ and
$\rho_1(\cdot)$ are even and bounded functions satisfying the properties of $\rho$-function defined by Maronna et al. \cite{maronna2019robust}.  Tukey's bisquare $\rho$-function is considered as a popular choice in robust regression and defined as $\rho_i(t)=1-\big( 1-(t/c_i)^2\big)^3 \mathbf{1}(|t| \leq c_i), i=0,1$ where $c_0$ and $c_1$ are chosen so that the desired normal efficiency $\zeta^*$ and breakdown point $\delta^*$ are attained for $\lambda=0$, respectively. This can be achieved by finding $c_0$ and $c_1$ that satisfy $\bE[\rho_0(t)]=\delta^*$ and $\big(\bE[\psi^{'}(t)]\big)^2/\bE[\psi^{2}(t)]=\zeta^*$ under the normality assumption of errors $t \sim \mathcal{N}(0,1)$ when $\lambda=0$, simultaneously. $\psi(t)$, $\psi_0(t)$, $\psi_1(t)$ and $W$ are defined as follows:
\begin{equation}
    \begin{split}
        \psi(t)=W\psi_0(t)+\psi_1(t), \\
        \psi_0(t)=\partial \rho_0(t)/ \partial t, \quad
        \psi_1(t)=\partial  \rho_1(t)/ \partial t, \\
        W=\big( 2\bE[\rho_1(t)]-\bE[\psi_1(t)t] \big)/\bE[\psi_0(t)t].
    \end{split}
\end{equation}

\textit{Computation}: In order to solve the optimization problem given in equation (\ref{eq:TAU-Lasso}), we employ the generalized gradient to minimize the composite objective function consisting of a non-convex term and a non-smooth $\ell_1$-norm penalty term. The generalized gradient of the objective function is defined as $\partial_{\bb} ([\hs_{\tau}^{(i)}]^2+\lambda\|\bb\|_{\ell_1})$ where $\partial_{\bb} [\hs_{\tau}^{(i)}]^2$ associated with the smooth, non-convex, continuously differentiable term is identical to its gradient $\nabla_{\bb} [\hs_{\tau}^{(i)}]^2$ and  $\partial_{\bb} (\lambda\|\bb\|_{\ell_1})$ associated with non-smooth, convex term coincides with its subdifferential \cite{freue2017pense}, \cite{clarke1990optimization}. It follows from the local lipschitzity of the composite objective function, any point $\bar{\bb} \in \bR^p$ at which $\mathbf{0} \in \partial_{\bb} ([\hs_{\tau}^{(i)}]^2+\lambda\|\bb\|_{\ell_1})$  is a local minimum of the $\tau$-Lasso estimation problem. Therefore, the generalized gradient of the objective function wrt $\bb$ may be leveraged to find the local minima of the given estimation problem. It can be shown the generalized gradient of the objective function is equivalent to the sub-gradient of the weighted least square penalized by $\ell_1$-norm except that the weights $w_l^{(i)}(\bb)$ here depend on the unknown $\bb$. Hence, the original optimization problem may be reformulated as follows:
\begin{equation}\label{eq:IRWLasso}
\begin{split}
\hbb^{(i)} &=\argmin_{\bb} \|\bOm^{(i)}(\cby^{(i)}-\cbX^{(i)}\bb)\|_{\ell_2}^2+\lambda^{'}\|\bb\|_{\ell_1},
\end{split}
\end{equation} 
where $\lambda^{'}=2b\lambda/\hs_b^{(i)}$, $\bOm^{(i)}$ is a diagonal matrix whose entries on diagonal are $\sqrt{w_l^{(i)}}$ and $w_l^{(i)}$ is given by,
\begin{equation}
    \begin{split}
        w_l^{(i)}=\frac{\big[w_{\tau}^{(i)}\psi_0(\tr_l^{(i)})+\psi_1(\tr_l^{(i)}) \big]}{\check{r}_l^{(i)}},\\
        w_{\tau}^{(i)}=\frac{\sum_{l=1}^b\big[2\rho_1(\tr_l^{(i)})-\psi_1(\tr_l^{(i)})\tr_l^{(i)} \big]}{\sum_{l=1}^b\psi_0(\tr_l^{(i)})\tr_l^{(i)}}.
    \end{split}
\end{equation}
where the notation $\check{r}_l^{(i)}$ is a shorthand for $\check{r}_l^{(i)}(\bb)$ and $\tr_l^{(i)}=\check{r}_l^{(i)}/\hs_b^{(i)}$. In the spirit of iteratively reweighted least-squares (IRLS) \cite{burrus1994iterative}-\cite{holland1977robust}, we use iteratively reweighted Lasso (IR-LASSO) alternating between finding the weight matrices $\bOm^{(i)}$, refining $\hbb^{(i)}$ and updating $\hs_b^{(i)}$. The M-scale estimates are calculated via fixed-point iterations at each step of IR-LASSO. \\

\subsubsection{Fusing variable selections}
Once parameter estimation and variable selection are performed at each node, the chosen variables are communicated to the fusion center or cloud. In the fusion center, a percentage-based voting rule is used to select the relevant variables for the entire large-scale data set. The selection results of all nodes are combined according to the following rule, that is,
, if a parameter is in the support within $100 \times K$ percent of subsets, it is selected to the support for the complete data set,
$ \hat{\mathcal{S}}=\{j:\sum_{i=1}^{s}\mathbf{1}(\hat{\beta}^{(i)}_{j} \neq 0) /s \geq K \}$.
 The chosen variables $\hat{\mathcal{S}}$ are broadcast into computation and storage nodes and deployed in the second stage of inference which uses the selected variables only.
 
Consider we skip the fusion and directly pass down the variables selected at each of $s$ nodes to the next stage. It is likely that models selected at some subsets of data are either overfitting or undefitting, which results in inaccurate inference results, thereby an appropriate rule of fusion is recommended. \\

%\subsection{Preprocessing: Data with Very High-Dimensional Subsets}

%In order to reduce the computational burden in settings with very high-dimensional subsets, an initial variable screening procedure called Density Power Divergence-SIS (DPD-SIS) \cite{ghosh2020robust} is employed to further reduce the model complexity, the number of variables to an order of sample size. Basically, the DPD-SIS extends the Sure Independence Screening (SIS) to address robustness in the presence of outliers. The robust screening procedure assigns a certain score to each predictor and then predictors are ranked in descending order based on the calculated score. In order to compute the score, the marginal estimate of each regression coefficient is obtained via a minimum DPD estimator and then its absolute value determines the score associated with each predictor. At each node, the DPD-SIS variable screening procedure is utilized to discard a certain number of irrelevant predictors and then distinct subsets with reduced dimensionality is passed down to next step for further processing, that is, $ \cbX \in \bR^{b \times p} \rightarrow \bar{\mathbf{X}} \in \bR^{b \times q}$ with $(q \ll p)$ an order of sample size. A reasonable choice for $q$ is the number of observations within subsets of data, that is, $q=b$. Note that variable screening procedure is dispensable in low-dimensional models, i.e. $q=p$. The algorithmic details of DPD-SIS can be found in \cite{ghosh2020robust}. \\

\subsection{Stage \RN{2}: Robust Inference }
\label{sec:inference_BLFRB}

\subsubsection{Construction of CIs (RSOB-$\tau$)}
In this part, we perform the actual inference over the chosen variables from the model selection stage by using the RSOB-$\tau$. The derived robust $\tau$-estimation equations are used to compute bootstrap replicates instead of MM-estimation equations in \cite{basiri2015robust}. Scalability is achieved by conducting inference in parallel on each distinct subset of data. Moreover, applying the RSOB-$\tau$ on only the selected variables eliminates the bias introduced by the regularization term.

In order to develop low-complexity inference procedures using bootstrap, it is required for the underlying estimator to be expressed as a fixed-point problem. It is well-known that $\tau$-estimator fulfills the above requirement and can be represented as a solution of fixed-point equations as follows:

\begin{equation}
\label{eq:initial_estimate}
  \hbth_b=\bbf(\hbth;\tilde{\bY}_b)   
\end{equation}
where $\bbf:\bR^{|\hat{\mathcal{S}}|+1} \rightarrow \bR^{|\hat{\mathcal{S}}|+1}$ denotes a smooth function, $\tilde{\bY}_b=\big(\cby,\tilde{\bX}\big)$ with the explaining variables chosen in the model selection stage and $\hbth_b$ is a fixed-point of $\bbf$. The terms $\hbth_b$ and $\bbf(\hbth_{b};\tilde{\bY}_b)$ are defined, respectively, as follows:
 
 \begin{equation}
 \begin{split}
    \hbth_b&=\begin{bmatrix}
\hbb_b\\
\\ \hs_b
\end{bmatrix} 
\\
\bbf(\hbth_{b};\tilde{\bY})&=\begin{bmatrix}
\Big( \sum_{l=1}^{b} \hw_l\tbx_{[l]} \tbx_{[l]}^{T}\Big)^{-1}\Big( \sum_{l=1}^{b} \hw_l \cy_l \tbx_{[l]}\Big),\\
\\ \sum_{l=1}^b \hat{v}_l\hr_l,
\end{bmatrix}
 \end{split}
\end{equation}
 where 
 \begin{equation}
     \begin{split}
         \hat{v}_l&=\frac{1}{b \delta_2} \times \frac{\rho_0\big(\tr_l\big)}{ \tr_l},\\
         \hw_l&=\frac{\hw_{\tau}\rho_0^{'}\big(\tr_l\big)+\rho_1^{'}\big(\tr_l\big)}{\hr_l},\\
         \hw_{\tau}&=\frac{\sum_{l=1}^b\Big[ 2\rho_1\big(\tr_l\big)-\rho_1^{'}\big(\tr_l\big)\tr_l\Big]}{\sum_{l=1}^b \rho_0^{'}\big(\tr_l\big)\tr_l},
         \\
         \hr_l&=\cy_l-\tbx_{[l]}^{T}\hbb_b,
         \\
         \tr_l&=\frac{
         \hr_l}{\hs_b}.
     \end{split}
 \end{equation}
$\hw_l$ down-weights the outlying observations. 
In order to construct confidence intervals of the selected variables, the bootstrap replications of $\hbth_b$ are computed as follows:
\begin{equation}
    \hbth_{n,b}^{\star}=\bbf(\hbth_{n,b}^{\star};\tilde{\bY}^{\star}),
\end{equation}
where $\tilde{\bY}^{\star}=\big(\tilde{\bY}_b,\bom^{\star}\big) $ denotes a bootstrap sample of size $n$ randomly drawn with replacement from the given subset of data. The multiplicity of observations is determined by the random weight vector $\bom^{\star} \in \bR^b$ drawn from a multinomial distribution $\big(n,(1/b)\mathbf{1}_b\big)$. Instead of computing a fully iterating bootstrap replicate $\hbth_{n,b}^{\star}$, we can approximate it via a one-step iteration $\hbth_{n,b}^{1 \star}$ as follows:

\begin{equation}
    \hbth_{n,b}^{1\star}=\bbf(\hbth_{b};\tilde{\bY}^{\star}),
\end{equation}
 However, the distribution of $\hbth_{n,b}^{1\star}$ may not exhibit the actual variability of the sampling distribution of $\hbth_{b}$, mainly because all bootstrap replicates $\hbth_{n,b}^{1\star}$ are calculated starting from the same initial value $\hbth_{b}$. It has been shown that one-step iteration of bootstrap replications for many estimators could be adjusted by a correction term to provide  asymptotically true estimate of bootstrap distribution \cite{schucany1991one}. In order to achieve asymptotically correct bootstrap estimates, the one-step improvement of  $\hbth_{n,b}^{1\star}$ with a linear correction term can be written as \cite{salibian2002bootrapping}, \cite{salibian2016robust}:

\begin{equation}
     \hbth_{n,b}^{R\star}=\hbth_{b}+\Big[ \mathbf{I}-\nabla \bbf(\hbth_{b};\tbY_b) \Big]^{-1}\Big(\hbth_{n,b}^{1\star}-\hbth_{b}\Big),
\end{equation}
where $\hbth_{n,b}^{R\star} \in \bR^{|\hat{\mathcal{S}}|+1}$ denotes the linearly corrected one-step bootstrap replication of $\hbth_{b}$ and $\nabla \bbf(\cdot)$ is a gradient matrix with respect to $\bth$. The linear correction term can be obtained by inverting the matrix $\Big[ \mathbf{I}-\nabla \bbf(\hbth_{b};\tbY_b) \Big]$ via the block matrix inversion lemma as described in Appendix.

Computational efficiency is attained because the correction factor and the initial estimate $\hbth_b$ are computed only once for each subset of data. Furthermore, one-step bootstrap replications of $\hbth_b$ are computationally inexpensive.We show in theorem 2 that $\hbth_{n,b}^{R\star}$ would estimate the same limiting distribution as the actual bootstrap distribution $\hbth_{n,b}^{\star}$ under certain regularity conditions.

The algorithm begins with generating $B$ bootstrap samples for each distinct subset of data $\tbY^{(i)}_b=\big(\cby^{(i)},\tbX^{(i)}\big) \in \bR^{b \times (|\hat{\mathcal{S}}|+1)}, i=1,\cdots,s$. Across all computing and storage nodes, linearly corrected one-step bootstrap replicates $\hbb_{n,b}^{R\star,(ij)}, j=1,\cdots,B$ are computed. Then, confidence intervals associated with selected variables from stage 1 are constructed by using bootstrap percentile method for each subset of data. The estimated confidence intervals are communicated from each computing node to the fusion center for performing inference on the complete large-scale data. In the fusion center, the confidence intervals for the complete large-scale data are produced by applying coordinate-wise averaging over upper bounds and lower bounds of transmitted confidence intervals as follows:

\begin{equation}
\begin{split}
     \overline{\text{CI}}_j=\frac{1}{s}\sum_{i=1}^{s} \overline{\text{CI}}_j^{\star(i)}\\
     \underline{\text{CI}}_j=\frac{1}{s}\sum_{i=1}^{s} \underline{\text{CI}}_j^{\star(i)}
\end{split}
\end{equation}
where $\overline{\text{CI}}_j$ and $\underline{\text{CI}}_j$ denote the lower bound and upper bound of confidence interval associated with entry $j$ of the non-zero parameter.

Herein, we assumed that the proportion of outliers withing each subset of data remains below $50\%$, implying the estimated confidence intervals are not corrupted. Therefore, we may avoid the undesirable effects of outliers on the veracity of confidence intervals. In case more than half of observations within each subset of data are contaminated by outliers, we recommend using the adaptive trimmed mean \cite{hogg1967some} or classical trimmed mean with manually set trimming ratio \cite{maronna2019robust}.

\begin{algorithm}
  \DontPrintSemicolon
  \KwData{\textbf{$s$}  distinct subsamples}
  %\KwData{$\bY=\big( \by,\bX \big)$}
  \KwOutput{$\mathbf{CI}$}
  %\Begin{
  % Generate $s$ distinct subsets of size $b$ by randomly resampling without replacement from the complete data set $\bY=\big( \by,\bX \big)$ \; % \tcc{This step can be skipped if there are already $s$ distinct subsets of data}
    \For{each subsample}{
    Generate $B$ bootstrap samples of size $n$ by randomly drawing with replacement as follows:
    $\tilde{\bY}^{\star}=\big(\tilde{\bY}_b,\bom^{\star}\big) $ \;
    Find the initial estimate $\hbth_b$ by solving the fixed-point problem given in equation (\ref{eq:initial_estimate})\;
    Calculate the one-step linearly corrected bootstrap replication corresponding to each bootstrap sample $\hbb_{n,b}^{R\star}$ from equation (\ref{eq:RSOBT_0}) \;
    Compute confidence intervals $\mathbf{CI}^{\star(i)}$ for each subset of data via bootstrap percentile method\;
    }
  Combine the confidence intervals by coordinate-wise averaging 
 % }
  \caption{The RSOB-$\tau$ procedure\label{RSOB-T}}
\end{algorithm}

\section{Statistical Robustness and Consistency Properties}
In this section, the robustness properties of $\tau$-Lasso estimators and linearly corrected one-step replications using RSOB-$\tau$ are characterized by deriving their breakdown points. Furthermore, asymptotic normality of the linearly corrected one-step bootstrap replications is established under certain regularity conditions.

\subsection{Robustness Properties of $\tau$-Lasso Estimators}

The finite-sample breakdown point of $\tau$-Lasso is derived. It measures the largest proportion of observations when arbitrarily replaced by outliers does not cause unbounded maximum bias or equivalently does not break down. Given a subset of data $\cbY=\big(\cby,\cbX \big) \in \bR^{b \times (p+1)}$ randomly drawn from the original data $\bY=\big( \by,\bX \big) \in \bR^{n \times (p+1)}$, the replacement finite-sample breakdown point (FBP) $\epsilon^{*} \big( \hbb;\cbY \big)$ of a regression estimator $\hbb \in \bR^{p}$ is defined as 

 \begin{equation}\label{eq:fbp}
     \epsilon^{*} \big( \hbb;\cbY \big)=\max \{ \frac{m}{b}: \sup_{\cbY_m \in \mathcal{Y}_m} \|\hbb \big(\cbY_m \big)\|_{\ell_2} < \infty \}
 \end{equation}
where the set $\mathcal{Y}_m$ contains all datasets $\cbY_m$ with $m$ $(0<m<b)$ out of the original $b$ observations replaced by arbitrary values. The bounded \textit{supremum} of $\ell_2$ term in the definition is equivalent to having a bounded maximum bias. The following theorem proves the results for finite-sample breakdown point of robust $\tau$-Lasso estimator by extending the theoretical result for that of robust S-PENSE estimator demonstrated in \cite{freue2017pense} to the class of $\tau$-Lasso estimators.\\

\textbf{Theorem 1}: \textit{Suppose $m(\delta)$ is the largest integer smaller than $b \min (\delta,1-\delta)$ for a subset of data $\cbY=\big(\cby,\cbX \big) \in \bR^{b \times (p+1)}$ and $\delta$ defined by equation (\ref{eq:Mscale}). Then the finite-sample breakdown point of the $\tau$-Lasso estimator is bounded from above and below as follows}:

\begin{equation}
    \frac{m(\delta)}{b} \leq \epsilon^{*}(\hbb;\cbY) \leq \delta
\end{equation}
where $\hbb$ denotes the $\tau$-Lasso estimator. \\
\\
\textbf{\textit{Proof}}: The complete proof is given in Supplemental Materials.

With minor modification of the proof given in the \textbf{Supplemental Material}, we can extend the above theorem for data models including an intercept term. 

\subsection{Asymptotic Properties of TSRD-$\tau$}
In this section, we show that the proposed inference procedure will enjoy desirable asymptotic properties for fixed $p$ under certain regularity conditions if the model selection procedure is consistent. The model selection consistency in the first stage would imply that the relevant variables are selected with probability converging to $1$ and the asymptotic distribution of modified bootstrap $\tau$-estimates for non-zero coefficients would be the same asymptotic normal distribution as if the true non-zero coefficients were known in advance. It seems reasonable to conjecture that the model selection procedure is consistent, extending the concept of Bolasso \cite{bach2008bolasso} to robust $\ell_1$-penalized estimators. We explore this possibility in simulations. The asymptotic properties of FRB for $\tau$-regression estimator were proven in Theorem 1 of \cite{salibian2016robust}. However, the inaccuracy in the linear correction factor in the proof is rectified here. The asymptotic properties of BLFRB and FRB for MM-regression estimator are established in \cite{basiri2015robust} and \cite{salibian2002bootrapping}. Also note that in this theorem, $\bb_0 \in \bR^{|\hat{\mathcal{S}}|}$ is a shorthand for $[\bb_0]_{\hat{\mathcal{S}}}$ where parameter vector $\bb_0 \in \bR^p$ as given in equation (\ref{eq:linear_model}) and $\xrightarrow{P}$ denotes convergence in probability.\\

\textbf{Theorem 2} : \textit{Let $\rho_0$ and $\rho_1$ be bounded $\rho$-functions satisfying the properties of bounded $\rho$-function defined by Maronna et al. \cite{maronna2019robust} and have continuous third derivatives. Assume the model selection stage produces consistent estimates, i.e., $\mathbb{P}(\hat{\mathcal{S}}=\mathcal{S})=o(1)$. Let $\hbb_b$ be the $\tau$-regression estimator for a subset of data $\big( \cby,\tbX \big) \in \bR^{b \times (|\hat{\mathcal{S}}|+1)}$ randomly drawn from the original data $\big( \by,\ubar{\bX} \big) \in \bR^{n \times (|\hat{\mathcal{S}}|+1)}$ and $\hs_b$ be the M-scale of residuals for the given subset of data and assume that they are consistent estimators, that is, $\hbb_b \xrightarrow{P} \bb_0$ and $\hs_b \xrightarrow{P} \sigma_0$. Given the following regularity conditions hold}:

\begin{enumerate}
    \item[1.] \textit{The following vectors and matrices exist and are finite}:
    \begin{enumerate}
        \item[1.1]  
          $\mathbb{E}\big[\big(\bar{w}_{\tau}\rho_0^{'}(r)+\rho_1^{'}(r)\big)/r\bx \bx^T \big]^{-1}$
          
        \item[1.2]  
           $\mathbb{E}\big[ \rho_0^{'}(r)\bx \big]$   
           
         \item[1.3]  
           $\mathbb{E}\big[\big(\Bar{w}_{\tau}\rho_0^{''}(r)+\rho_1^{''}(r)\big) \bx \bx^T \big]$
           
         \item[1.4]  
           $ \mathbb{E}\big[
        \big(\bar{w}_{\tau}\rho_0^{''}(r)+\rho_1^{''}(r) \big)r\bx \big]$
    \end{enumerate}
    \item[2.] $\mathbb{E}[\rho_0^{'}(r)r] \neq 0$ \textit{ and finite,}
    \item[3.] $\rho_0^{'}(u)/u$, $\rho_1^{'}(u)/u$, $\big(\rho_0^{'}(u)-\rho_0^{''}(u)u\big)/u^2$ and $\big( \rho_1^{'}(u)-\rho_1^{''}(u)u \big)/u^2$ are continuous.
\end{enumerate}
\textit{where $\bar{w}_{\tau}=\big(2\bE[\rho_1(r)]-\bE[\rho_1^{'}(r)r]\big)/\bE[\rho_0^{'}(r)r]$. Then, the distribution of $\sqrt{n}\big(\hbb_{n,b}^{R\star}-\hbb_b\big)$ converges weakly to the limiting distribution of $\sqrt{b}\big(\hbb_b-\bb_0 \big)$ and consequently to the limiting distribution of $\sqrt{n}\big(\hbb_n-\bb_0 \big)$ as $n$ and $b$ approach infinity.}\\

 \textbf{\textit{Proof}}: The complete proof is presented in Supplemental Materials.

\subsection{Statistical Robustness of RSOB-$\tau$}

In this section, we are interested in studying the robustness properties of the proposed RSOB-$\tau$. The confidence intervals for regression parameters were constructed by using RSOB-$\tau$ quantiles and thus, the breakdown point of quantiles gives an insight into the robustness and reliability of the inferences made using the proposed method. Here, we derive some theoretical results about the breakdown point of quantile estimates of RSOB-$\tau$ and FRB using $\tau$-estimators.

Before proceeding, we define few concepts essential in comprehending the robustness results. Given $t \in \big( 0,1 \big)$, $\hat{q}_t$ is defined as the $t_{th}$ upper quantile of a statistic $\hbb_b$  such that $P[\hbb_b>\hat{q}_t]=t$. According to \cite{singh1998breakdown}, the upper breakdown point of a bootstrap estimate $\hat{q}_t^{\star}$ is defined as the minimum proportion of arbitrarily large outliers in the subset of data $\tbY$ that can drive $\hat{q}_{t}^{\star}$ into infinity. In what follows theoretical results on robustness properties of estimated RSOB-$\tau$ and FRB quantiles using robust $\tau$-estimator are demonstrated. The following theorems are proved by using similar techniques as in the proofs given for BLFRB and FRB quantiles using robust MM-estimator \cite{basiri2015robust} and \cite{salibian2002bootrapping}. Note that the theorems below also hold for subsets of data $\tbY \in \bR^{b \times |\hat{\mathcal{S}}|}$ having $|\hat{\mathcal{S}}|$ predictors selected from model selection stage.
\\

\textbf{Theorem 3}:
\textit{Suppose  $\bY=\big(\by,\bX \big) \in \bR^{n \times (p+1)}$ is a large-scale data following the linear model given in equation (\ref{eq:linear_model}). Except, here no sparsity assumption is imposed on the parameter vector $\bb_0$. Assume that $\bY$ is in general position, i.e., any subset of $p$ observations will result in a unique determination of $\bb_0$. Let $\hbb_n$ be a robust $\tau$-estimate of $\bb_0$ based on the data $\bY$ whose breakdown point is $\epsilon$. Then, the breakdown point of the $t_{th}$ FRB quantile estimate of the regression parameters $[\beta_0]_{(l)} \mbox{, }l=1,\cdots,p$ is determined by $\min(\epsilon_t^{\star},\epsilon)$ where $\epsilon_t^{\star}$ is the smallest $\delta_c \in [0, 1]$ that satisfies the following inequality,}

\begin{equation}
    P\big[Binomial(n,1-\delta_c)<p\big] \geq t
\end{equation}
Note that $\delta_c$ is different from $\delta_0$ and $\delta_1$.

\textbf{\textit{Proof}}: The details of the poof are given in Appendix.
\\

\textbf{Theorem 4}: \textit{Suppose $\cbY=\big(\cby,\cbX \big) \in \bR^{b \times (p+1)}$ is a subset of the large-scale data  $\bY=\big(\by,\bX \big) \in \bR^{n \times (p+1)}$ formed by random resampling without replacement of the original full data set. Except, here no sparsity assumption is imposed on the parameter vector $\bb_0$. Assume that $\cbY$ is in general position, i.e., any subset of $p$ observations will determine a bounded least-square estimate. Let $\hbb_b$ be a robust $\tau$-estimator of $\bb_0$ based on the bag of data $\cbY$, $\epsilon^{*} (\hbb_b,\cbY)$ denotes the finite-sample breakdown point of $\hbb_b$ and $\delta_1=0.5$. Then, all the RSOB-T bootstrap quantiles estimated using the $\tau$-estimator that are constructed over $\cbY$ will have equal asymptotic breakdown point to $\hbb_b$}.\\

\textbf{\textit{Proof}}: A detailed proof is given in Appendix.

\section{SIMULATIONS AND RESULTS}

In this section, the performance of the proposed method is investigated in simulations considering both variable selection  and inference. In particular, correct identification of sparse basis, statistical robustness of bootstrap estimates, the quality of the parameter estimates and confidence intervals are studied. The results in Theorem 2 is validated through computer simulations, indicating the distribution of bootstrap replications by using RSOB-$\tau$ asymptotically converges to the sampling distribution of $\tau$-estimator. The performance of the proposed method is assessed through computer simulations by using different proportions of outliers, large-scale data in both low-dimensional $(p < b \text{ or } n)$ versus high-dimensional $(p \approx n \text{ or } \gg b )$ settings. 
%The performance is compared to an inference method using MM-Lasso and MM-estimators in \cite{mozafari2019robust}.
It is assumed the large-scale data follow a linear regression model where the parameter vector $\bb_0 \in \bR^p$ is sparse with $k_s$ non-zero entries. The measurement noise vector, $\bv$, is an additive white Gaussian noise with a variance (AWGN) $\sigma_v^2=\|\bX \bb_0\|_{\ell_2}^2 10^{-\mbox{SNR}/10}/n$ (SNR in dB).  

\subsection{Simulation Setting}
\label{subsec:sim_sett}
Throughout simulations, the confidence intervals are reported in a nominal level of $100\times(1-\alpha_{cl})\%=90\%$ with $\alpha_{cl}=0.1$. 
 We create a decreasing grid of $70$ lambda values with logarithmic spacing of $1.1$, spanning $\lambda_1$ to $\lambda_{70}$ where $\lambda_1$ is set to $\lambda_{\max}$. The maximum number of iteration in $\tau$ and MM-Lasso estimators is fixed at $30$. The robustness of DPD-SIS procedure is adjusted by a tuning parameter $\alpha$ set to $0.4$ (stable for  a range of contamination levels) to provide robustness without significant loss in efficiency. The proposed algorithm was implemented in MATLAB except for the estimation of initial S-Lasso which was done in R using PENSE \cite{kepplingerpackage}. It provides a good initial estimate by constructing clean subsamples of data, potentially removing outlying observations. This is achieved by using the principal sensitivity components (PSCs) for EN estimator and removing the observations with most extreme PSCs from the subsamples in an iterative manner. A detailed description of initialization for PENSE can be found in Supplementary Materials of \cite{freue2019robust}. In addition, we used the Dual Augmented Lagrangian \cite{tomioka2011super} implementation to solve the IR-LASSO, and MM- and S-estimators as in \cite{croux2004robust}.
 
 The outliers in all simulations except for \textbf{Section \ref{subsec:robust_boot}} are introduced by randomly choosing
the observations in y and replacing them with random values
chosen from a standard Gaussian distribution with $\sigma_e=250$.

\subsection{Data Standardization}
Across all simulations for estimation problems using robust Lasso, it is assumed the linear regression model has an intercept component and all columns of the augmented regression matrix $[\mathbf{1}_{b \times 1} \cbX^{(i)}]$ except the first one are robustly standardized by centring the columns using a bisquare location estimator and scaling them using bisquare scale estimators \cite{maronna2011robust}. The response vector $\cby^{(i)}$ is centred using the bisquare location estimator \cite{maronna2011robust}.

\subsection{Calibration of Tuning Constants $c_0$ and $c_1$}
In order to tune $c_0$ for the M-scale of residuals within all the robust estimators discussed in this paper, we set $\delta_{i}=0.5$ for $i=0,1,2$ to achieve the maximum robustness against outliers. $\delta_{i}$ controls the breakdown point according to Theorem 4.1 in \cite{freue2017pense} and $\delta_0$ is associated with initial S-Lasso estimator. It’s  worth  mentioning $0.5$ is  the  largest  value $\delta_{i}$ can  take on. Note that $c_0$ is tuned for the desired breakdown point with the assumption $\lambda=0$. In particular, it is recommended to have maximum robustness because bootstrapping may exacerbate the malicious behaviour of outliers in resampled data sets. 
The tuning constant $c_1$ of $\tau$-Lasso and MM-Lasso is respectively adjusted to $6.08$ and $4.68$ to attain $95\%$ efficiency under Gaussian errors when $\lambda=0$. Likewise, $c_1=6.08$ for $\tau$-estimator and $c_1=4.68$ for MM-estimator are adjusted to provide $95\%$ efficiency under Gaussian errors for bootstrap replications.

\subsection{Choice of Fusion Parameter $K$}
$K$ denotes the proportion of data subsets classified the given variable as relevant. Herein, the fusion parameter $K$ for voting rule is set to $0.5$, indicating the variables selected within at least $50\%$ of subsets are regarded as relevant variables, i.e, majority voting scheme.  The specific value of $0.5$ is a compromise between reducing the false positive rate and maximizing the selection of true relevant variables. 

\subsection{Choice of Batch Size $b$}
The choice of batch size $b$ is a design parameter, depending on
computational and storage resources available plus either model selection
performance or prediction performance. On one hand, randomly partitioning the entire large-scale data of size $n$
into subsets of size $b$ and independently analyzing smaller subsets of data leads to
significant computational speed-up and data storage benefits  \cite{kleiner2014scalable, chen2014split}.
On the other hand, the choice of batch size $b$ trades off between overfitting and underfitting in the variable selection procedure in additional to statistical correctness of confidence intervals. Setting the batch size $b$ to very large values results in rejecting relevant variables. This limits the invertibility of sub-design matrices as the eigenvalues are no longer strictly positive. However, one shall tune $b$ to be large enough to discard the irrelevant variables. Moreover, very small batch size $b$ equals to large number of data subsets results in wasted computation as
we are occupying unnecessarily many computation and storage nodes while only a small proportion of them would be needed. The known information-theoretic results \cite{wainwright2009information} impose a condition on the batch size $b$ to ensure the support recovery for any method, $b\geq c_s k_s \log(p)$ for a sufficiently large constant $c_s$. On the other hand, BLB \cite{kleiner2014scalable} requires the batch size to some values $n^{0.6}<b< n^{0.9}$ such that the statistical correctness is obtained. Combining these two constraints, one may choose the batch size in the range of  $\max(n^{0.6},c_s k_s \log(p))\leq b \leq n^{0.9}$. We then need to take into account practical implications of computational constraints and choose an appropriate batch size $b$.Given the batch size $b$, one can obtain the number of subsamples $s$ and $\gamma$ in a straightforward manner.

We provide an example describing an appropriate batch size $b$. Let $n=1000000$, $b=500$ and $k_s=50$, we then obtain that the minimum batch size from support recovery point of view is $b\geq 311 c_s$. Setting the constant $c_s=30$, sufficiently large for sparse recovery, implies that $b\geq 9400$. On the other hand, reliable statistical correctness is attained by BLB for $b\geq (n^{0.6}\approx 4000)$. We shall occupy many resources over $100$ computing cores. If we set $b=n^{0.75}\approx 32000$, we shall need $33$ computing cores and computations over $32000$ is manageable by single computing cores, which gives a good compromise between computational burden and statistical efficiency. Setting $b\approx n^{0.9}$ would mean almost $4$ subsets of size $250000$ which require higher computational power for each node and not very desirable.

\subsection{Tuning of regularization parameter}  In order to calibrate the robust penalized estimation problem, $\lambda_{\max}$ is initially estimated by using the method introduced by Khan et al. \cite{khan2007robust} and then improved upon via a binary search \cite{smucler2017robust}.  A set of candidate lambdas in decreasing order starting from $\lambda_{max}$ is formed. Selection of $\lambda$ is carried out through a robust version of Bayesian Information Criterion (BIC) \cite{schwarz1978estimating}, defined as
 
 \begin{equation}
     \mbox{RBIC}(\lambda)=b\log (\hs_b^2(\lambda))+C(b,p)\|\hbb(\lambda)\|_{\ell_0}
 \end{equation}
 where $\hs_b$ denotes the robust M-scale of residuals and $C(b,p)$ is set to $\log(b)$ for settings where the dimensionality $p$ is much smaller than the sample size $n$. $C(b,p)=\log(\log(b))\log(p)$ is chosen for $\log(p)/b \rightarrow 0$ as $p \rightarrow \infty$ to attain better empirical performance \cite{fan2013tuning,kim2012consistent, ghosh2020ultrahigh}. Here, standard BIC is modified by replacing the non-robust estimate of scale with the robust M-estimate of scale to deal with outliers. Finally, the optimal tuning parameter $\lambda$ minimizes the RBIC over the pre-defined grid of lambdas
 
 \begin{equation}
     \lambda^{*}=\argmin_{\lambda \in \Lambda} \mbox{RBIC}(\lambda).
 \end{equation}
 
\subsection{Scenarios}
We consider the following scenarios for which simulation studies are carried out.
\begin{itemize}
    \item \textbf{Scenario 1:}  We set the simulation parameters as follows: $n=27000$, $p=30000$, $b=900$ ($p/b=33.33$), $\gamma=0.6667$, $\mbox{SNR}=15 \mbox{ dB}$, $B=1000$.  $\bb_0 \in \bR^p$ is sparse with $k_s=40$ non-zero entries. $[\bb_0]_{\mathcal{S}}$ is set to $3\times \mathbf{1}_{\mathcal{S}}$ and their positions are chosen randomly. The covariate vectors $\bx_{[i]}, i=1,\cdots,n$ are drawn independently from a multivariate Gaussian distribution $\mathcal{N}\big(\mathbf{0},\boldsymbol{\Sigma}\big)$ with $\Sigma_{ij}=\rho^{|i-j|}$ (Toeplitz covariance structure, $\rho=0.5$).
    \\
    
    \item \textbf{Scenario 2:} The simulation parameters are set as follows: $n=2000000$, $p=80$, $b=40000$ ($p/b=1/500$), $\gamma=0.730367$, $\mbox{SNR}=30 \mbox{ dB}$, $B=400$.  $\bb_0 \in \bR^p$ is sparse with $k_s=20$ non-zero entries. $[\bb_0]_{\mathcal{S}}$ is set to $3\times \mathbf{1}_{\mathcal{S}}$ and their positions are chosen randomly. Explaining variables $\bx_{[i]}$ in the regression matrix are i.i.d, randomly drawn from a multivariate Gaussian distribution $\mathcal{N}\big(\mathbf{0},\mathbf{I}_p\big)$. 
    \\
    
    \item \textbf{Scenario 3:} The simulation parameters are set as follows: $n=80000$, $p=100$, $b=4000$ ($p/b=1/40$), $\gamma=0.73466$, $\mbox{SNR}=15 \mbox{ dB}$, $B\!=\!300$. The regression matrix $\bX$ is randomly generated from mutually independent observations drawn from $\mathcal{N}(\mathbf{0},\boldsymbol{\Sigma})$ with $\Sigma_{ij}=\rho^{|i-j|}$ (Toeplitz covariance structure, $\rho\!=\!0.5$). $\bb_0 \in \bR^p$ is sparse with $k_s=10$ non-zero entries. $[\bb_0]_{\mathcal{S}}$ is set to $3\times \mathbf{1}_{\mathcal{S}}$ and their positions are chosen randomly. 
    \\
    
    \item \textbf{Scenario 4:}
    We set the simulation parameters as follows: $n=4900$, $p=6000$, $b=350$ ($p/b=17.14$), $\gamma=0.6895$, $\mbox{SNR}=15 \mbox{ dB}$, $B=1000$.  $\bb_0 \in \bR^p$ is sparse with $k_s=10$ non-zero entries. $\bb_0$ is set to 
    \begin{equation*}
      \bb_0=[2.5, 2.5, 2.5, 2, 3, 3, 3, 3.5, 3.5, 3.5, \mathbf{0}_{p-k_s}^T]^T  
    \end{equation*}
     The covariate vectors $\bx_{[i]}, i=1,\cdots,n$ are drawn independently from a multivariate Gaussian distribution $\mathcal{N}\big(\mathbf{0},\boldsymbol{\Sigma}\big)$ with $\Sigma_{ij}=\rho^{|i-j|}$ (Toeplitz covariance structure, $\rho=0.5$).\\
    
    \item \textbf{Scenario 5:}
    We set the simulation parameters as follows: $n=20000$, $p=80$, $\mbox{SNR}=15 \mbox{ dB}$.  $\bb_0 \in \bR^p$ is sparse with $k_s=15$ non-zero entries. $\bb_0$ is set to 
    \begin{equation*}
      \bb_0=[3.5, 3.5, 3.5, 5, 5, 5, 2.5, 2.5, 2.5, 1.5, 2 \times \mathbf{1}_5^T, \mathbf{0}_{p-k_s}^T]^T  
    \end{equation*}
     The covariate vectors $\bx_{[i]}, i=1,\cdots,n$ are drawn independently from a multivariate Gaussian distribution $\mathcal{N}\big(\mathbf{0},\boldsymbol{\Sigma}\big)$ with $\Sigma_{ij}=\rho^{|i-j|}$ (Toeplitz covariance structure, $\rho=0.5$).
    
\end{itemize}

\subsection{Variable Selection Performance with Different $\ell_1$-penalized Estimators}

In this subsection, we first substitute different $\ell_1$-penalized estimators with $\tau$-Lasso estimators used in the model selection stage of TSRD-$\tau$ introduced in \textbf{Section \ref{sec:TSRD}}. We then carry out simulations to compare the variable selection performance of the resulting procedures with that of TSRD-$\tau$ and TSRD-MM. For this purpose, we use $\ell_1$-penalized estimators such as RA-Lasso \cite{zheng2017robust} and Sparse-LTS \cite{alfons2013sparse}. The former is implemented in MATLAB by following exactly the computation algorithm described in \cite{zheng2017robust} whereas we use the well-known R package \textbf{robustHD} for Sparse LTS regression \cite{alfons2021robusthd}. We run the simulations on the synthetic data set described by \textbf{Scenario 5} except for setting $\text{SNR}=10 \text{ dB}$ with normal errors. In this experiment, we introduce contamination in the regression matrix $\bX$ and the response vector $\by$, simultaneously. For each simulation, we consider four contamination schemes as follows:

\begin{itemize}
    \item \textbf{Scheme 1:} $10 \%$ of observations in the response vector $\by$ are replaced with random values drawn from standard Gaussian with $\sigma_e=250$. We also replace the corresponding observations in $\bX$ with random values chosen from standard multivariate Gaussian distribution with $\mathbf{\Sigma}=\sigma_e^2 \mathbf{I}_{p}$. (large-variance outliers) 
    
    \item \textbf{Scheme 2:} $10 \%$ of observations in the response vector $\by$ are replaced with random values drawn from a Gaussian distribution $\mathcal{N}(250,1)$. We also replace the corresponding observations in $\bX$ with random values chosen from a multivariate Gaussian distribution  $\mathbf{N}(50 \times \mathbf{1}_p,\mathbf{I}_p )$. (gross outliers)
    
    \item \textbf{Scheme 3:} The additive noise is heavy-tailed Student's $t$-distribution with one degree of freedom. 
    
    \item \textbf{Scheme 4:} The outlier contamination follows the same procedure as in Scheme 1 with the assumption of heavy-tailed additive errors, thereby combining Scheme 1 and 3.
    
\end{itemize}

Across all schemes, the same set of observations are replaced with outliers. Herein, we perform a Monte-Carlo study of 20 trials where a random realization of outlier in $\by$ is used at each trial. \textbf{Table \ref{tab:comparison_msel}} shows the result of variable selection method with different $\ell_1$-penalized estimators for four contamination schemes, averaged over 20 trials. We observe that the proposed variable selection algorithms using $\tau$-Lasso estimator and MM-Lasso estimator perfectly recover the true relevant variables while keeping the false positive rates low. In contrast, variable selection method using Sparse-LTS results in significantly larger false positive rates for smaller subsample sizes under contamination schemes 1 and 2. Except for contamination scheme 2 (heavy-tailed errors), the variable selection with RA-Lasso results in extremely overfitted models.

\vspace{-10pt}
\begin{table*}
\caption{ \textbf{Comparison of Variable Selection Performance with Different $\ell_1$- penalized Estimators:} The Proposed Algorithms using MM-Lasso and $\tau$-Lasso Exhibit a Reliable Performance in Recovering The True Sparse Basis ($\text{TPR=1}$) while Keeping False Positive Rates low. Variable Selection Method using Sparse-LTS fails to suppress the False Positives for Schemes  1 and 2 (Smaller subsample Size). Variable Selection with RA-Lasso Performs Reliably Only for Scheme 3, Heavy-Tailed Errors with No Outlier Contamination}
\label{tab:comparison_msel}
\centering
\ra{1.2}
\begin{threeparttable}
\begin{tabular}{@{}llllclllclllclll@{}} \toprule
 & \multicolumn{3}{c}{scheme 1} & &\multicolumn{3}{c}{ scheme 2}  & &\multicolumn{3}{c}{ scheme 3} & &\multicolumn{3}{c}{ scheme 4}\\
\cmidrule{2-4} \cmidrule{6-8} \cmidrule{10-12} \cmidrule{14-16} 
 & TP & FP & CER  && TP & FP & CER && TP & FP & CER && TP & FP & CER \\ \midrule
  \multicolumn{4}{c}{$b=625$}&& & & && & &\\
\cmidrule{1-4} 
TSRD-$\tau$
& 1 & 0.0085 & 0.0069 && 1 & 0 & 0 && 1& 0 & 0 &&   1 & 0 & 0\\
TSRD-MM & 1& 0& 0 && 1& 0& 0 && 1 & 0 & 0 &&  1 & 0 & 0 \\
Sparse-LTS
 & 1& 0.14& 0.1138 && 1 & 0.2008& 0.1631 && 1 & 0.0008 & 0.0006 &&  1 & 0.0023 & 0.0019\\
RA-Lasso
 & 1& 1& 0.8125 && 1 & 0.1461 & 0.1187 && 1 & 0 & 0 &&  1 & 1 & 0.8125\\
 \midrule
 \multicolumn{4}{c}{$b=800$}&& & & && & &\\
\cmidrule{1-4} 
TSRD-$\tau$
& 1 & 0 & 0 && 1 & 0 & 0 && 1 & 0 & 0 &&  1 & 0 & 0\\
TSRD-MM & 1& 0& 0 && 1& 0& 0 && 1 & 0 & 0&& 1 & 0 & 0\\
Sparse-LTS
 & 1& 0.0146& 0.0119 && 1 & 0.05 & 0.0406 && 1 & 0 & 0 &&  1 & 0 & 0\\
RA-Lasso
 & 1 & 1 & 0.8125 && 1 & 0.2477 & 0.2019 && 1 & 0 & 0 &&  0.9966 & 0.9938 & 0.8081 \\
 \midrule
 \multicolumn{4}{c}{$b=1000$}&& & & && & &\\
\cmidrule{1-4} 
TSRD-$\tau$
& 1 & 0.0015 & 0.0013 && 1 & 0 & 0 && 1 & 0 & 0 &&  1 & 0 & 0\\
TSRD-MM & 1& 0& 0 && 1& 0& 0 && 1 & 0 & 0 &&  1 & 0 & 0\\
Sparse-LTS
 & 1& 0.0023& 0.0019 && 1 & 0.0077& 0.0062 && 1 & 0 & 0 &&  1 & 0 & 0\\
RA-Lasso
 & 0.92 & 0.9231 & 0.7650 && 1 & 0.4661 & 0.3787 && 1 & 0 & 0 &&  0.3733 & 0.3569 & 0.4075\\
\bottomrule
\end{tabular}
\end{threeparttable}
\end{table*}

\subsection{Robustness of Bootstrap Replications}
\label{subsec:robust_boot}
In this subsection, the robustness of RSOB-$\tau$ is quantified by an uncertainty measure in comparison to BLB. In particular, we calculate the standard deviation based on the bootstrap replications produced by both methods and verify the results in Theorem 4 by assessing the relative error. The bootstrap estimate of standard deviation is computed as follows: 

\begin{equation}
    \widehat{\mbox{SD}}(\hbb_n)=\frac{1}{|\hat{\mathcal{S}}|}\sum_{l=1}^{|\hat{\mathcal{S}}|} \Bigg(\frac{1}{s}\sum_{i=1}^{s}\bigg(\sum_{j=1}^{B}\frac{\Big([\hbb_{n,b}^{\star(ij)}]_l-[\hbb_{n,b}^{\star(i.)}]_l\Big)^2}{B-1}\bigg)^{1/2}\Bigg)
\end{equation}
where $[\hbb_{n,b}^{\star(i.)}]_l$ is given by:

\begin{equation}
    [\hbb_{n,b}^{\star(i.)}]_l=\frac{1}{B}\sum_{j=1}^{B} [\hbb_{n,b}^{\star(ij)}]_l
\end{equation}
In order to measure how accurate the bootstrap estimate of standard deviation approximates the average standard deviation of $\hbb_n$, we use the relative error criterion that is defined as follows:

\begin{equation}
    \varepsilon=\frac{\widehat{\mbox{SD}}(\hbb_n)-\overline{\mbox{SD}}(\hbb_n)}{\overline{\mbox{SD}}(\hbb_n)}
\end{equation}
where the average standard deviation $\overline{\mbox{SD}}(\hbb_n)$ is defined as $\sigma/\sqrt{n \mathcal{O}}$ based on asymptotic covariance of $\tau$-estimator \cite{yohai1988high}. Here, $\mathcal{O}$ is set to $1$ for least square estimator in BLB and $\mathcal{O}=0.95$ for $\tau$-estimator in RSOB-$\tau$ tuned to have $95\%$ Gaussian efficiency. 

First, we show that even one outlying observation could drive  $\widehat{\mbox{SD}}(\hbb_n)$ based on bootstrap estimates obtained by BLB into infinity whereas those based on bootstrap replications obtained by the RSOB-$\tau$ remain resistant to outlier. Herein, we run the simulations on the synthetic data set described by \textbf{Scenario 2}. In this experiment, a data point within the original data set is randomly drawn and its response vector is multiplied by an extreme value $\alpha_o$ powers of $10$ to imitate the situation where outlier is introduced by misplacement of decimal point. In regard to model selection, the robust and non-robust two-stage methods exhibit reliable performance in selecting the true sparse basis of the parameter vector with a $\text{TP}=1$ and $\text{CER}=0$. One might have expected that the non-robust inference would fail in model selection. However, only the bag of data containing the outlier yielded unreliable estimates but the voting scheme in the fusion center reduces the adversarial effect of outlier.

\begin{figure} 
	\centering
	\includegraphics[width=10cm]{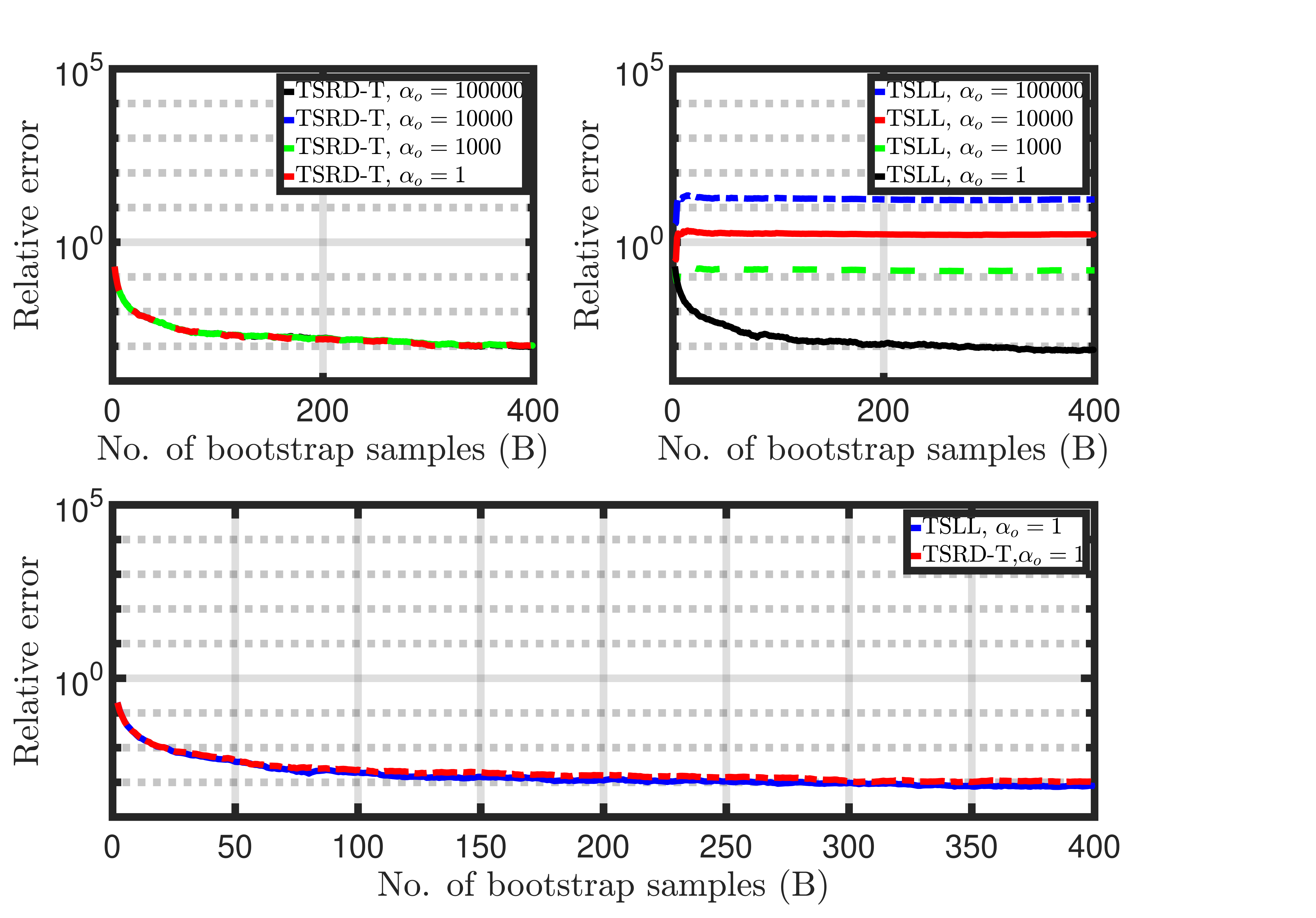}
	\vspace{-20pt}
	\caption{The presence of only one extreme outlier in the data can drive the bootstrap estimate of standard deviation obtained by BLB in two-stage Lasso-LS into infinity. However, the bootstrap estimates of standard deviation obtained by RSOB-$\tau$ in TSRD-$\tau$ remains almost unaltered to the presence of one extreme outlier. The curves produced by TSRD-$\tau$ for all different values of $\alpha$ overlap, implying that one outlier has almost no effect on bootstrap estimates of standard deviation based on TSRD-$\tau$.}
	\label{fig:Rerr_robust}
\end{figure}

In the stage 2, the bootstrap estimates of $\widehat{\mbox{SD}}(\hbb_n)$ are computed based on the data set generated by the selected predictors from the model selection stage. As it is observed in \textbf{Fig. \ref{fig:Rerr_robust}}, both two-stage algorithms TSRD-$\tau$ and TSLL perform remarkably well in terms of relative error when there are no outliers present within the data set. However, the bootstrap estimates of standard deviation obtained by BLB are severely influenced by the presence of even one outlier. As the magnitude of $\alpha_o$ increases, the relative error gets larger, implying that BLB is not robust to outliers. On contrary, the relative error of standard deviation obtained by TSRD-$\tau$ is not influenced at all by the presence of one outlier regardless of its magnitude.

\begin{figure}
	\centering
    \includegraphics[width=14cm, trim = 0.4cm 0.5cm 0.5cm 0.1cm, clip]{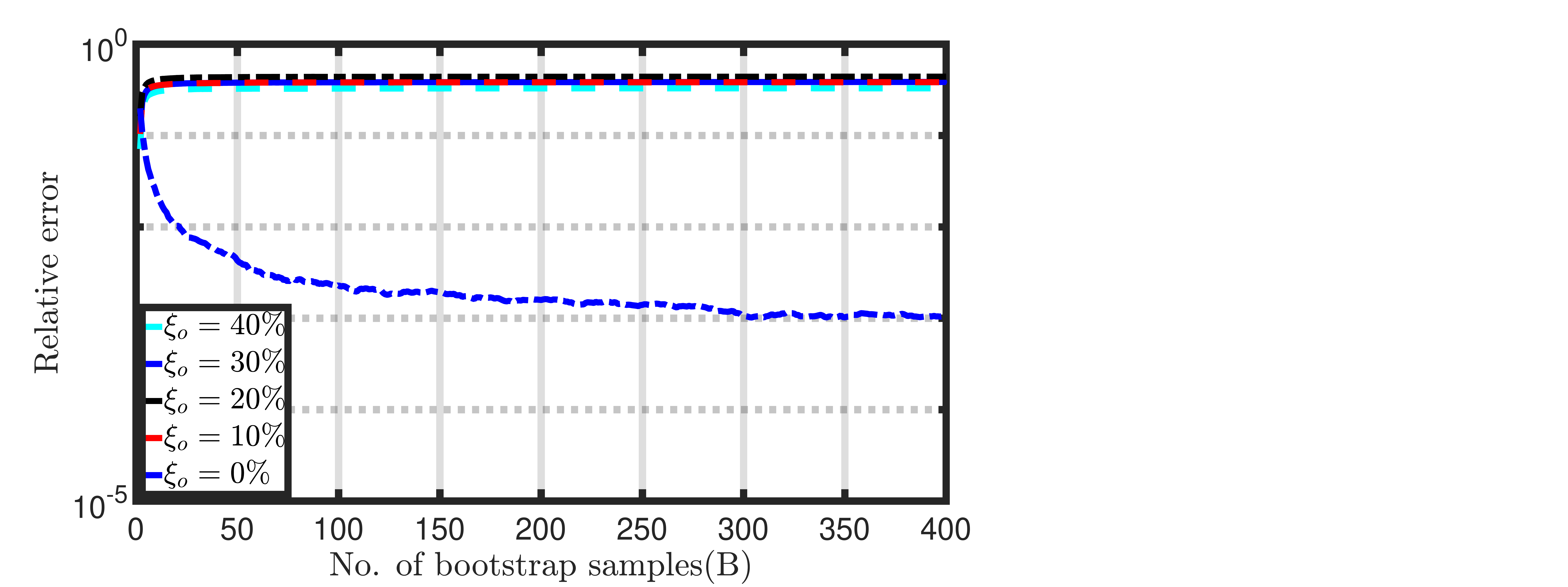}
    \vspace{-15pt}
	\caption{The proposed inference method RSOB-$\tau$ at the stage 2 of TSRD-$\tau$ exhibit strong resilience to outlier even when $40\%$ of observations are contaminated by outliers and relative error of standard deviation remains bounded.}
	\label{fig:Rerr_robust_severe}
\end{figure}

Theorem 4 states the upper breakdown point of RSOB-$\tau$ bootstrap quantile estimates is $0.499$ for the simulation set-up described above. In order to examine the robustness of bootstrap replications, we show that the TSRD-$\tau$ bootstrap replications are robust in the face of outliers even if the data is contaminated severely by outliers. In this experiment, the outliers are introduced by randomly choosing a percentage of the observations in $\by$ and multiplying them with a random value $\alpha_o=100000$. The proposed method TSRD-$\tau$ correctly recovers the true sparse basis with zero false positive rate even in the presence of severe contamination. As shown in \textbf{Fig. \ref{fig:Rerr_robust_severe} }, the bootstrap estimate of standard deviation based on the RSOB-$\tau$ is only slightly influenced by the outliers at contamination levels as high as $40 \%$, hence verifying the results in Theorem 4. In other words, the impact of outliers on the bootstrap estimates is bounded. Note that the curves in \textbf{Fig. \ref{fig:Rerr_robust}} and \textbf{Fig. \ref{fig:Rerr_robust_severe} } are obtained by averaging over 15 trials of Monte Carlo simulations

\subsection{Statistical Convergence }
\begin{figure} [H]
	\centering
	\includegraphics[width=9.5cm]{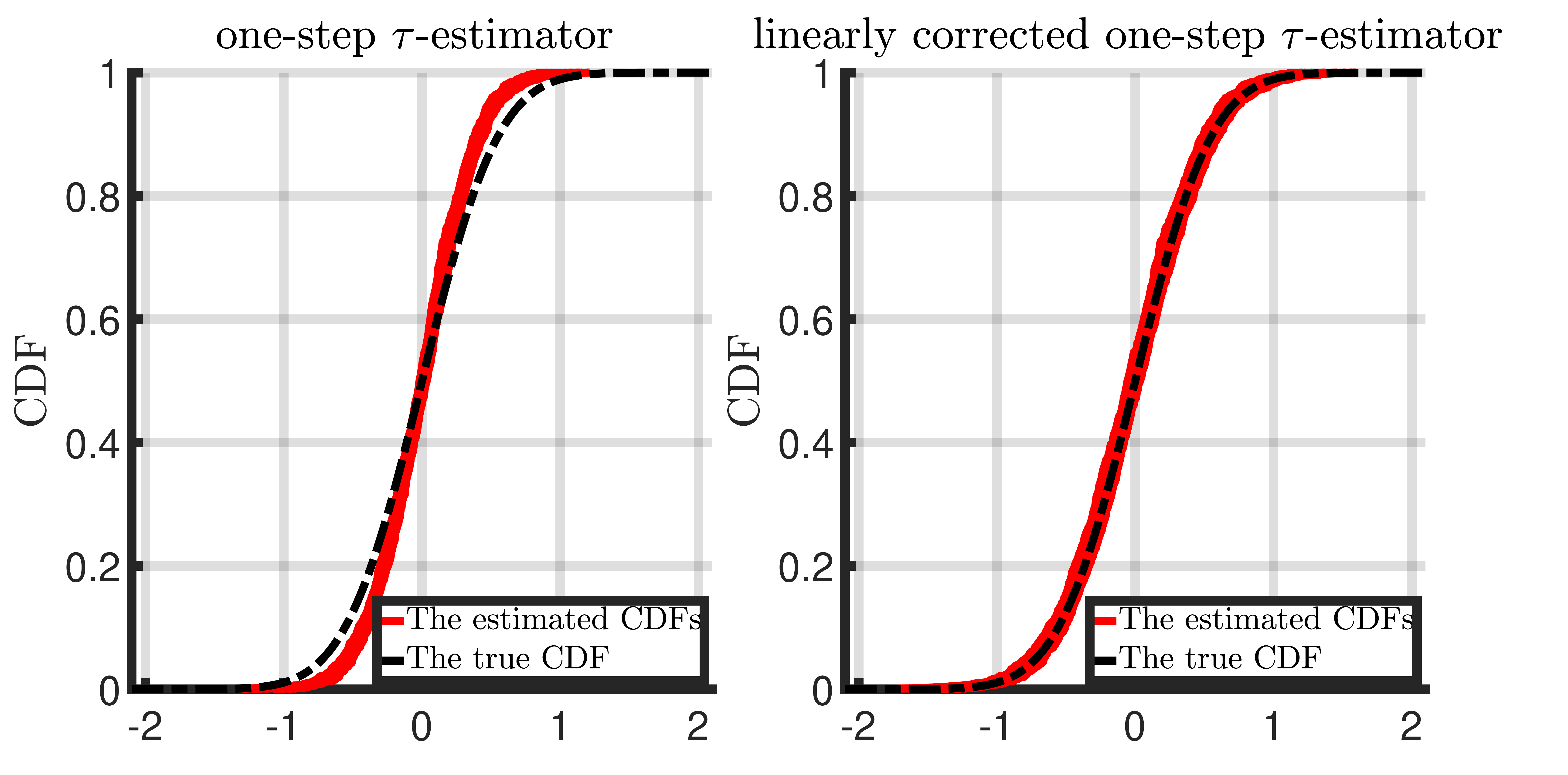}
	\vspace{-25pt}
	\caption{The distribution of bootstrap replicates produced by linearly corrected one-step $\tau$-estimator in RSOB-$\tau$ overlaps the true limiting distribution of $\tau$-estimator whereas the distribution of bootstrap replicates produced by one-step $\tau$-estimator underestimates the variability of the true limiting distribution of $\tau$-estimator. }
	\label{fig:convergence}
\end{figure}

In this subsection, the correctness of Theorem 2 is verified by computer simulations. In other words, we show the distribution of $\sqrt{n}\big(\hbb_{n,b}^{R\star}-\hbb_b\big)$ converges to the limiting distribution of $\sqrt{n}\big(\hbb_{n}-\bb_0\big)$ as $n$ and $b$ approach infinity. Here, we run the simulations on the synthetic dataset described by \textbf{Scenario 2}, the same settings as in section \ref{subsec:robust_boot} are used for the sake of convenience to study the statistical convergence. We assume that large-scale data is not contaminated by outliers. However, the number of bootstrap samples within each subset of data is set to $1000$. Under the condition $\tau$-estimator is tuned for $95 \%$ normal efficiency, the limiting distribution of $\sqrt{n}\big(\hbb_{n}-\bb_0\big)$ obeys a multivariate Gaussian distribution $\mathcal{N}\big( \mathbf{0},(\sigma^2/0.95 )\mathbf{I}_{k_s}\big)$. The distribution of $\sqrt{n}\big(\hbb_{n,b}^{R\star}-\hbb_b\big)$ is formed by randomly drawing a subset $\big(\cby,\tbX \big)$ from the original data set $\big(\by,\ubar{\bX} \big) \in \bR^{n \times |\hat{\mathcal{S}}|+1}$, computing the initial $\tau$-estimate $\hbb_b$ and performing one-step linear correction of initial $\tau$-estimates for bootstrap samples by using the derived equations. The plot on the right-hand side of \textbf{Fig. \ref{fig:convergence}} shows the empirical distribution of $\sqrt{n}\big(\hbb_{n,b}^{R\star}-\hbb_b\big)$ overlaps the true limiting distribution of $\sqrt{n}\big(\hbb_{n}-\bb_0\big)$ for all elements of $\hbb_{n,b}^{R\star}$. On contrary, the plot on the left-hand side of \textbf{Fig. \ref{fig:convergence}} shows the empirical distribution of $\sqrt{n}\big(\hbb_{n,b}^{1\star}-\hbb_b\big)$ underestimates the variability of the true limiting distribution of $\sqrt{n}\big(\hbb_{n}-\bb_0\big)$ for all elements of $\hbb_{n,b}^{1\star}$. Therefore we can conclude the distribution of $\sqrt{n}\big(\hbb_{n,b}^{R\star}-\hbb_b\big)$ provides a reliable approximation of the distribution of $\sqrt{n}\big(\hbb_{n}-\bb_0\big)$. 

\vspace{-10pt}

\subsection{Computational Complexity}

\begin{figure} [H]
	\centering
	\includegraphics[width=7cm, trim = 0.2cm 0cm 0.5cm 0.1cm, clip]{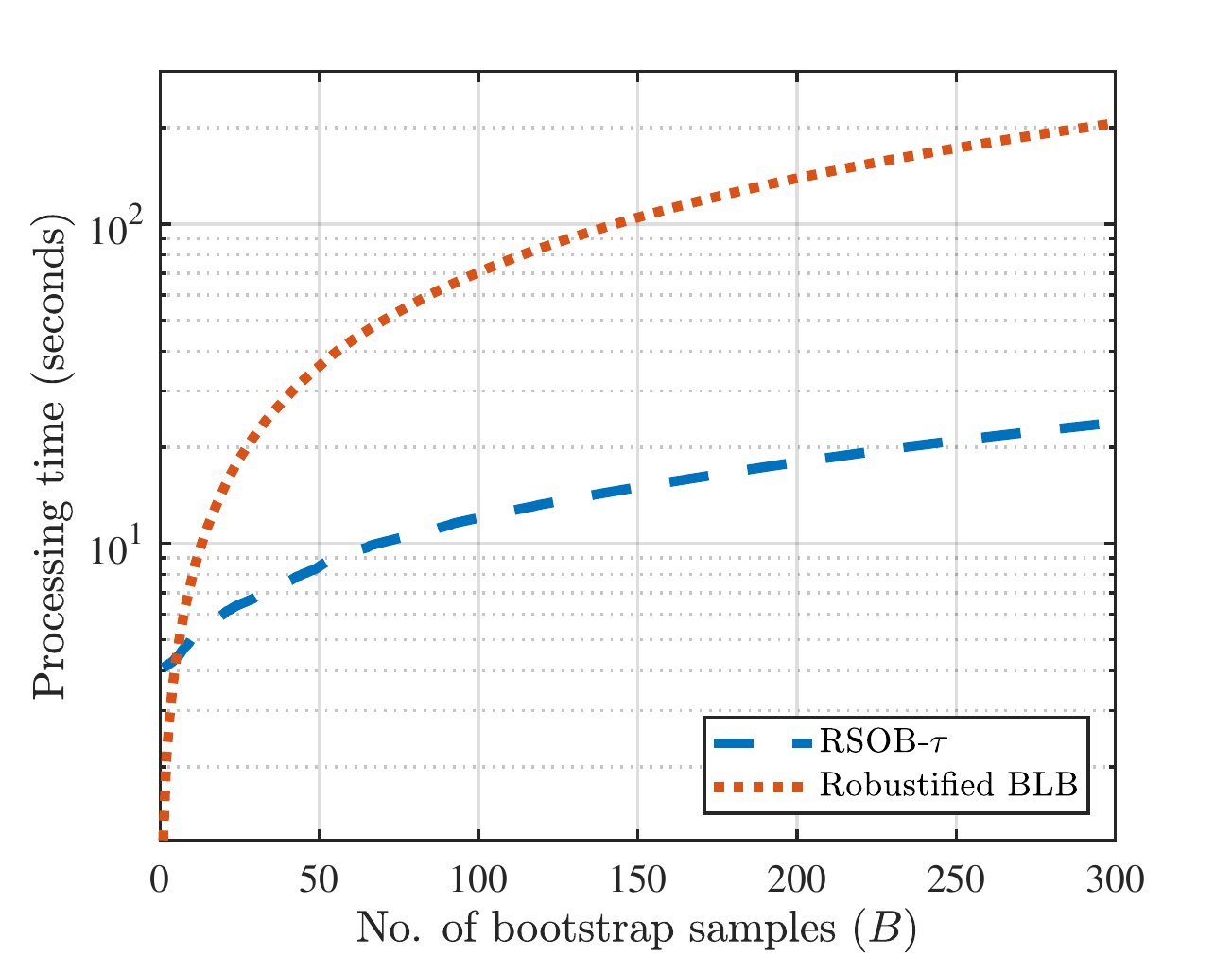}
	\vspace{-5pt}
	\caption{The RSOB-T method employing linearly corrected one-step $\tau$-estimator is significantly faster than robustified BLB. }
	\label{fig:time}
\end{figure}

 In this subsection, we compare the computational complexity of the proposed distributed inference method, RSOB-$\tau$ at stage 2 of TSRD-$\tau$ to a robust realization of BLB method employing $\tau$-estimator for computing bootstrap replicates. We run the simulations on the synthetic data illustrated by \textbf{Scenario 3} where the proportion of outliers is set to $10 \%$. The experiment was conducted in parallel on a single node of a high-performance computing cluster (Triton) where $22$ computing cores and $14$ GB of memory were requested and a Dell PowerEdge C$4130$ node was granted. The cumulative processing time is recorded after each iteration where new set of bootstrap samples are successively added to the bags. The cumulative processing time of RSOB-$\tau$ versus robustified BLB is demonstrated in \textbf{Fig. \ref{fig:time}}. As the number of bootstrap samples increases, the proposed RSOB-$\tau$  requires significantly less processing time in comparison to robustified BLB. This implies the RSOB-$\tau$ achieves remarkably higher computational efficiency. 

\vspace{-20pt}
\vspace{-20pt}
\subsection{Overfitting versus Underfitting and their implications on inference}
In order to make valid inferences, we shall ensure the model selection procedure performs reliably. Herein, we examine how confidence intervals are influenced by overfitting and underfitting in the model selection. To do so, we run the simulations on the synthetic data described by \textbf{Scenario 5} where the proportion of outliers is set to $10\%$. We set the number of data partition $s=40$.   
We study the validity of the statistical inference procedure TSRD-$\tau$ when model selection can not perfectly recover the true support, resulting in either overfitted models or underfitted models.
We first consider a scenario where the model selection procedure fails to reject all irrelevant variables associated with zero coefficients of the parameter vector, resulting in an extremely overfitted model. We further consider the case that the model selection procedure fails to select all relevant variables and $10$ out of $15$ relevant variables are classified as irrelevant variables, resulting in an underfitted model.

\begin{figure} [H]
	%\centering
	\includegraphics[width=10cm,trim = 0.3cm 0.5cm 0.7cm 0.1cm, clip]{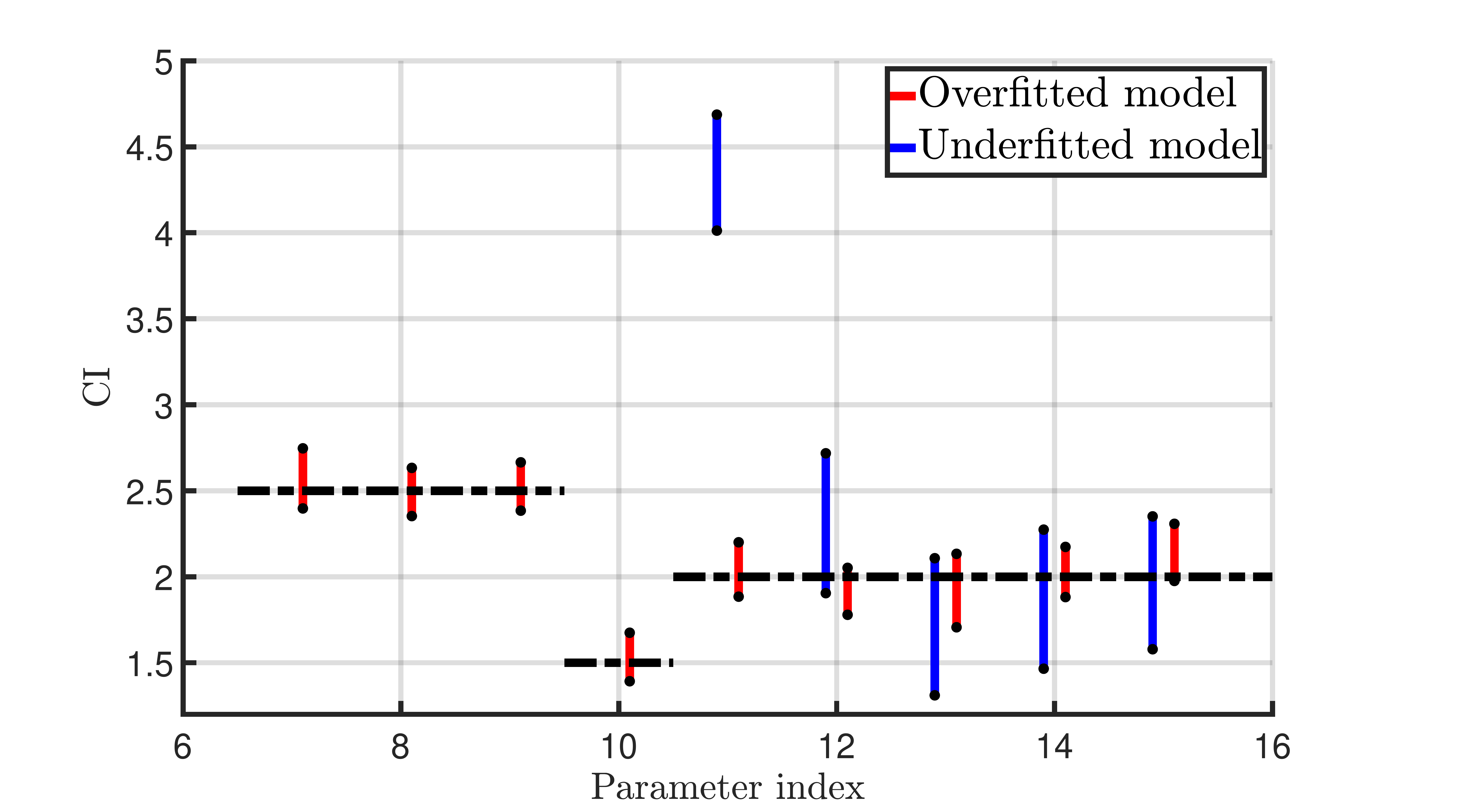}
	%\hspace{-10pt}
	\vspace{-15pt}
	\caption{ Confidence intervals for non-zero coefficients of underfitted model are much larger than that of overfitted model. Besides, confidence interval for one non-zero coefficient of the underfitted model is extremely biased.  (\textit{the dash-dot lines indicate the true value of parameter vector for the corresponding entry})  }
	\label{fig:CI_40}
\end{figure}

\begin{figure} [H]
	\includegraphics[width=10cm,trim = 1.9cm 0.5cm 0.7cm 0.1cm, clip]{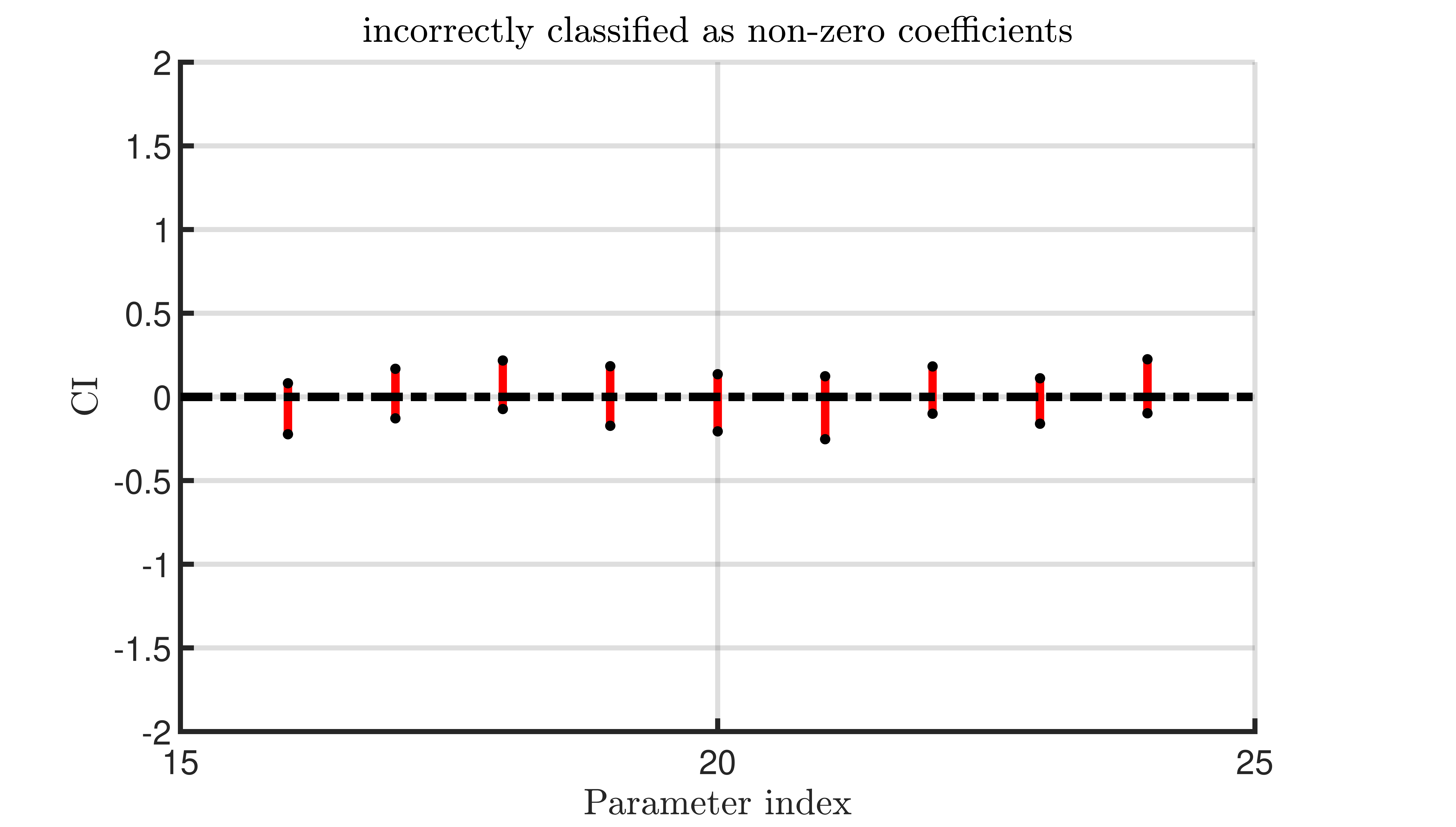}
	\vspace{-20pt}
	\caption{In the overfitted model, we observe that confidence intervals for false positives associated with zero coefficients of the parameter vector cover zero for the given entries. This implies that we can identify the variables associated with confidence intervals covering zero as irrelevant. ( \textit{the dash-dot lines indicate the true value of parameter vector  for the corresponding entry}) }
	\label{fig:CI_400}
\end{figure}

  We observe that confidence intervals for non-zero coefficients of underfitted model are much larger than that of overfitted model, as indicated by \textbf{Fig. \ref{fig:CI_40}}. In particular, confidence interval for one coefficient of underfitted model is very biased. Therefore, we do not only lose information about the relevant variables not chosen by the model selection in underfitted model. Also, the inference results may not reliably reflect confidence intervals for the non-zero coefficients of selected model. Moreover, we observe from \textbf{Fig. \ref{fig:CI_400}} that the confidence intervals for the given coefficients of parameter vector associated with false positives cover $0$. We can identify these coefficients as zero and variables corresponding to them as irrelevant. In this experiment, the confidence intervals associated with only 5 coefficients did not contain zero. Hence, the variable selection can still be improved even when one faces overfitting.

\vspace{-10pt}

\vspace{-10pt}

\subsection{Fusing the Variables Selected by S-Lasso Estimates}
We now study how the variable selection algorithm performs when instead of the variables selected by $\tau$-Lasso estimates, the variables selected by the initial S-Lasso estimates across nodes are aggregated via the majority voting scheme. To do so, we carry out a series of Monte-Carlo simulations of 20 trials where a random realization of outlier is used at each trial. We run the simulations on the synthetic data set described by \textbf{Scenario 3} for low-dimensional regime and the synthetic data set described by \textbf{Scenario 4} for high-dimensional regime.

\textup{Table \ref{tab_fusion_slasso}} shows the result of simulations. We observe that if we aggregate the variables selected by the initial S-Lasso estimates instead of those selected by $\tau$-Lasso estimates, the performance will remain almost the same. This could be explained by the fact that both $\tau$-Lasso and S-Lasso estimators promote sparsity and are robust to gross outliers. We suspect that low Gaussian efficiency has little effect on the performance of sparse recovery.

\begin{table}[H]
\caption{ \textbf{Fusing the Variables Selected by S-Lasso Estimates:} If we use the variables selected by S-Lasso estimates instead of those selected by $\tau$-Lasso, the model selection will remain almost unchanged.}
\label{tab_fusion_slasso}
%\centering
\ra{1.2}
\begin{threeparttable}
\begin{tabular}{@{}lllllclll@{}} \toprule
 & & \multicolumn{3}{c}{TSRD-$\tau$} & &\multicolumn{3}{c}{Fusion / S-Lasso }\\
\cmidrule{3-5} \cmidrule{7-9}
& $\xi_o $ & TP & FP & CER  && TP & FP & CER \\ \cmidrule{2-9}

\multirow{2}{*}{LD}& $0.1$  & 1& 0 & 0 && 1& 0.0011 & 0.0010 \\
& $0.2$
 & 1& 0& 0 && 1 & 0.0017& 0.0015  \\
 \cmidrule{2-9}
  \multirow{2}{*}{HD}& $0.1$  & 1& 0& 0 && 1& 0& 0 \\
& $0.2$
 & 1& 0& 0 && 1 & 0& 0 \\
\bottomrule
\end{tabular}
\end{threeparttable}
\end{table}

\section{Conclusion}

This paper introduced robust and distributed inference procedures for large scale data where data exhibits an underlying low-dimensional structure and is contaminated by outliers. We propose two-stage inference procedures called TSRD-$\tau$ and TSRD-MM. The former employs the class of robust $\tau$-estimators whereas the latter employs that of MM-estimators. In the first stage, active explaining variables are selected by local variable selection employing robust Lasso estimators. The selections from each node are combined by applying a fusion rule at the fusion center or cloud. The selection is broadcast to the computational nodes. In the second stage, actual inferences on the selected variables are performed by using the robust and computationally efficient bootstrap procedures. Confidence intervals are constructed, parameter estimates are found, and standard deviations are quantified. The favorable statistical properties including consistency and robustness of the proposed method were established using analytical methods and verified in simulations. Moreover, the quantitative robustness properties of robust $\tau$-Lasso were established, in particular its finite-sample breakdown point.

Future directions of research include extending the classical asymptotic analysis to high-dimensional asymptotic analysis where $p$ grows with $n$ to infinity.
 It is also an open question how one can establish asymptotic results for local optima. Finally, it would be interesting to devise a rigorous proof for the model consistency of proposed TSRD-$\tau$ and TSRD-MM procedures.

% or


\appendices

\section{Derivation of the Linear Correction Term}

The linear correction may be derived by inverting a block-matrix as follows:
\begin{equation}
    \Big[ \mathbf{I}-\nabla \bbf(\hbth_{b};\tbY) \Big]^{-1}=\begin{bmatrix} 
\mathcal{A}& \boldsymbol{\eta}\\
\boldsymbol{\zeta} & a\end{bmatrix}^{-1},
\end{equation}

where $\mathcal{A}$, $\boldsymbol{\eta}$, $\boldsymbol{\zeta}$ and $a$ are given by
\begin{equation}\label{eq:BLFRB_1}
    \begin{split} 
      \mathcal{A}&=\big(\hat{\bA}_b\big)^{-1} \Bigg[\mathcal{A}_2-\mathcal{A}_1\Bigg],
     \\
     \boldsymbol{\eta}&=\big(\hat{\bA}_b\big)^{-1}\Bigg[\mathcal{\eta}_2 -\mathcal{\eta}_1
     \Bigg],
     \\
       \boldsymbol{\zeta}&=\frac{1}{b \delta_2} \sum_{l=1}^b \rho_0^{'}(\tr_l) \tbx_{[l]}^T,
       \\
       a&=\frac{1}{b \delta_2 }\sum_{l=1}^{b} \rho_0^{'}(\tr_l)\tr_l.
    \end{split}
\end{equation}
$\mathcal{A}_1$, $\mathcal{A}_2$, $\boldsymbol{\eta}_1$ and $\boldsymbol{\eta}_2$ are calculated as follows:
\begin{equation}
\begin{split}
     \mathcal{A}_1&=\frac{1}{b}\sum_{l=1}^{b}\tbx_{[l]}\nabla_{\bb} w_{\tau} \rho_0^{'}(\tr_l), \\
     \mathcal{A}_2&=\frac{1}{b\hs_{b}}\sum_{l=1}^{b}\Big[  \hw_{\tau}\rho_0^{''}(\tr_l) +\rho_1^{''}(\tr_l)\Big] \tbx_{[l]}\tbx_{[l]}^T,  \\
     \mathcal{\eta}_1&=\frac{1}{b}\sum_{l=1}^{b}\nabla_{\sigma} w_{\tau}  \rho_0^{'}(\tr_l)\tbx_{[l]},
     \\
     \mathcal{\eta}_2&=\frac{1}{b\hs_{b}}\sum_{l=1}^{b} \big[ \hw_{\tau}\rho_0^{''}(\tr_l)+\rho_1^{''}(\tr_l) \Big]\tbx_{[l]}\tr_l, 
\end{split}
\end{equation}
where $\nabla_{\bb} w_{\tau} $ and $\nabla_{\sigma} w_{\tau} $ are shorthands for $\partial w_{\tau}(\hbth_b)/ \partial \bb$ and $\partial w_{\tau}(\hbth_b)/ \partial \sigma$, respectively. $\hat{\bA}_b$, $\nabla_{\bb} w_{\tau}$ and $\nabla_{\sigma} w_{\tau}$ are given by
\begin{equation}
 \hat{\bA}_b=\frac{1}{b} \sum_{l=1}^{b} \hw_l^{(i)}\tbx_{[l]} \tbx_{[l]}^T, 
\end{equation}

\begin{equation}\label{eq:derivations_10}
\begin{split}
\nabla_{\bb} w_{\tau}=\frac{\sum_{l=1}^{b}\Big[\rho_1^{''}(\tr_l) \tr_l-\rho_1^{'}(\tr_l)\Big]\tbx_{[l]}^T/\hs_b}{\sum_{l=1}^b \rho_0^{'}(\tr_l)\tr_l},\\
+\frac{\sum_{l=1}^{b}\Big[\rho_0^{''}(\tr_l) \tr_l+\rho_0^{'}(\tr_l)\Big]\tbx_{[l]}^T/\hs_b}{\sum_{l=1}^b \rho_0^{'}(\tr_l)\tr_l}\hw_{\tau}.
\end{split}
\end{equation}

\begin{equation} \label{eq:derivations_115}
    \begin{split}
        \nabla_{\sigma} w_{\tau}=
        \frac{\sum_{l=1}^{b} \Big[\rho_1^{''}(\tr_l)\tr_l-\rho_1^{'}(\tr_l)\Big]\tr_l/\hs_b}{\sum_{l=1}^{b}\rho_0^{'}(\tr_l)\tr_l}\\
        +\frac{\sum_{l=1}^{b}\Big[\rho_0^{''}(\tr_l)\tr_l+\rho_0^{'}(\tr_l)\Big]\tr_l/\hs_b}{\sum_{l=1}^{b}\rho_0^{'}(\tr_l)\tr_l}\hw_{\tau}
    \end{split}
\end{equation}
On the other hand,
\begin{equation}
    \begin{bmatrix} 
\mathcal{A}& \boldsymbol{\eta}\\
\boldsymbol{\zeta} & a\end{bmatrix}^{-1}=\begin{bmatrix} 
\mathbf{M}_{b}& \mathbf{d}_{b}\\
\mathbf{N}_{b} & \mathbf{q}_b\end{bmatrix}.
\end{equation}

\section{Derivation of the One-step Bootstrap Replicates}

The one-step bootstrap replicates $\hbb_{n,b}^{1\star}$ and $\hs_{n,b}^{1\star}$ are calculated as follows:
\begin{equation}\label{eq:BLFRB_2}
    \begin{split}
        \hbb_{n,b}^{1\star}&=\Big(\sum_{l=1}^{b} \omega_l^{\star} w_l^{\star} \tbx_{[l]} \tbx_{[l]}^{T}\Big)^{-1}  \sum_{l=1}^{b} \omega_l^{\star} w_l^{\star} \cy_l \tbx_{[l]},
        \\
        \hs_{n,b}^{1\star}&=\sum_{l=1}^{b} \omega_l^{\star} \check{v}_l^{\star} \big(\cy_l-\tbx_{[l]}^{T}\hbb_b\big),
    \end{split}
\end{equation}
where $v_l^{\star}$, $w_{l}^{\star}$ and $w_{\tau}^{\star}$ are computed as follows:
\begin{equation}\label{eq:BLFRB_3}
    \begin{split}
    v_l^{\star}&=\frac{b}{n}\hat{v}_l,\\
    w_l^{\star}&=\frac{w_{\tau}^{\star}\rho_0^{'}(\tr_l)+\rho_1^{'}(\tr_l )}{\hr_l},\\
    w_{\tau}^{\star}&=
    \frac{\sum_{l=1}^{b} \omega_l^{\star} \Big[2\rho_1(\tr_l)-   \rho_1^{'}(\tr_l) \tr_l \Big]}{\sum_{l=1}^{b} \omega_l^{\star}\rho_0^{'}(\tr_l)\tr_l}.
    \end{split}
\end{equation}

Therefore, the linearly corrected one-step bootstrap replications using $\tau$-estimators are calculated as follows:
\begin{equation}
\label{eq:RSOBT_0}
\begin{split}
    \hbb_{n,b}^{R\star}&=\hbb_{b}+\mathbf{M}_{b}\Big(\hbb_{n,b}^{1\star}-\hbb_{b}\Big)+\mathbf{d}_{b}
    \Big(\hs_{n,b}^{1\star}-\hs_{b}\Big),
    \\
          \mathbf{M}_{b}&=\Big(\mathcal{A}-\boldsymbol{\eta}
          a^{-1}\boldsymbol{\zeta}\Big)^{-1},
      \\
      \mathbf{d}_{b}&=-\mathcal{A}
      ^{-1}\boldsymbol{\eta}\Big(a-\boldsymbol{\zeta}\mathcal{A}^{-1}\boldsymbol{\eta}\Big)^{-1}.
\end{split}
\end{equation}


%\section*{Acknowledgment}


\ifCLASSOPTIONcaptionsoff
  \newpage
\fi



%\bibliographystyle{IEEEtran}

\bibliographystyle{IEEEtran}
\bibliography{refs}

\begin{IEEEbiography}{Emadaldin Mozafari-Majd}
\end{IEEEbiography}

%\begin{IEEEbiographynophoto}{Visa Koivunen}
\begin{IEEEbiography}{Visa Koivunen}
\end{IEEEbiography}

\newpage

This supplemental material contains a section on initial variable screening (Preprocessing) for very high-dimensional settings along with additional simulation results and technical proofs of the theorem discussed in the paper.

\section*{Initial Variable Screening (Preprocessing): Data with Very High-Dimensional Subsets}

In order to reduce the computational burden in settings with very high-dimensional subsets, an initial variable screening procedure called Density Power Divergence-SIS (DPD-SIS) \cite{ghosh2020robust} is employed to further reduce the model complexity, the number of variables to an order of sample size $n$. In distributed and parallel architecture, we use the variable screening only in very high-dimensional settings where $p$ is much larger than $b$. Basically, the DPD-SIS extends the Sure Independence Screening (SIS) to address robustness in the presence of outliers. The robust screening procedure assigns a certain score to each predictor and then predictors are ranked in descending order based on the calculated score. In order to compute the score, the marginal estimate of each regression coefficient is obtained via a minimum DPD estimator and then its absolute value determines the score associated with each predictor. At each node, the DPD-SIS variable screening procedure is utilized to discard a certain number of irrelevant predictors and then distinct subsets with reduced dimensionality is passed down to next step for further processing, that is, $ \cbX \in \bR^{b \times p} \rightarrow \bar{\mathbf{X}} \in \bR^{b \times q}$ with $(q \ll p)$ an order of sample size. A reasonable choice for $q$ is the number of observations within subsets of data, that is, $q=b$. Note that variable screening procedure is dispensable in low-dimensional models, i.e. $q=p$. Subdividing a high-dimensional data into smaller subsets may be allowed as long as the batch size $b$ satisfies the requirements for sparse recovery and statistical correctness of bootstrap computations as specified in \textbf{Section V.E}, \textit{Choice of Batch Size}, within the main body of the paper. The algorithmic details of DPD-SIS can be found in \cite{ghosh2020robust}. 

 The DPD-SIS is initially applied to the distinct subsets of data and top $b$ predictors, the exact order of sample size, within each subset of data are kept based on their score and the remaining predictors are discarded. We then form a set of predictors appearing within at least half of data subsets. We now place these set of predictors on top of the ranking within each of $s$ data subsets, keep the top $b$ predictors and reject the remaining ones. The model selection proceeds with the data set of reduced dimensionality.

\section*{Additional Simulation Results}
\subsection{High-dimensional: Model Selection and Inference}

In this part, we study the performance of the proposed methods, TSRD-$\tau$ and TSRD-MM, for a large-scale high-dimensional data set ($p>b$) in terms of model selection and robustness of confidence intervals to outliers. We run the simulations on the synthtetic dataset described by \textbf{Scenario 1}. In the current high-dimensional setting, an initial variable screening procedure is employed to reduce the dimensionality prior to model selection. The DPD-SIS is initially applied to the distinct subsets of data and top $b$ covariates, the exact order of sample size, within each subset of data are kept based on their score and the remaining covariates are discarded. The model selection proceeds with the data set of reduced dimensionality.
%\vspace{-15pt}
\begin{table}
\caption{ \textbf{Model Selection in High-Dimensional:} The Proposed Two-Stage Robust Inference Methods Achieve a Perfect Recovery of True Sparse Basis ($\text{TP=1}$) with Small Number of False Positives  under Zero Contamination to Moderate Contamination. In Contrast, the Two-Stage Lasso-LS Completely Fails at Recovering the Sparse Basis ($\text{TP}=0$) for all Scenarios Except for Outlier-Free.}
\label{tabhigh_dim}
%\centering
\ra{1.2}
\begin{threeparttable}
\begin{tabular}{@{}llllclllclll@{}} \toprule
 & \multicolumn{3}{c}{TSRD-$\tau$} & &\multicolumn{3}{c}{TSRD-MM}  & &\multicolumn{3}{c}{TSLL}\\
\cmidrule{2-4} \cmidrule{6-8} \cmidrule{10-12}
$\xi_o $ & TP & FP & CER  && TP & FP & CER && TP & FP & CER \\ \midrule

$0$
& 1 & 0.0004 & 0.0004 && 1 & 0.0013 & 0.0013 && 1 & 0 & 0\\
$0.1$  & 1& 0.0003& 0.0003 && 1& 0.0009& 0.0009 && 0 & 0 & 0.0013\\
$0.2$
 & 1& 0.0002& 0.0002 && 1 & 0.0002& 0.0002 && 0 & 0 & 0.0013\\
\bottomrule
\end{tabular}
\begin{tablenotes}
    \item[1] Although no truly active variables are selected by TSLL in the model selection stage, one might be mislead by low CER. This can be attributed to fact that the number of truly active variables $k_s$ are insignificant compared to $p$.
\end{tablenotes}
\end{threeparttable}
\end{table}
The proposed two-stage robust inference algorithms are compared to their two-stage non-robust counterpart employing Lasso in the first stage and BLB based on least square estimator in the second stage. For the sake of brevity, the non-robust two-stage inference method is regarded as \textit{two-stage Lasso-LS} (TSLL). Before using Lasso, variable screening is carried out by the DPD-SIS to ensure the data with the reduced dimensionality contains relevant variables. In regard to model selection, the performance is quantified by using confusion matrix and CER as represented in \textbf{Table \ref{tabhigh_dim}}. The model selection algorithms in the stage 1 of TSRD-$\tau$ and TSRD-MM could perfectly identify all true non-zero parameters for different proportions of outliers at the highly underdetermined setting, $p/b=33.33$. In contrast, the non-robust TSLL algorithm fails completely at identifying the sparse basis of the parameter vector except for outlier-free scenario. 

%\vspace{-15pt}
\begin{figure} 
	%\centering
    \hspace{-15pt}
	\includegraphics[width=10.5cm]{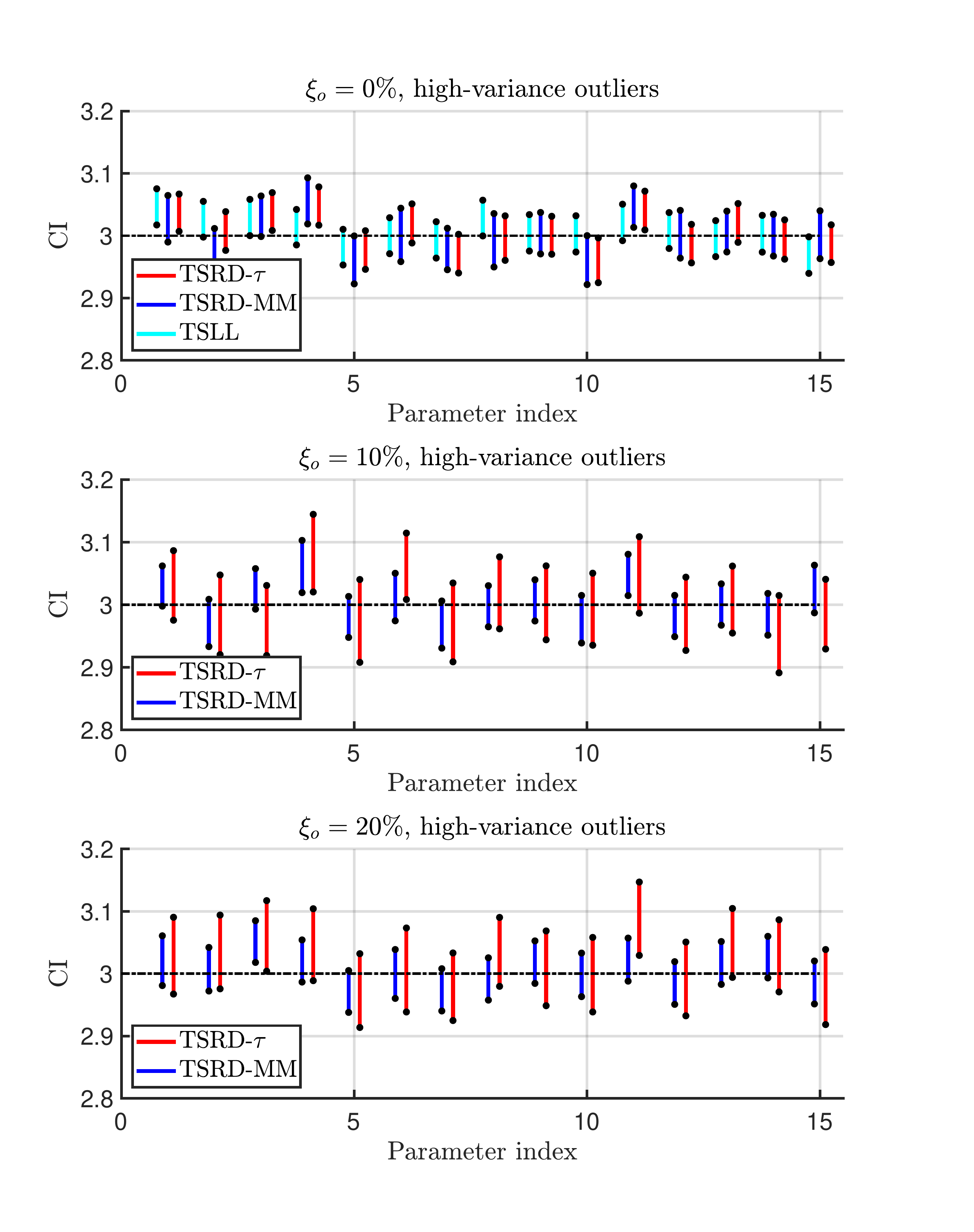}
	\vspace{-30pt}
	\caption{ When there are no outlying observations, the confidence intervals based on bootstraps methods in TSRD-$\tau$ and TSRD-MM provide reliable estimates of the CIs based on bootstrap percentiles of least-square estimator. The confidence intervals produced by TSRD-T and TSRD-MM methods exhibit robustness to outliers and their lengths are slightly affected by an increase in the proportion of outliers. Thus, reliable parameter estimates and confidence intervals are obtained even in the presence of outliers (\textit{the dash-dot lines indicate the true value of non-zero entries of parameter vector}). }
	\label{fig:CI_High}
	%\captionsetup{skip=0pt}
\end{figure}

The selected variables from the first stage are used to construct confidence intervals based on the bootstrap methods RSOB-$\tau$ and BLFRB. The confidence intervals constructed for the first $15$ selected variables are shown in \textbf{Fig.  \ref{fig:CI_High}}. The CIs formed by robust inference methods at the stage 2 of TSRD-$\tau$ and TSRD-MM remain resistant to contamination and length of CIs are  slightly inflated with an increase in the proportion of outliers. In regard to outlier-free scenario, the robust bootstrap methods provide reliable estimates of the CIs constructed by bootstrap percentiles of least-square estimator. Hence, it can be concluded that the proposed two-stage inference methods, TSRD-$\tau$ and TSRD-MM, can be used to perform robust statistical inference for large-scale high-dimensional data sets. Note that no comparison is made to two-stage Lasso-LS in the presence of outlying observations due to zero true positive rate, i.e., no element of sparse basis was identified.

\subsection{Effect of Substituting $\tau$-Lasso with non-regularized $\tau$- and MM-estimators on Variable Selection}

We study the effect of substituting the $\tau$-Lasso estimator with $\tau$-estimator and MM-estimator under the settings described by \textbf{Scenario 3} and $10 \%$ outlier ratio. We conduct the above experiment with $20$ trials for which a random realization of outlier is used at each trial. The $\tau$- and MM-estimators were initialized with S-Lasso estimates of the parameter vector $\bb$. We observe in our simulations that both $\tau$- and MM-estimators fail in recovering the correct support and all irrelevant variables are selected within the estimated support In contrast, TSRD-$\tau$ employing the $\tau$-Lasso estimator succeeds in recovering the correct support, indicated by the following contingency table.
\begin{table}[H]
\caption{ \textbf{Model Selection:} The Proposed TSRD-$\tau$ Method Achieves a Perfect Recovery of True Sparse Basis ($\text{TP=1}$) with no False Positives ($\text{FP}=0$). In Contrast, all irrelevant variables are selected within the estimated support when one replaces the $\tau$-Lasso estimator with non-regularized estimators}
\label{tab:tab2}
\ra{1.2}
\begin{threeparttable}
\begin{tabular}{@{}llllclllclll@{}} \toprule
 & \multicolumn{3}{c}{TSRD-$\tau$} & &\multicolumn{3}{c}{$\tau$}  & &\multicolumn{3}{c}{MM}\\
\cmidrule{2-4} \cmidrule{6-8} \cmidrule{10-12}
$\xi_o $ & TP & FP & CER  && TP & FP & CER && TP & FP & CER \\ \midrule
$0.1$  & 1& 0& 0 && 1& 1& 0.9 && 1 & 1 & 0.9\\
 \\
\bottomrule
\end{tabular}
\end{threeparttable}
\end{table}
Note that we obtained the results demonstrated in the above table by averaging over 20 trials.

\subsection{Effect of Initial Estimate on Variable Selection}

In order to explore the influence of initial estimates on the variable selection procedure, we perform a series of simulations with high-dimensional and low-dimensional data. We run the $\tau$-Lasso estimator when randomly initialized and then when initialized with non-regularized $\tau$-estimates (applicable only to low-dimensional regime) and compare the results of model selection with recommended procedure where the $\tau$-Lasso estimator is using S-Lasso estimates as the initial point. In both high-dimensional and low-dimensional data, we carry out a Monte-Carlo study of 20 trials where a random realization of outlier is used at each trial. In case of $\tau$-Lasso estimator with random initialization, we initiate the algorithm with a randomly distributed multivariate Gaussian $\mathcal{N}(1000\times \mathbf{1}_{p+1},(250)^2 \times \textbf{I}_{p+1})$ for each trial and batch of data, chosen to be far from the true coefficient $\bb_0$. We run the simulations on the synthetic data set described by \textbf{Scenario 3} for low-dimensional regime and the synthetic data set described by \textbf{Scenario 4} for high-dimensional regime.

As shown in \textup{Tables \ref{tab_initial_estld}-\ref{tab_initial_esthd}}, the model selection algorithm using $\tau$-Lasso achieves the exact support recovery regardless of initialization across all trials for the low-dimensional and high-dimensional regimes. Although the model selection procedure succeeds in perfectly recovering the true support when $\tau$-Lasso estimator initialized randomly. In fact, the $\tau$-Lasso optimization problem is solved via alternating minimization where the sub-problems are non-convex themselves. Recent results show that many well-known nonconvex optimization problems possess a well behaved landscape where all second-order stationary points are global minima \cite{li2019non} and \cite{chi2019nonconvex}. We conjecture that the variable succeeds in recovering the true support when the $\tau$-Lasso optimization problem is initialized randomly due to potential nice landscape of optimization problem in conjunction with collaborative nature of fusion procedure. We suspect this result may not entirely generalize to all scenarios and this topic requires further study from the optimization perspective. 

%\vspace{-15pt}
\begin{table}
\caption{ \textbf{Influence of Initial Estimate on Model Selection in Low-Dimensional Data:} Regardless of How the $\tau$-Lasso Estimator is Initialized, the Model Selection Method Succeeds in Perfectly Recovering the True Sparse Basis ($\text{TP=1}$) with no False Positives ($\text{FP}=0$) under Contamination.}
\label{tab_initial_estld}
%\centering
\ra{1.2}
\begin{threeparttable}
\begin{tabular}{@{}llllclllclll@{}} \toprule
 & \multicolumn{3}{c}{TSRD-$\tau$} & &\multicolumn{3}{c}{random init. } & &\multicolumn{3}{c}{init. by $\tau$-estimates}\\
\cmidrule{2-4} \cmidrule{6-8} \cmidrule{10-12}
$\xi_o $ & TP & FP & CER  && TP & FP & CER && TP & FP & CER \\ \midrule

$0.1$  & 1& 0& 0 && 1& 0& 0 && 1 & 0 & 0\\
$0.2$
 & 1& 0& 0 && 1 & 0& 0 && 1 & 0 & 0\\
 \\
\bottomrule
\end{tabular}
\end{threeparttable}
\end{table}

%\vspace{-15pt}
\begin{table}
\centering
\caption{ \textbf{Influence of Initial Estimate on Model Selection in High-Dimensional Data:} Regardless of How the $\tau$-Lasso Estimator is Initialized, the Model Selection Method Succeeds in Perfectly Recovering the True Sparse Basis ($\text{TP=1}$) with no False Positives ($\text{FP}=0$) under Contamination.}
\label{tab_initial_esthd}
%\centering
\ra{1.2}
\begin{threeparttable}
\begin{tabular}{@{}llllclll@{}} \toprule
 & \multicolumn{3}{c}{TSRD-$\tau$} & &\multicolumn{3}{c}{random init. }\\
\cmidrule{2-4} \cmidrule{6-8}
$\xi_o $ & TP & FP & CER  && TP & FP & CER\\ \midrule

$0.1$  & 1& 0& 0 && 1& 0& 0\\
$0.2$
 & 1& 0& 0 && 1 & 0& 0 \\
 \\
\bottomrule
\end{tabular}
\end{threeparttable}
\end{table}

\subsection{The Effect of Sample Size, Dimensionality , and Number of Subsamples on Computational Complexity of RSOB-$\tau$}

\begin{figure} 
	\centering
	\includegraphics[width=7cm, trim = 0.2cm 0cm 0.5cm 0.1cm, clip]{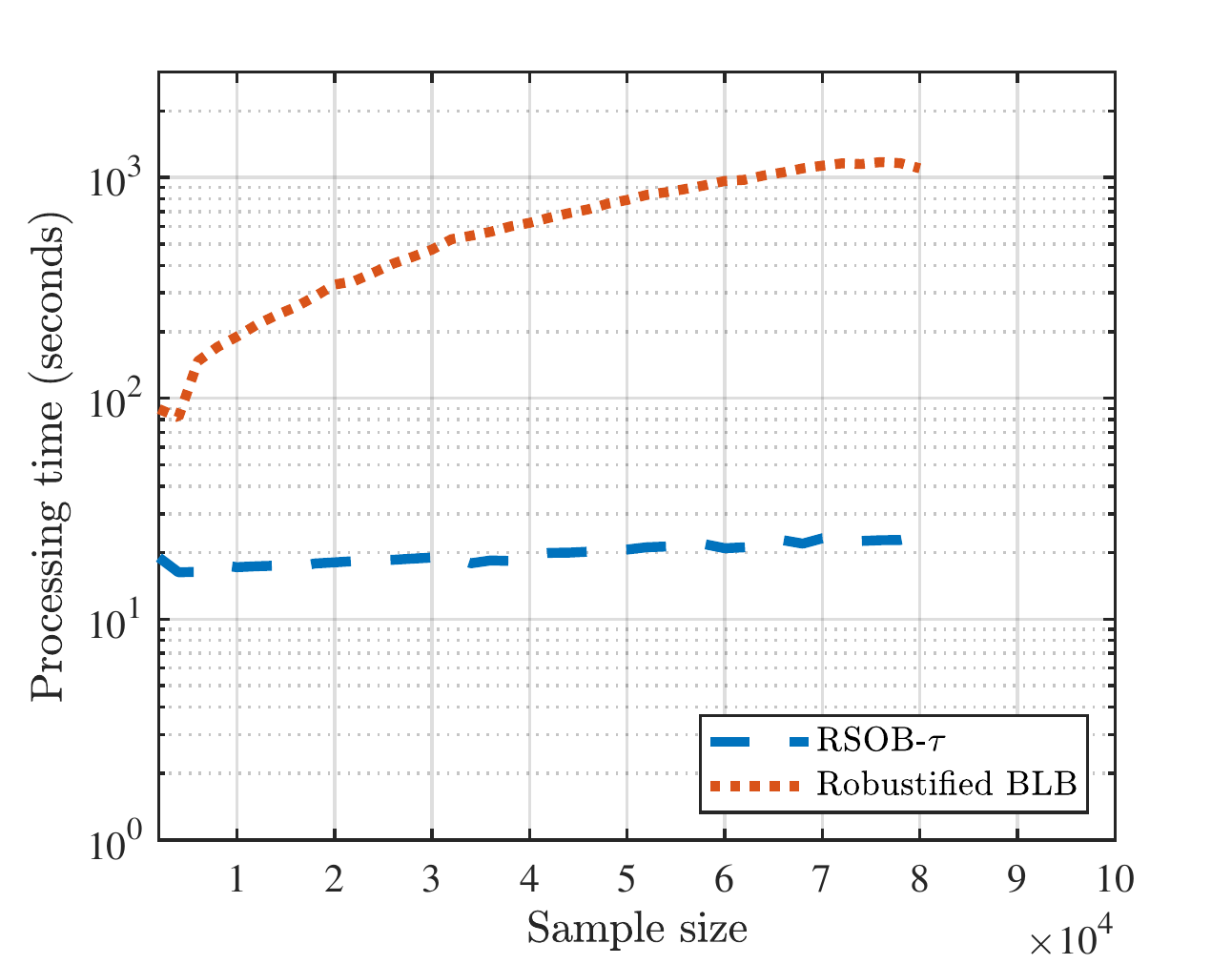}
	\vspace{-5pt}
	\caption{The processing time of RSOB-$\tau$ method employing linearly corrected one-step $\tau$-estimator scales much better with sample size than the robustified BLB method.}
	\label{fig:time_n}
\end{figure}

\begin{figure} [H]
	\centering
	\includegraphics[width=7cm, trim = 0.2cm 0cm 0.5cm 0.1cm, clip]{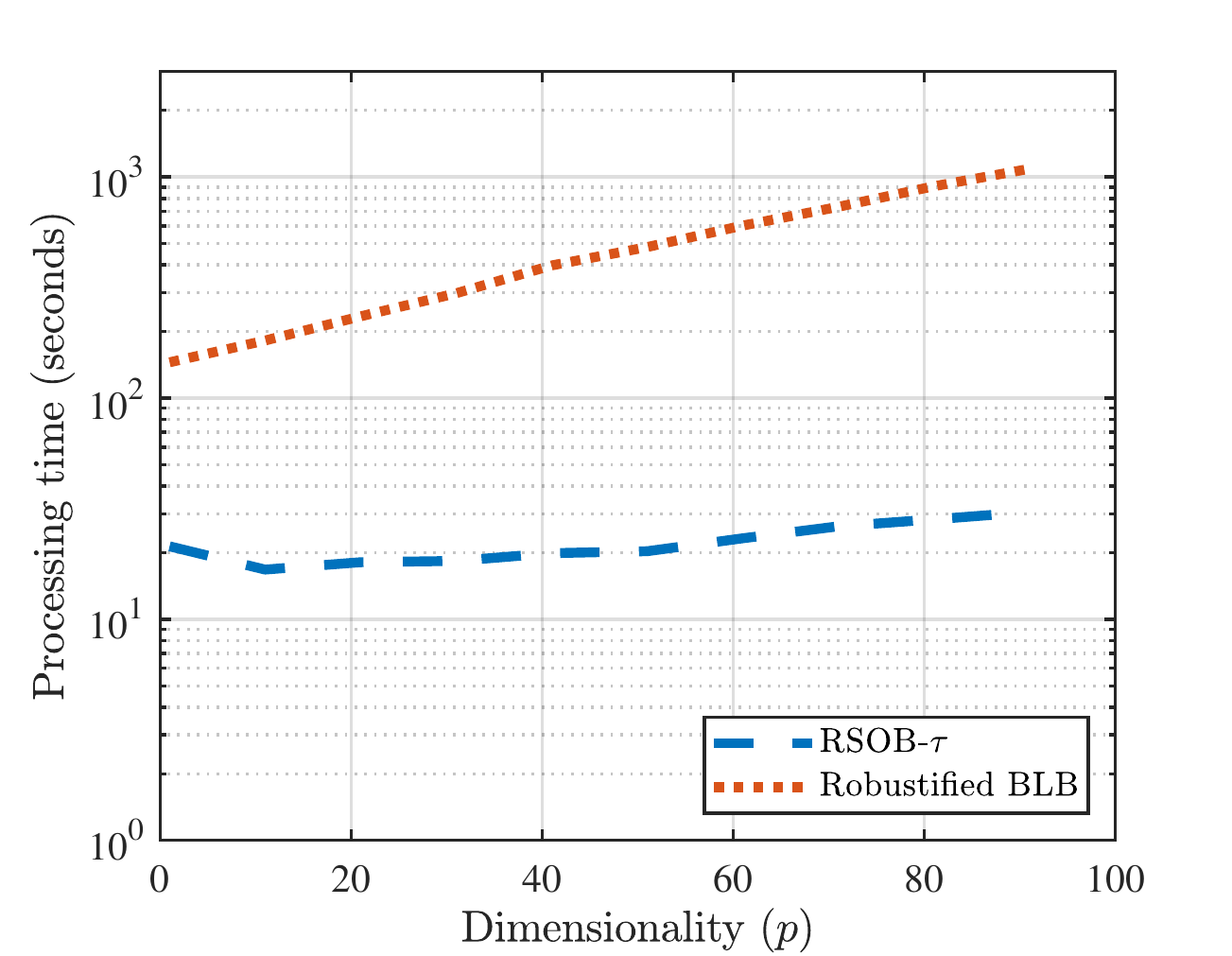}
	\vspace{-5pt}
	\caption{The processing time of RSOB-$\tau$ method employing linearly corrected one-step $\tau$-estimator scales much better with dimensionality than the robustified BLB method.}
	\label{fig:time_p}
\end{figure}

\begin{figure} [H]
	\centering
	\includegraphics[width=7cm, trim = 0.2cm 0cm 0.5cm 0.1cm, clip]{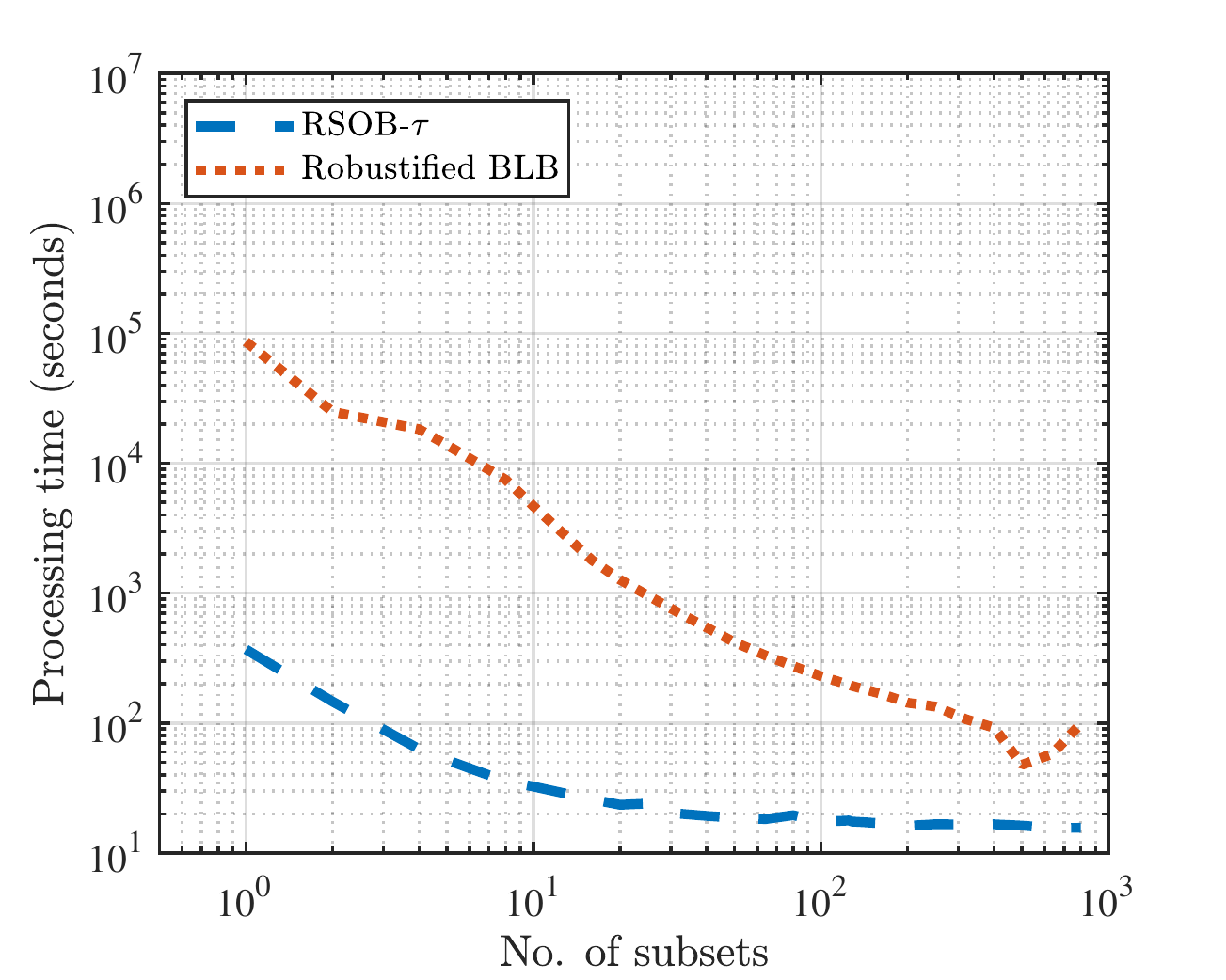}
	\vspace{-5pt}
	\caption{The above plot indicates if one excessively increases the number of data partitions, small computational gains are made  by using RSOB-$\tau$ than using the robustified BLB.}
	\label{fig:time_s}
\end{figure}

We conduct a number of experiments to examine how the computational complexity of RSOB-$\tau$ scales with sample size, dimensionality and the number of subsamples. To do so, we plot the processing time against one of the above parameters while keeping the remaining parameters at their defaults. This process continues until all parameters have been allowed to vary for a range of values. We run the simulations on the data set described by \textbf{Scenario 3} where $10 \%$ of observations are contaminated by outliers. We then compare them to the result of simulations obtained using robust realization of BLB method based on $\tau$-estimator and report the corresponding results in \textbf{Fig. \ref{fig:time_n}-\ref{fig:time_s}}. We conduct the experiment on a single node of a high-performance computing cluster (Triton) where 22 computing cores and $25$ GB of memory were used. Note that we report the processing time of RSOB-$\tau$ versus that of robustified BLB where a distinct subsample is allocated to each computing core. The number of bootstrap samples is fixed at $B=300$.

 \textbf{Fig. \ref{fig:time_n}} shows that as we increase the sample size $n$, the processing time associated with RSOB-$\tau$ grows at much slower rate than the processing time associated with the BLB method based on $\tau$-estimator, thereby achieving higher computational efficiency for larger sample size. In \textbf{Fig. \ref{fig:time_p}}, we obtain similar results when plotting the processing time against dimensionality $p$. That is, RSOB-$\tau$ achieves higher computational efficiency for larger dimensions $p$.

\vspace{-10pt}

\textbf{Fig. \ref{fig:time_s}} shows that as we increase the number of subsets (data partitions), the processing time associated with both RSOB-$\tau$ and the robustified BLB decreases and simultaneously the gap between the two curves shrinks. Therefore, there would little benefit in excessively increasing the number of subsets when considering the processing time. We assume that each subsample is processed by allocating one computing core to each subsample. Setting $s$ to very large values leads to wasted computation as we are occupying so many resources while a small portion of each resource is needed. One could use $\lfloor n/n^{0.9}\rfloor<s<\lfloor{n/\max(c_s k_s\log(p), n^{0.6})}\rfloor$ as a crude approximation and choose $ b = n/s$ so that it uses the computational and storage capabilities of the computational nodes efficiently. As it would be difficult to allocate computing cores for large numbers, we record the processing time associated with each subsample and their maximum is considered to be the processing time for better interpretability.

\section*{Proof of Theorem 1} The finite-sample breakdown point of $\tau$-Lasso estimator can be derived in two separate stages: first stage is dedicated to show the boundedness of finite-sample breakdown point from above and second stage is dedicated to show the boundedness of finite-sample breakdown point from below. Once the boundedness from above and below is established, the above theorem is proved immediately. Before proceeding any further, we extend the lemma 5.1 \cite{maronna2019robust} for M-scale estimators to $\tau$-scale estimators as follows: \\

\textbf{Lemma 1}: \textit{Consider any sequence of samples $\Big( \hat{\bY}^{(k)}=\big(\hat{\by}^{(k)},\hat{\bX}^{(k)}\big) \in \bR^{b \times (p+1) }\Big)_{k \in \mathbb{N}}=\Big(\hat{\bY}^{(1)},\hat{\bY}^{(2)}, \hat{\bY}^{(3)}, \cdots \Big)$ and  corresponding residual vector $\mathbf{r}^{(k)}=\hat{\by}^{(k)}-\hat{\bX}^{(k)}\hbb^{(k)}$ for $\hat{\bY}^{(k)}$. Suppose $r^{(k)}_l$ denotes the residual for  $\big(\hat{y}^{(k)}_{l},(\hat{\bx}_{[l]}^{(k)})^T \big)$ a row in $\hat{\bY}^{(k)}$,  Then \\
\begin{enumerate}
    \item Let $C=\{l:|r_{l}^{(k)}| \rightarrow \infty \}$, if $\#(C)>b \delta$, then $\hs_{\tau}(\mathbf{r}^{(k)}) \rightarrow \infty$ as $k \rightarrow \infty$.
    \item Let $D=\{l:|r_{l}^{(k)}|\mbox{ is bounded } \}$, if $\#(D)>b-b \delta$, then $\hs_{\tau}(\mathbf{r}^{(k)})$ is bounded.
\end{enumerate}}
where $\rightarrow \infty$ denotes the left-hand side of arrow tends to infinity. The part $1$ of the lemma implies that if the number of unbounded entries of $\br^{(k)}$ exceeds $b \delta$, then $\tau$-scale estimate of $\br^{(k)}$ goes to infinity. On the other hand, if the number of bounded entries of $\br^{(k)}$ exceeds $b-b \delta$, then $\tau$-scale estimate of $\br^{(k)}$ remains bounded. To prove the above lemma, we use the results from lemma 3.2 in \cite{yohai1986high}, the $\tau$-scale is bounded from above and below as follows:

\begin{equation}
\label{eq:tau_M_relation}
  \bar{c}_i \hs_{\text{M}}(\mathbf{u}) \leq \hs_{\tau}(\mathbf{u})\leq \sqrt{\sup_{t \in \bR} \rho_1(t)} \hs_{\text{M}}(\mathbf{u})  \quad \forall \mathbf{u} \in \bR^b,
\end{equation}
where $\bar{c}_i$ is a positive constant, $\hs_{\text{M}}$ denotes the M-scale estimate and $\hs_{\tau}$ denotes the $\tau$-scale estimate. Now, we can take advantage of the inequality associating M-scale estimate with $\tau$-scale estimate. We extend the lemma 5.1 in p.184 of \cite{maronna2019robust} for M-scale estimators and derive similar results for $\tau$-scale estimators. According to part 1 of lemma 5.1  in \cite{maronna2019robust},

\begin{equation}
\begin{split}
        \mbox{if }\#(C)>b \delta \Rightarrow  \hs_{\text{M}}(\mathbf{r}^{(k)}) \rightarrow \infty,\\
    \bar{c}_i \times \infty  \leq \hs_{\tau}(\mathbf{r}^{(k)})\leq \sqrt{\sup_{t \in \bR} \rho_1(t)} \times \infty, 
\end{split}
\end{equation}
which implies both upper and lower bound of $\tau$-scale estimate goes to infinity. This proves part 1 of the lemma,  if $\#(C)>b\delta$ then $\hs_{\tau}(\mathbf{r}^{(k)}) \rightarrow \infty$.
To prove the second part of lemma, we know from part 2 of lemma 5.1 \cite{maronna2019robust},

\begin{equation}
\begin{split}
        \#(D)>b-b\delta \Rightarrow  \hs_{\text{M}}(\mathbf{r}^{(k)}) \mbox{ is bounded},\\
     \bar{c}_i \times \mbox{bounded}  \leq \hs_{\tau}(\mathbf{r}^{(k)})\leq  \sqrt{\sup_{t \in \bR} \rho_1(t)} \times \mbox{bounded},
\end{split}
\end{equation}
which implies both lower and upper bounds of $\tau$-scale estimates are bounded. This proves the second part of the lemma, if $\#(D)>b-b\delta$, then $\hs_{\tau}(\mathbf{r}^{(k)})$ is bounded. Now, we need to prove in two separate stages that the finite-sample breakdown point of the $\tau$-Lasso estimator is bounded from the above and the below.
\subsection{Bounded From Below}
In order to establish the boundedness of $\hbb$ from below, we need to show that the sequence of $\tau$-scale estimates $\big( \hbb^{(k)} \in \bR^p \big)_{k \in \mathbb{N}}$ is bounded for any arbitrary sequence of contaminated samples $\big(\hat{\bY}_m^{(k)}\big)_{k \in \mathbb{N}}$ with $m \leq m(\delta)$. The method of proof by contradiction can be used to establish the boundedness from below. That is, it is assumed the sequence $\big( \hbb^{(k)} \big)_{k \in \mathbb{N}}$ is unbounded and then shown that $\hbb^{(k)}$ violates the optimality condition where it is assumed to attain the minimum of the $\tau$-Lasso objective function. Thus, $ \hbb^{(k)} $ should be bounded to be a minimizer of the objective function.

Suppose $\tilde{\bb} \in \bR^p$ has a bounded $\ell_1$-norm such that $\|\tilde{\bb}\|_{\ell_1}=K_b < \infty$. For the uncontaminated observations $(\check{y}_l,\check{\bx}_{[l]}^T)$ within the contaminated sample $\hat{\bY}_m^{(k)}$, the corresponding residuals $|r_{l}^{(k)}\big(\tilde{\bb},\hat{\bY}_m^{(k)}\big)|=|\check{y}_l-\cbx_{[l]}^T\tilde{\bb}|< \infty$ are bounded based on the triangle inequality. Without loss of generality, it is assumed $b\delta=b \min(\delta,1-\delta)$. Since the inequality $m \leq m(\delta) \leq b\delta$ holds based on the theorem assumption, it can be shown the number of bounded residuals $\#(D)\geq b-m \geq b-b\delta$. Therefore, we can conclude $\hs_{\tau}\Big(\mathbf{r}^{(k)}\big(\tilde{\bb},\hat{\bY}_m^{(k)}\big)\Big) $ can be large but still bounded based on the above lemma,

\begin{equation}
    \sup_{k \in \mathbb{N}} \hs_{\tau}\Big(\mathbf{r}^{(k)}\big(\tilde{\bb},\hat{\bY}_m^{(k)}\big)\Big) < \infty.
\end{equation}

Now, let the sequence $\big(\| \hbb^{(k)}\|_{\ell_1}\big)_{k \in \mathbb{N}}$ be unbounded. An unbounded sequence does not converge, i.e. sequences that contain arbitrarily large numbers. Hence, there exists a sequence index $k_0$ such that $\|\hbb^{(k_0)}\|_{\ell_1} > K_b+\frac{1}{\lambda} \sup_{k \in \mathbb{N}} \hs_{\tau}^2\Big(\mathbf{r}^{(k)}\big(\tilde{\bb},\hat{\bY}_m^{(k)}\big)\Big)$ and as a result we can say for every $k^{'} \geq k_0$,

\begin{equation}
\label{eq:TAU-Lasso}
\begin{split}
    \overbrace{\hs_{\tau}^2\Big(\mathbf{r}^{(k^{'})}\big(\hbb^{(k^{'})},\hat{\bY}_m^{(k^{'})}\big)\Big)+\lambda \| \hbb^{(k^{'})}\|_{\ell_1}}^{\mathcal{L}\big(\hbb^{(k^{'})},\hat{\bY}_m^{(k^{'})}\big)} \\> \hs_{\tau}^2\Big(\mathbf{r}^{(k^{'})}\big(\hbb^{(k^{'})},\hat{\bY}_m^{(k^{'})}\big)\Big)\\+\lambda\Bigg(K_b+\frac{1}{\lambda} \sup_{k \in \mathbb{N}} \hs_{\tau}^2\Big(\mathbf{r}^{(k)}\big(\tilde{\bb},\hat{\bY}_m^{(k)}\big)\Big)\Bigg) 
    \\
    \geq \hs_{\tau}^2 \Big(\mathbf{r}^{(k^{'})}\big(\tilde{\bb},\hat{\bY}_m^{(k^{'})}\big)\Big) +\lambda K_b=\mathcal{L}\big(\tilde{\bb},\hat{\bY}_m^{(k^{'})}\big) \Rightarrow 
    \\
    \mathcal{L}\big(\hbb^{(k^{'})},\hat{\bY}_m^{(k^{'})}\big) \geq \mathcal{L}\big(\tilde{\bb},\hat{\bY}_m^{(k^{'})}\big)
\end{split}
\end{equation}
where $\lambda$ determines the amount of regularization imposed by the $\ell_1$-norm and $\mathcal{L}(\cdot)$ is $\tau$-Lasso objective function as given in equation (\ref{eq:TAU-Lasso}). The above result contradicts the fact that $\hbb^{(k^{'})}$ is the minimizer of the objective function, as the loss function is larger for $\hbb^{(k^{'})}$. Thus, $\hbb^{(k)}$ should be bounded for $m \leq m(\delta)$ and the boundedness from below is proved.\\

\subsection{Bounded From Above}

The boundedness from above can be established by showing that the estimator breaks down for $m > b\delta$. In order to prove such a property, proof by contradiction is used in this work. That is, it is assumed that the sequence of estimates $\big(\hbb^{(k)}\big)_{k\in \mathbb{N}}$ minimizing the $\tau$-Lasso objective function over the contaminated sample $\hat{\bY}_m^{(k)}$ is bounded. and then shown the $\tau$-Lasso objective function evaluated at $\hbb^{(k)}$ achieves larger value than the $\tau$-Lasso objective function evaluated at a given unbounded $\tilde{\bb}^{(k)}$. This contradicts the assumption that $\hbb^{(k)}$ is the minimum of $\tau$-Lasso objective function. Thus, it can be concluded $\hbb^{(k)}$ must be unbounded for $m> b \delta$ establishing the boundedness from above.

As the $\tau$-Lasso objective function is comprised of three primary components, its evaluation is carried out in three separate steps as follows:
\begin{enumerate}
    \item \textit{M-scale of residuals, component 1}
    \item $\frac{1}{b} \sum_{l=1}^b \rho_1 \Bigg( \frac{\hat{r}_{l}^{(k)}}{\hs_{\text{M}}\big(\hat{\br}^{(k)}\big)}\Bigg)$, \textit{component 2}
    \item $\lambda \| \hbb^{(k)}\|_{\ell_1}$, \textit{component 3}
\end{enumerate}

Once these components are evaluated, the $\tau$-Lasso objective function is formed by a simple addition and multiplication operation over these components. In order to proceed with the proof, the $\tau$-Lasso objective function is evaluated for the bounded sequences $\hbb^{(k)}$ and unbounded sequences $\tilde{\bb}^{(k)}$ and then compared with each other to arrive at a contradiction. \\

\subsubsection{Evaluation of \texorpdfstring{$\tau$}{tau}-Lasso Objective Function for \texorpdfstring{$\hat{\boldsymbol{\beta}}^{(k)}$}{beta}}

 \paragraph{Component 1} Suppose the set $C \subset \{ 1,\cdots,b\}$ denotes the indices of the observations within the original contaminated-free sample $\check{\bY}$ replaced by outliers to construct the contaminated sample $\hat{\bY}_m^{(k)}$ with $m= \#(C)$. To simplify the proof without loss of the generality, we choose an arbitrary $\bx_0 \in \bR^{p}$ with unit $\ell_2$-norm, $\|\bx_0\|_{\ell_2}=1$. Now, we construct a contaminated sequence of samples with $m$  outliers given by
\begin{equation}
    \big(\hat{y}_{l}^{(k)},\hat{\bx}_{[l]}^{(k)}\big)=\begin{cases}
  \big( k^{\nu+1},\bx_0 k \big) & l \in C\\    
  \big( \check{y}_l,\cbx_{[l]} \big) & l \notin C
\end{cases},
\end{equation}
where $0 < \nu \leq 1$ and $\big( k^{\nu+1},\bx_0 k \big)$ are chosen to account for outlying observations. The sequence for outlying observations diverges as $k$ goes to infinity.

First, we assume that $\hbb^{(k)}$ is bounded in norm and consequently, we have $|r_{l}^{(k)}\big( \hbb^{k},\hat{\bY}_m^{(k)} \big)|=|\hat{y}_l-(\hat{\bx}_{[l]}^{(k)})^T \hbb^{(k)}|< \infty$ is bounded for $l \notin C$ and $k \in \mathbb{N}$. $\hat{r}_l^{(k)}$ is a shorthad for $r_{l}^{(k)}\big( \hbb^{k},\hat{\bY}_m^{(k)} \big)$. On the other hand, the residuals corresponding to contaminated observations, $l \in C$ are lower bounded by

\begin{equation}
\begin{split}
        |\hat{r}_l^{(k)}|=|k^{\nu+1}-k\bx_0^T\hbb^{(k)}|=k|k^{\nu}-\bx_0^T\hbb^{(k)}| \mbox{ and } \bx_0^T\hbb^{(k)} \\\leq \|\bx_0\|_{\ell_1} \|\hbb^{(k)}\|_{\ell_1} \Rightarrow
    |k^{\nu}-\bx_0^T\hbb^{(k)}|\geq |k^{\nu}-\|\bx_0\|_{\ell_1}  \|\hbb^{(k)}\|_{\ell_1}| \\ \Rightarrow
    |\hat{r}_l^{(k)}| \geq k |k^{\nu}-\|\bx_0\|_{\ell_1}  \|\hbb^{(k)}\|_{\ell_1}|
\end{split}
\end{equation}

In addition, the right-hand side of inequality goes to infinity as $k$ approaches infinity which implies the residuals $\hat{r}_l^{(k)}$ go to infinity for $l \in C$ as well. Based on the above lemma and lemma 5.1 in p.184 of \cite{maronna2019robust} , we conclude that both $\hs_{\tau}\big(\hat{\br}^{(k)}\big)$ and $\hs_{\text{M}}\big(\hat{\br}^{(k)}\big)$ go to infinity for $\#(C) >b \delta$. As a result, we can decompose the M-estimation of scale equation as follows:
\begin{equation}
     \sum_{l \notin C}\rho_0\Bigg(\frac{\hat{r}_{l}^{(k)}}{\hs_{\text{M}}\big(\hat{\br}^{(k)}\big)}\Bigg)+ \sum_{l \in C}\rho_0\Bigg(\frac{\hat{r}_{l}^{(k)}}{\hs_{\text{M}}\big(\hat{\br}^{(k)}\big)}\Bigg)=b\delta
\end{equation}

Recalling the proof of Theorem 4.1 in \cite{freueproteomic}, it follows

\begin{equation}
    \rho_0\big(\frac{1}{\gamma}\big)=\frac{b\delta}{m}
\end{equation}
where 
\begin{equation}
    \gamma=\lim_{k \rightarrow \infty}\frac{\hs_{\text{M}}\big( \hat{\br}^{(k)}\big)}{k^{\nu+1}}
\end{equation}

\paragraph{Component 2}

On the other hand, we have 
\begin{equation}
\begin{split}
  \frac{1}{b} \sum_{l=1}^b \rho_1 \Bigg( \frac{\hat{r}_{l}^{(k)}}{\hs_{\text{M}}\big(\hat{\br}^{(k)}\big)}\Bigg)=\frac{m}{b} \rho_1\Bigg(\frac{1-\bx_0^T \hbb^{(k)}/k^{\nu}}{\hs_{\text{M}}\big(\hat{\br}^{(k)}\big)/k^{\nu+1}}\Bigg) \Rightarrow\\
\lim_{k \rightarrow \infty}\frac{m}{b} \rho_1\Bigg(\frac{1-\bx_0^T \hbb^{(k)}/k^{\nu}}{\hs_{\text{M}}\big(\hat{\br}^{(k)}\big)/k^{\nu+1}}\Bigg)=\frac{m}{b} \rho_1\Big(\frac{1}{\gamma}\Big)  
\end{split}
\end{equation}

\paragraph{Component 3}

 $\lim_{k \rightarrow \infty} \|\hbb^{(k)}\|_{\ell_1}$ remains bounded according to the assumption of the proof.

\paragraph{Deriving $\tau$-Lasso Objective Function}

Now, we can evaluate the $\tau$-Lasso loss function according to equation (\ref{eq:TAU-Lasso}) as follows:
\begin{equation}
\begin{split}
      \lim_{k \rightarrow \infty} \frac{\hs_{\tau}^2\big(\hat{\br}^{(k)}\big)}{k^{2\nu+2}}= \lim_{k \rightarrow \infty}  \frac{\hs_{\text{M}}^2\big(\hat{\br}^{(k)}\big)}{k^{2\nu+2}}  \times \frac{m}{b} \rho_1\Bigg(\frac{1-\bx_0^T \hbb^{(k)}/k^{\nu}}{\hs_{\text{M}}^2\big(\hat{\br}^{(k)}\big)/k^{\nu+1}}\Bigg)\\
      \Rightarrow \lim_{k \rightarrow \infty} \frac{\hs_{\tau}^2\big(\hat{\br}^{(k)}\big)}{k^{2\nu+2}}=\frac{m\gamma^2}{b} \rho_1\Big(\frac{1}{\gamma}\Big) \Rightarrow 
       \lim_{k \rightarrow \infty} \frac{\mathcal{L}\big(\hbb^{(k)},\hat{\bY}_m^{(k)}\big)}{k^{2\nu+2}}\\=\lim_{k \rightarrow \infty} \Bigg( \frac{\hs_{\tau}^2\big(\hat{\br}^{(k)}\big)}{k^{2\nu+2}}+\lambda \frac{\|\hbb^{(k)}\|_{\ell_1}}{k^{2\nu+2}}\Bigg)=\frac{m\gamma^2}{b} \rho_1\Big(\frac{1}{\gamma}\Big)
\end{split}
\end{equation}
where $\lim_{k \rightarrow \infty} \|\hbb^{(k)}\|_{\ell_1}/k^{2\nu+2}=0$ due to the bounded norm of $\hbb^{(k)}$.
\subsubsection{Evaluation of \texorpdfstring{$\tau$}{tau}-Lasso Objective Function for \texorpdfstring{$\tilde{\boldsymbol{\beta}}^{(k)}$}{beta}}
Now, we evaluate the $\tau$-Lasso objective function for an unbounded sequence $\tilde{\bb}^{(k)}$ in a three-step procedure as follows:

\paragraph{Component 1}

The unbounded sequence is assumed to be  $\tilde{\bb}^{(k)}=\frac{k^{\nu}}{2}\bx_0$. The residuals for this sequence become 
\begin{equation}
    \hat{r}_{l}^{(k)}=\begin{cases}
  k^{\nu+1}-\frac{k^{\nu+1}}{2}\bx_0^T \bx_0  & l \in C\\    
  \cy_l-\frac{k^{\nu}}{2}\bx_0^T \cbx_{[l]} & l \notin C
\end{cases}= \begin{cases} \frac{k^{\nu+1}}{2}  & l \in C\\    
  \cy_l-\frac{k^{\nu}}{2} \bx_0^T \cbx_{[l]} & l \notin C
\end{cases}
\end{equation}
where $\bx_0^T \bx_0=\|\bx_0\|_{\ell_2}^2=1$ and all residuals go to infinity as $k \rightarrow \infty$. Hence, both  $\hs_{\tau}\big(\hat{\br}^{(k)}\big)$ and $\hs_{\text{M}}\big(\hat{\br}^{(k)}\big)$ tend to infinity. The decomposition of M-estimation of scale equation yields,

\begin{equation}
          \sum_{l \notin C}\rho_0\Bigg(\frac{ \cy_l-\frac{k^{\nu}}{2}\bx_0^T \cbx_{[l]}}{\hs_{\text{M}}\big(\hat{\br}^{(k)}\big)}\Bigg)+ \sum_{l \in C}\rho_0\Bigg(\frac{k^{\nu+1}/2}{\hs_{\text{M}}\big(\hat{\br}^{(k)}\big)}\Bigg)=b\delta 
\end{equation}

Using the proof of Theorem 4.1 in \cite{freueproteomic} , it can be inferred
\begin{equation}
\rho_0\Bigg(\frac{1}{\lim_{k \rightarrow \infty}\frac{\hs_{\text{M}}\big(\hat{\br}^{(k)}\big)}{k^{\nu+1}/2}}\Bigg)=\frac{b\delta}{m} 
\end{equation}
where
\begin{equation}
    \lim_{k \rightarrow \infty}\frac{\hs_{\text{M}}\big(\hat{\br}^{(k)}\big)}{k^{\nu+1}/2}=\gamma
\end{equation}

\paragraph{Component 2}

On the other hand, we have 
\begin{equation}
\begin{split}
  \frac{1}{b} \sum_{l=1}^b \rho_1 \Bigg( \frac{\hat{r}^{(k)}_l}{\hs_{\text{M}}\big(\hat{\br}^{(k)}\big)}\Bigg)=\frac{m}{b} \rho_1\Bigg(\frac{1}{\frac{\hs_{\text{M}}\big(\hat{\br}^{(k)}\big)}{k^{\nu+1}/2}}\Bigg) \Rightarrow\\
\lim_{k \rightarrow \infty}\frac{m}{b} \rho_1\Bigg(\frac{1}{\frac{\hs_{\text{M}}\big(\hat{\br}^{(k)}\big)}{k^{\nu+1}/2}}\Bigg)=\frac{m}{b} \rho_1\Big(\frac{1}{\gamma}\Big)  
\end{split}
\end{equation}

\paragraph{Component 3}

 $\lim_{k \rightarrow \infty} \|\tilde{\bb}^{(k)}\|_{\ell_1}$ diverges as $k$ goes to infinity.\\

\paragraph{Deriving $\tau$-Lasso objective function}

Now, we can evaluate the $\tau$-Lasso loss function according to equation (\ref{eq:TAU-Lasso}) as follows:

\begin{equation}
\begin{split}
      \lim_{k \rightarrow \infty} \frac{\hs_{\tau}^2\big(\hat{\br}^{(k)}\big)}{k^{2\nu+2}}= \lim_{k \rightarrow \infty}  \frac{\hs_{\text{M}}^2\big(\hat{\br}^{(k)}\big)}{k^{2\nu+2}} \times \frac{m}{b} \rho_1\Bigg(\frac{1}{\frac{\hs_{\text{M}}\big(\hat{\br}^{(k)}\big)}{k^{\nu+1}/2}}\Bigg)\\
      \Rightarrow \lim_{k \rightarrow \infty} \frac{\hs_{\tau}^2\big(\hat{\br}^{(k)}\big)}{k^{2\nu+2}}=\frac{m\gamma^2}{4b} \rho_1\Big(\frac{1}{\gamma}\Big) \Rightarrow 
       \lim_{k \rightarrow \infty} \frac{\mathcal{L}\big( \tilde{\bb}^{(k)},\hat{\bY}_m^{(k)}\big)}{k^{2\nu+2}}\\=\lim_{k \rightarrow \infty} \Bigg( \frac{\hs_{\tau}^2\big(\hat{\br}^{(k)}\big)}{k^{2\nu+2}}+\lambda \frac{\|\tilde{\bb}^{(k)}\|_{\ell_1}}{k^{2\nu+2}}\Bigg)=\frac{m\gamma^2}{4b} \rho_1(\frac{1}{\gamma})
\end{split}
\end{equation}

where
\begin{equation}
\begin{split}
   \lim_{k \rightarrow \infty}\|\tilde{\bb}^{(k)}\|_{\ell_1}/k^{2\nu+2}=\lim_{k \rightarrow \infty}\|k^{\nu}\bx_0/2\|_{\ell_1}/k^{2\nu+2}\\=\lim_{k \rightarrow \infty}\|\bx_0/2\|_{\ell_1}/k^{\nu+2}=0 
\end{split}
\end{equation}

\subsubsection{Comparison}
Now, we can compare the $\tau$-Lasso objective function for the given bounded and unbounded sequences and conclude that for large enough $k_0$,

\begin{equation}
    \frac{\mathcal{L}\big(\tilde{\bb}^{(k)},\hat{\bY}_m^{(k)}\big)}{k^{2\nu+2}}<\frac{\mathcal{L}\big(\hbb^{(k)},\hat{\bY}_m^{(k)}\big)}{k^{2\nu+2}}, \quad \forall k \geq k_0
\end{equation}

The above results contradict the fact that the bounded $\hbb^{(k)}$ is the minimum of the $\tau$-Lasso objective function for the contaminated sample with $m>b\delta$. Because the $\tau$-Lasso objective function for the unbounded $\tilde{\bb}^{(k)}$ is smaller than that of $\hbb^{(k)}$. This implies the $\hbb^{(k)}$ have to be unbounded and thus, the robust $\tau$-Lasso estimator breaks down for $m>b \delta$.\\

\section*{Proof of Theorem 2}
To begin with the proof, it follows from the first-order condition that $\tau$-estimates of regression parameter and scale for the given subset of data, $\hbb_b$ and $\hs_b$ must satisfy the following equations \cite{yohai1988high}:

\begin{equation} \label{eq:proof_cond}
\begin{split}
    \frac{1}{b}\sum_{l=1}^{b} \big[\hat{w}_{\tau}\rho_0^{'}\Big(\frac{\hr_l}{\hs_b}\Big)+\rho_1^{'}\Big(\frac{\hr_l}{\hs_b}\Big)\big] \tbx_{[l]}=\mathbf{0}, \\
    \frac{1}{b}\sum_{l=1}^{b} \rho_0\Big(\frac{\hr_l}{\hs_b}\Big) =\delta,
\end{split}
\end{equation}
where $\hr_l=\cy_l-\tbx_{[l]}^T \hbb_b$ and $\hat{w}_{\tau}$ is given by

\begin{equation} \label{eq:proof_weight}
   \hat{w}_{\tau}=
\frac{\sum_{l=1}^{b} \Big[ 2  \rho_1\Big(\frac{\hr_l}{\hs_b}\Big)- \rho_1^{'}\Big(\frac{\hr_l}{\hs_b}\Big) \frac{\hr_l}{\hs_b}\Big]}{\sum_{l=1}^{b} \rho_0^{'}\big(\frac{\hr_l}{\hs_b}\Big)\frac{\hr_l}{\hs_b}}. 
\end{equation}

Therefore, we can obtain $\tau$-estimates of regression $\hbb_b$ and scale $\hs_b$ for the subset of data with observations $\big(\cby,\tbX \big)$ as follows:

\begin{equation} \label{eq:proof_sol1}
\begin{split}
    \hbb_b=\bA_b(\hbb_b,\hs_b)^{-1}\bv_b (\hbb_b,\hs_b), \\
  \hs_b=\hs_b u_b(\hbb_b,\hs_b),
\end{split}
\end{equation}
where
\begin{equation} \label{eq:proof_sol2}
\begin{split}
    \bA_b(\hbb_b,\hs_b)= \frac{1}{b} \sum_{l=1}^{b} \hat{w}_l \tbx_{[l]} \tbx_{[l]}^{T}, \\
    \bv_b(\hbb_b,\hs_b)=\frac{1}{b} \sum_{l=1}^{b} \hat{w}_l \check{y}_l\tbx_{[l]}, \\
    u_b(\hbb_b,\hs_b)=\frac{1}{b \delta}\sum_{l=1}^{b} \rho_0\Big(\frac{\hr_l}{\hs_b}\Big),\\
    \hat{w}_l=\frac{\hat{w}_{\tau}\rho_0^{'}\Big(\frac{\hr_l}{\hs_b}\Big)+\rho_1^{'}\Big(\frac{\hr_l}{\hs_b}\Big)}{\hr_l}.
\end{split}
\end{equation}
Alternatively, the robust $\tau$-estimates of regression parameters and scale can be written as the solution of a fixed-point problem as follows:
\begin{equation}
    \hbth_b=\bbf (\hbth_b;\tbY),
\end{equation}
where $\bbf: \bR^{(|\hat{\mathcal{S}}|+1)} \rightarrow \bR^{(|\hat{\mathcal{S}}|+1)}$, $\hbth_b \in \bR^{(|\hat{\mathcal{S}}|+1)}$ and $\bbf(\hbth_b;\tbY)$ are given by

\begin{equation}
    \begin{split}
        \hbth_b=
        \begin{bmatrix}
         \hbb_b\\
         \mbox{ }\\
         \hs_b\\
        \end{bmatrix},
        \\
        \bbf(\hbth_b;\tbY)=
         \begin{bmatrix}
         \bA_b(\hbb_b,\hs_b)^{-1}\bv_b(\hbb_b,\hs_b) \\
         \mbox{ }\\
         \hs_b u_b(\hbb_b,\hs_b)\\
        \end{bmatrix}.
    \end{split}
\end{equation}
Conditioned on $\hat{\mathcal{S}}=\mathcal{S}$, $|\hat{\mathcal{S}}|$ can be replaced with $k_s$. Given $\rho_0$ and $\rho_1$ are differentiable functions, we can expand $\bbf$ by using Taylor expansion around the limiting values, $\bth_0=[\bb_0, \sigma_0]^T$, as follows:
\begin{equation}\label{eq:taylor_1}
\begin{split}
      \hbth_b
    =\bbf(\bth_0)+\nabla \bbf(\bth_0)(\hbth_b-\bth_0)\\
    +\underbrace{\frac{1}{2}\big[\mathbf{I}_{k_s} \otimes (\hbth_b-\bth_0)^T \big] \nabla^2\bbf(\bar{\bth})(\hbth_b-\bth_0)}_{R_b}  ,
\end{split}
\end{equation}
where $\bbf(\bth_0)$ is a short-hand for $\bbf(\bth_0; \tbY)$, $\nabla \bbf(\cdot) \in \bR^{(k_s+1) \times (k_s+1)}$ is the matrix of partial derivatives, $\nabla^2 \bbf(\cdot) \in \bR^{(k_s+1)^2 \times (k_s+1)}$ is the Hessian matrix of $\bbf(\cdot)$, $\otimes$ denotes the Kronecker product and $\bar{\bth}$ lies on the line segment between $\bth_0$ and $\hbth_b$. The term $\big[\mathbf{I}_{k_s} \otimes (\hbth_b-\bth_0)^T \big] \nabla^2\bbf(\bar{\bth})$ in the remainder is a $(k_s+1)^2 \times (k_s+1)$ matrix whose $(i,j)$-entry is given by

\begin{equation} \label{eq:fp_4}
\begin{split}
M_{ij}=(\hbth_b-\bth_0)^T \Bigg[\frac{\partial^2 \bbf(\bar{\bth})}{\partial \theta_{i} \partial \theta_{j}}\Bigg].\\
\end{split}
\end{equation}

In addition, $\nabla \bbf(\cdot)$ is defined as follows:
\begin{equation}
 \nabla \bbf(\bth)=\begin{bmatrix}
\partial [(\bA_b)^{-1} \bv_b]/\partial \bb
& \partial [(\bA_b)^{-1} \bv_b]/\partial \sigma\\
\partial[\sigma u_b]/\partial \bb & \partial[\sigma u_b]/\partial \sigma\\
\end{bmatrix}.
\end{equation}
Here, $u_b$, $\bA_b$ and $\bv_b$ are short-hands for $u_b(\bth)$, $\bA_b(\bth)$ and $\bv_b(\bth)$, respectively. Tedious but straightforward calculations show that the second-order terms $\partial^2 \bbf(\bar{\bth})/(\partial \theta_{i} \partial \theta_{j})$ are a combination of sample mean products. Taking into account the convergence of the sample mean products to their corresponding population mean according to Lemma 2 in \cite{salibian2002bootrapping}, (an extension of law of large numbers) and continuity of derivatives of $\rho_0$ and $\rho_1$, we can guarantee $\partial^2 \bbf(\bar{\bth})/(\partial \theta_{i} \partial \theta_{j})=O_p(1)$ for $i,j=1,\cdots,k_s+1$. On the other hand, it follows from the root-$n$ consistency of estimators \cite{shao2006mathematical} that $\|\hbth_b-\bth_0\|_{\ell_2}=O_p(1/\sqrt{b})$. Hence, the $(i,j)$-entry $M_{ij}=O_p(1/\sqrt{b})$. Noting that $i_{th}$ entry in the remainder is a linear combination $\sum_{j=1}^{k_s+1} M_{ij}( [\hat{\theta}_{b}]_{j}-[\theta_0]_{j})=o_p(1/\sqrt{b})$, implying the remainder term $R_b=o_p(1/\sqrt{b})$. Therefore, we can re-express the Taylor expansion given in equation (\ref{eq:taylor_1}) as follows:
\begin{equation} \label{eq:taylor_2}
\begin{split}
(\hbth_b-\bth_0)=(\bbf(\bth_0)-\bth_0)+\nabla \bbf (\bth_0)(\hbth_b-\bth_0) +o_p(1/\sqrt{b}) \Rightarrow
\\
[\mathbf{I}-\nabla \bbf(\bth_0)](\hbth_b-\bth_0)=(\bbf(\bth_0)-\bth_0)+o_p(1/\sqrt{b}) \Rightarrow
\\
  \sqrt{b}(\hbth_b-\bth_0)=[\mathbf{I}-\nabla \bbf(\bth_0)]^{-1} \sqrt{b}[\bbf(\bth_0)-\bth_0]
+o_p(1).  
\end{split}
\end{equation}
On the other hand, we know that $o_p(1)$ term converges to $\mathbf{0}_{k_s+1}$ in probability as $b$ tends to infinity. As a result, both sides of the following converge to the same limiting distribution.

\begin{equation} \label{eq:taylor_3}
\begin{split}
  \sqrt{b}(\hbth_b-\bth_0) \sim [\mathbf{I}-\nabla \bbf(\bth_0)]^{-1} \sqrt{b}[\bbf(\bth_0)-\bth_0]
\end{split},
\end{equation}
where the notation $\sim$ stands for weak convergence of both sides to the same limiting distribution. Consider $n$ bootstrap samples are drawn from the given subset of data, we approximate the actual bootstrap estimates $\hbth_{n,b}^{\star}$ for the given subset of data using linearly corrected one-step bootstrap estimates $\hbth_{n,b}^{R\star}$ as follows:

\begin{equation} \label{eq:BLFRB}
\begin{split}
  \sqrt{n}(\hbth_{n,b}^{R\star}-\hbth_b) \sim [\mathbf{I}-\nabla \bbf(\hbth_b)]^{-1} \sqrt{n}[\bbf^{\star}(\hbth_b)-\hbth_b]
\end{split},
\end{equation}
where $n$ denotes the number of observations in the complete data set and $\bbf^{\star}(\hbth_b)$ is one-step bootstrap estimate and shorthand for $\bbf^{\star}(\hbth_b; \tbY^{\star})$ where $\tbY^{\star}=\Big(\cby,\tbX;\bom^{\star}\Big)\in \bR^{n \times (k_s+1)}$. Since $[\mathbf{I}-\nabla \bbf(\hbth_b)]^{-1}$ is a consistent estimator of $[\mathbf{I}-\nabla \bbf(\bth_0)]^{-1}$, we only need to show that $\sqrt{n}[\bbf^{\star}(\hbth_b)-\hbth_b]$ converges to the same limiting distribution as $\sqrt{b}[\bbf(\bth_0)-\bth_0]$. To proceed with estimation of the correction matrix $[\mathbf{I}-\nabla \bbf(\hbth_b)]^{-1}$, we compute the gradient matrix $\mathbf{I}-\nabla \bbf(\hbth_b)$ as follows:

\begin{equation} \label{eq:inverse_mat}
\mathbf{I}-\nabla \bbf(\hbth_b)=
\begin{array}{cccccc}
\cline{1-6}
\multicolumn{1}{|c}{} & &  & &  \multicolumn{1}{c|}{} &  
\multicolumn{1}{c|}{} \\
\multicolumn{1}{|c}{}  & &  & &  \multicolumn{1}{c|}{}   &\multicolumn{1}{c|}{}  \\
\multicolumn{1}{|c}{}  & & \mathcal{A} &  &\multicolumn{1}{c|}{}   & \multicolumn{1}{c|}{\boldsymbol{\eta}}\\
\multicolumn{1}{|c}{}  & &  & &  \multicolumn{1}{c|}{} &   \multicolumn{1}{c|}{} \\
 \multicolumn{1}{|c}{} &  & & &  \multicolumn{1}{c|}{} & \multicolumn{1}{c|}{}\\
\cline{1-6}
\multicolumn{1}{|c}{} &  & \boldsymbol{\zeta}  & &\multicolumn{1}{c|}{} &    \multicolumn{1}{c|}{a}\\
\cline{1-6}
\end{array}.
\end{equation}
where
\begin{equation} \label{eq:matix_elements}
\begin{split}
\mathcal{A}=\mathbf{I}-\frac{\partial [(\bA_b)^{-1} \bv_b]}{\partial \bb} \Big|_{\hbth_b} \mbox{ ,    } \boldsymbol{\eta}=-\frac{\partial [(\bA_b)^{-1} \bv_b]}{\partial \sigma}\Big|_{\hbth_b}\\, a=1-\frac{\partial [\sigma u_b]}{\partial \sigma} \Big|_{\hbth_b}, 
\boldsymbol{\zeta}=-\frac{\partial [\sigma u_b]}{\partial \bb}\Big|_{\hbth_b}.
\end{split}
\end{equation}
Let's begin with calculating $\boldsymbol{\zeta}$,
\begin{equation} \label{eq:derivations_1}
\begin{split}
   \boldsymbol{\zeta}=-\frac{\partial [\sigma u_b]}{\partial \bb}  \Big|_{\hbth_b} \Rightarrow -\frac{\partial [\sigma u_b]}{\partial \bb}=-(\frac{\partial \sigma}{\partial \bb}u_b+\sigma \frac{\partial u_b}{\partial \bb})
  \Rightarrow \\\boldsymbol{\zeta}=\frac{1}{b \delta }\sum_{l=1}^b \rho_0^{'} (\bvr_l)\tbx_{[l]}^T.
\end{split}
\end{equation}
where $\check{r}_l$ denotes a shorthand for $\check{r}_l(\bb)=\cy_l-\tbx_{[l]}^T \bb$ and $\bvr_l=\hr_l/\hs_b$. In order to calculate $a$, we need to derive $\partial [\sigma u_b]/\partial \sigma$,

\begin{equation} \label{eq:derivations_2}
\begin{split}
   \frac{\partial [\sigma u_b]}{\partial \sigma}  =(\frac{\partial \sigma}{\partial \sigma}u_b+\sigma \frac{\partial u_b}{\partial \sigma})= 
    \frac{1}{b \delta}\Big[\sum_{l=1}^b \rho_0(\frac{\check{r}_l}{\sigma})-\sum_{l=1}^{b} \rho_0^{'}(\frac{\check{r}_l}{\sigma})\frac{\check{r}_l}{\sigma} \Big].
\end{split}
\end{equation}

Therefore, $a$ can be derived as follows:

\begin{equation}
    \begin{split}
        a=1-\frac{\partial [\sigma u_b]}{\partial \sigma}\Big|_{\hbth_b}=\frac{1}{b \delta}\sum_{l=1}^{b} \rho_0^{'} (\bvr_l)\bvr_l.
    \end{split}
\end{equation}

Finding $\mathcal{A}$ requires differentiating $ (\bA_b)^{-1} \bv_b$ with respect to $\bb$. To do so, $\boldsymbol{\alpha}_b=(\bA_b)^{-1} \bv_b$ is defined to simplify the derivations as follows:

\begin{equation} \label{eq:derivations_4}
\begin{split} 
\bv_b=\bA_b \boldsymbol{\alpha}_b  \Rightarrow \frac{\partial}{\partial \bb}[\bA_b\boldsymbol{\alpha}_b]=\frac{\partial \bv_b}{\partial \bb}.
\end{split}
\end{equation}
To avoid confusion, the subscripts are dropped from $\boldsymbol{\alpha}_b$, $\bA_b$ and $\bv_b$. Hence, we can express $\frac{\partial}{\partial \bb}[\bA \boldsymbol{\alpha}]$ as

\begin{equation} \label{eq:derivations_5}
\begin{split}
\frac{\partial}{\partial \bb}[\bA \boldsymbol{\alpha}]=\bA \frac{\partial \boldsymbol{\alpha}}{\partial \bb}  +\begin{bmatrix}
\big| & \vdots & \vdots & \big| \\
& & &
\\
\frac{\partial[\bA]}{\partial \beta_1}\boldsymbol{\alpha} & \vdots & \vdots &  \frac{\partial[\bA]}{\partial \beta_{k_s}} \boldsymbol{\alpha}\\
& & &  
\\
\big| & \vdots & \vdots & \big|
\end{bmatrix} \Rightarrow
\\
\frac{\partial \boldsymbol{\alpha}}{\partial \bb}=\bA^{-1}\Bigg[\frac{\partial \bv}{\partial \bb}- \begin{bmatrix}
\big| & \vdots & \vdots & \big| \\
& & &
\\
\frac{\partial[\bA]}{\partial \beta_1}\boldsymbol{\alpha} & \vdots & \vdots &  \frac{\partial[\bA]}{\partial \beta_{k_s}} \boldsymbol{\alpha}\\
& & &  
\\
\big| & \vdots & \vdots & \big|
\end{bmatrix} \Big].
\end{split}
\end{equation}
Next, we find the expression for $\frac{\partial \bv}{\partial \bb}$

\begin{equation} \label{eq:derivations_6}
\begin{split}
\frac{\partial \bv}{\partial \bb}=
    \frac{1}{b}\sum_{l=1}^{b} \frac{\check{r}_l \tbx_{[l]} \partial w_{\tau}/\partial \bb \rho_0^{'}(\check{r}_l/\sigma)}{\check{r}_l^2} \check{y}_l \\ +\frac{1}{b}\sum_{l=1}^{b}  
    \Big[ \frac{\check{r}_l\Big(-w_{\tau}\rho_0^{''}(\check{r}_l/\sigma) \tbx_{[l]}/\sigma - \rho_1^{''}(\check{r}_l/\sigma)\tbx_{[l]}/\sigma\Big) }{\check{r}_l^2} \\
 +\frac{\tbx_{[l]}\Big(w_{\tau}\rho_0^{'}(\check{r}_l/\sigma)+\rho_1^{'}(\check{r}_l/\sigma)\Big)}{\check{r}_l^2}\Big]
\check{y}_l \tbx_{[l]}^T, \end{split}
\end{equation}
where $w_{\tau}$ is given by

\begin{equation}
    w_{\tau}=\frac{\sum_{l=1}^{b} \Big[ 2  \rho_1\Big(\frac{\check{r}_l}{\sigma}\Big)- \rho_1^{'}\Big(\frac{\check{r}_l}{\sigma}\Big) \frac{\check{r}_l}{\sigma}\Big]}{\sum_{l=1}^{b} \rho_0^{'}\big(\frac{\check{r}_l}{\sigma}\Big)\frac{\check{r}_l}{\sigma}}. 
\end{equation}

 On the other hand, $\frac{\partial w_{\tau}}{\partial \bb}$ can be calculated as follows:

\begin{equation} \label{eq:derivations_8}
\begin{split}
\frac{\partial w_{\tau}}{\partial \bb}
=\frac{ \sum_{l=1}^{b}  \Big[\rho_1^{''}(\check{r}_l/\sigma)\tbx_{[l]}^T \check{r}_l/\sigma^2-\rho_1^{'}(\check{r}_l/\sigma)\tbx_{[l]}^T/\sigma\Big]}{\sum_{l=1}^b \rho_0^{'}(\check{r}_l/\sigma)\check{r}_l/\sigma} \\+ \frac{\sum_{l=1}^{b}\Big[\rho_0^{''}(\check{r}_l/\sigma)\tbx_{[l]}^T\check{r}_l/\sigma^2+\rho_0^{'}(\check{r}_l/\sigma)\tbx_{[l]}^T/\sigma \big] }{\sum_{l=1}^b \rho_0^{'}(\check{r}_l/\sigma)\check{r}_l/\sigma} \times w_{\tau}.
\end{split}
\end{equation}
Now, we need to compute $\partial [\bA] / \partial \beta_j$,

\begin{equation}\label{eq:derivations_12}
\begin{split}
\frac{\partial [\bA]}{\partial \beta_j}\boldsymbol{\alpha}
=\frac{1}{b} \sum_{l=1}^{b} \Big[\frac{\partial w_{\tau}/ \partial \beta_j \rho_0^{'}(\check{r}_l/\sigma)-w_{\tau}\rho_0^{''}(\check{r}_l/\sigma)\tilde{x}_{lj}/\sigma}{\check{r}_l} \\
-\frac{\rho_1^{''}(\check{r}_l/\sigma)\tilde{x}_{lj}/\sigma}{\check{r}_l}+\frac{\Big(w_{\tau} \rho_0^{'}(\check{r}_l/\sigma)+\rho_1^{'}(\check{r}_l/\sigma)\Big)\tilde{x}_{lj}}{\check{r}_l^2}
\Big]\tbx_{[l]} \tbx_{[l]}^{T}\boldsymbol{\alpha} \\
\end{split}
\end{equation}

Therefore, we have

\begin{equation}
\label{eq:derivations_17}
\begin{split}
\begin{bmatrix}
\big| & \vdots & \vdots & \big| \\
& & &
\\
\frac{\partial[\bA]}{\partial \beta_1}\alpha & \vdots & \vdots &  \frac{\partial[\bA]}{\partial \beta_{k_s}} \alpha\\
& & &  
\\
\big| & \vdots & \vdots & \big|
\end{bmatrix}
  =\frac{1}{b}\sum_{l=1}^{b} \frac{\check{r}_l \tbx_{[l]} \partial w_{\tau}/\partial \bb \rho_0^{'}(\check{r}_l/\sigma)}{\check{r}_l^2}\tbx_{[l]}^{T} \boldsymbol{\alpha}\\
  +\frac{1}{b}\sum_{l=1}^{b} \Big[ \frac{\check{r}_l\Big(-w_{\tau}\rho_0^{''}(\check{r}_l/\sigma) \tbx_{[l]}/\sigma - \rho_1^{''}(\check{r}_l/\sigma)\tbx_{[l]}/\sigma\Big) }{\check{r}_l^2} \\
 +\frac{\tbx_{[l]}\Big(w_{\tau}\rho_0^{'}(\check{r}_l/\sigma)+\rho_1^{'}(\check{r}_l/\sigma)\Big)}{\check{r}_l^2}\Big]
\tbx_{[l]}^{T} \tbx_{[l]}^{T} \boldsymbol{\alpha}
\end{split}
\end{equation}
Now, we can calculate $\frac{\partial \boldsymbol{\alpha}}{\partial \bb}$ as follows:

\begin{equation}\label{eq:derivations_18}
    \begin{split}
      \frac{\partial \boldsymbol{\alpha}}{\partial \bb}
=\bA^{-1}\Bigg[ \frac{1}{b}\sum_{l=1}^{b} \frac{\tbx_{[l]} \partial w_{\tau}/\partial \bb \rho_0^{'}(\check{r}_l/\sigma)}{\check{r}_l} (\check{y}_l-\tbx_{[l]}^{T} \boldsymbol{\alpha})
     \\ +\frac{1}{b}\sum_{l=1}^{b}  \Big[ \frac{-w_{\tau}\rho_0^{''}(\check{r}_l/\sigma) \tbx_{[l]}/\sigma - \rho_1^{''}(\check{r}_l/\sigma)\tbx_{[l]}/\sigma }{\check{r}_l} 
\\ +\frac{\tbx_{[l]}\Big(w_{\tau}\rho_0^{'}(\check{r}_l/\sigma)+\rho_1^{'}(\check{r}_l/\sigma)\Big)}{\check{r}_l^2}\Big]
 \tbx_{[l]}^{T} (\check{y}_l-\tbx_{[l]}^{T} \boldsymbol{\alpha})\Bigg]\\
\end{split}
\end{equation}
Plugging in $\hbth_b$ into $\frac{\partial \boldsymbol{\alpha}}{\partial \bb}$, we can proceed with $\mathcal{A}$,

\begin{equation}
    \begin{split}
        \mathcal{A}=\mathbf{I}-\frac{\partial [\bA^{-1} \bv]}{\partial \bb} \Big|_{\hbth_b}=\big(\hat{\bA}_b\big)^{-1}\Bigg[- \frac{1}{b}\sum_{l=1}^{b}  
     \tbx_{[l]} \nabla_{\bb} w_{\tau} \rho_0^{'}(\bvr_l)\\+\frac{1}{b}\sum_{l=1}^{b}\Big[ \hat{w}_{\tau}\rho_0^{''}(\bvr_l)  + \rho_1^{''}(\bvr_l) \Big]\tbx_{[l]}\tbx_{[l]}^{T}/\hs_b \Bigg]
    \end{split}
\end{equation}
 where  $\boldsymbol{\alpha}\Big|_{\hbth_b}=\hbb_b$ is given by the fixed-point assumption. Now, we turn our attention to deriving the last missing expression, $\frac{\partial [\bA^{-1}\bv]}{\partial \sigma}$,

\begin{equation}\label{eq:derivations_113}
    \begin{split}
        \frac{\partial \bv}{\partial \sigma}=\bA \frac{\partial \alpha}{\partial \sigma}+\frac{\partial \bA}{\partial \sigma}\alpha \Rightarrow \bA \frac{\partial \alpha }{\partial \sigma}=\frac{\partial \bv}{\partial \sigma}-\frac{\partial \bA}{\partial \sigma}\alpha \Rightarrow   \\
        \frac{\partial \alpha}{\partial \sigma}=\bA^{-1}(\frac{\partial \bv}{\partial \sigma}-\frac{\partial \bA}{\partial \sigma}\alpha)
    \end{split}
\end{equation}

Subsequently, $\frac{\partial \boldsymbol{\alpha}}{\partial \sigma}$ can be derived by following analogous procedures to $\frac{\partial \boldsymbol{\alpha}}{\partial \bb}$ as follows:

\begin{equation}
    \begin{split}
        \frac{\partial \boldsymbol{\alpha}}{\partial \sigma}
        =\bA^{-1}\Bigg[ \frac{1}{b}\sum_{l=1}^{b}  \Big[\frac{\partial w_{\tau}/\partial \sigma \rho_0^{'}(\check{r}_l/\sigma)-w_{\tau}\rho_0^{''}(\check{r}_l/\sigma)\check{r}_l/\sigma^2}{\check{r}_l}\\
        -\frac{\rho_1^{''}(\check{r}_l/\sigma)\check{r}_l/\sigma^2}{\check{r}_l}\Big] \tbx_{[l]} (\check{y}_l-\tbx_{[l]}^{T}\boldsymbol{\alpha})\Bigg]
    \end{split}
\end{equation}
where $\frac{\partial w_{\tau}}{\partial \sigma}$ is given as follows:
\begin{equation} \label{eq:derivations_22}
\begin{split}
\frac{\partial w_{\tau}}{\partial \sigma}
=\frac{\sum_{l=1}^{b}\Big[\rho_1^{''}(\check{r}_l/\sigma)\check{r}_l^2/\sigma^3-\rho_1^{'}(\check{r}_l/\sigma)\check{r}_l/\sigma^2\big]}{\sum_{l=1}^{b}\rho_0^{'}(\check{r}_l/\sigma)\check{r}_l/\sigma }
\\+ \frac{\sum_{l=1}^{b}\Big[\rho_0^{''}(\check{r}_l/\sigma)\check{r}_l^2/\sigma^3+\rho_0^{'}(\check{r}_l/\sigma)\check{r}_l/\sigma^2\Big]}{\sum_{l=1}^{b}\rho_0^{'}(\check{r}_l/\sigma)\check{r}_l/\sigma } \times w_{\tau}
\end{split}
\end{equation}
Having $\frac{\partial \boldsymbol{\alpha}}{\partial \sigma}$, we can compute $\boldsymbol{\eta}$ as follows:
\begin{equation}
    \begin{split}
        \boldsymbol{\eta}=-\frac{\partial[(\bA^{-1}\bv]}{\partial \sigma } \big|_{\hbth_b}
          =\big(\hat{\bA}_b\big)^{-1} \Bigg[ \frac{1}{b}\sum_{l=1}^{b}  \Big[-\nabla_{\sigma} w_{\tau} \rho_0^{'}(\bvr_l)\\
          +\hat{w}_{\tau}\rho_0^{''}(\bvr_l)\bvr_l/\hs_b+\rho_1^{''}(\bvr_l)\bvr_l/\hs_b \Big] 
        \tbx_{[l]}  \Bigg]
    \end{split}
\end{equation}

Consequently, we can exploit the block matrix inversion lemma to compute $[\mathbf{I}-\nabla \bbf(\hbth_{b})]^{-1}$ as follows:
\begin{equation}\label{eq:correction_proof}
\begin{split}
        \Big[ \mathbf{I}-\nabla \bbf(\hbth_{b}) \Big]^{-1}=\begin{bmatrix} 
\mathcal{A}& \boldsymbol{\eta}\\
\boldsymbol{\zeta} & a\end{bmatrix}^{-1}=\begin{bmatrix} 
\mathbf{M}_{b}& \mathbf{d}_{b}\\
\mathbf{N}_{b} & \mathbf{q}_{b} \end{bmatrix},
\\
 \mathbf{M}_{b}=\Big(\mathcal{A}-\boldsymbol{\eta}a^{-1}\boldsymbol{\zeta}\Big)^{-1},
      \\
      \mathbf{d}_{b}=-\mathcal{A}^{-1}\boldsymbol{\eta}\Big(a-\boldsymbol{\zeta}\mathcal{A}^{-1}\boldsymbol{\eta}\Big)^{-1},
      \\
      \mathbf{N}_{b}=-\Big(a-\boldsymbol{\zeta}\mathcal{A}^{-1}\boldsymbol{\eta}\Big)^{-1}\boldsymbol{\zeta}\mathcal{A}^{-1},\\
      \mathbf{q}_{b}=\Big(a-\boldsymbol{\zeta}\mathcal{A}^{-1}\boldsymbol{\eta}\Big)^{-1}.
\end{split}
\end{equation}

Now, we only need to prove both $\sqrt{n}[\bbf^{\star}(\hbth_b)-\hbth_b]$ and $\sqrt{b}[\bbf(\bth_0)-\bth_0]$ are convergent to the same limiting distribution. We show that  $\bbf(\bth_0)-\bth_0$ can be expressed as smooth function of means. Therefore, we can exploit the results on central limit theorem, and its extension to bootstrapping of smooth functions of means \cite{bickel1981some} and show that $\sqrt{n}[\bbf^{\star}(\hbth_b)-\hbth_b]$ and $\sqrt{b}[\bbf(\bth_0)-\bth_0]$ are convergent to the same limiting distribution. Let's define $\mathbf{Q}(\bth_0)$ and its expected value $\boldsymbol{\mu}(\bth_0)$ as follows:

\begin{equation}
    \begin{split}
      \mathbf{Q}(\bth_0)\\= \Big( \frac{\rho_0^{'}(r/\sigma_0)}{r}\bx \bx^T,\frac{\rho_1^{'}(r/\sigma_0)}{r}\bx \bx^T,
      \frac{\rho_0^{'}(r/\sigma_0)}{r} y \bx ,\frac{\rho_1^{'}(r/\sigma_0)}{r} y \bx,\\
 2 \rho_1(r/\sigma_0)-\rho_1^{'}(r/\sigma_0)r/\sigma_0 ,\rho_0^{'}(r/\sigma_0)r/\sigma_0, \sigma_0 \frac{\rho_0(r/\sigma_0)}{\delta }  \Big),\\
\boldsymbol{\mu}(\bth_0)=\bE[\mathbf{Q}(\bth_0)]=
\Big(\mathbf{Z}_1,\mathbf{Z}_2,\mathbf{z}_3,\mathbf{z}_4,z_5,z_6,\sigma_0 \Big) 
\end{split}
\end{equation}
where $\mathbf{Q}(\bth_0) \mbox{ and } \boldsymbol{\mu}(\bth_0) \in  \bR^{k_s \times k_s} \times \bR^{k_s \times k_s} \times \bR^{k_s} \times \bR^{k_s} \times \bR \times \bR \times \bR$ and

\begin{equation}
    \begin{split}
        r=y-\bx^T \bb_0 \\
        \bb_0=\big(\frac{z_5}{z_6}\times \mathbf{Z}_1+\mathbf{Z}_2\big)^{-1} \times \big(\frac{z_5}{z_6}\times \mathbf{z}_3+\mathbf{z}_4\big)
    \end{split}
\end{equation}
Given $b$ observations of $\Big(\check{y}_l,\tbx_{[l]}\Big)$, the sample mean $\bar{\mathbf{Q}}_b(\bth_0)$ is given by

\begin{equation}
    \begin{split}
      \bar{\mathbf{Q}}_b(\bth_0)\\
      = \Big( \frac{1}{b}\sum_{l=1}^b\frac{\rho_0^{'}(\check{r}_l(\bb_0)/\sigma_0)}{\check{r}_l(\bb_0)}\tbx_{[l]} \tbx_{[l]}^T,\frac{1}{b}\sum_{l=1}^b\frac{\rho_1^{'}(\check{r}_l(\bb_0)/\sigma_0)}{\check{r}_l(\bb_0)}\tbx_{[l]} \tbx_{[l]}^T\\,
      \frac{1}{b}\sum_{l=1}^b\frac{\rho_0^{'}(\check{r}_l(\bb_0)/\sigma_0)}{\check{r}_l(\bb_0)}\cy_l\tbx_{[l]}, \frac{1}{b}\sum_{l=1}^b\frac{\rho_1^{'}(\check{r}_l(\bb_0)/\sigma_0)}{\check{r}_l(\bb_0)}\cy_l\tbx_{[l]},\\
 \frac{1}{b}\sum_{l=1}^b \big[2 \rho_1(\Check{r}_l(\bb_0)/\sigma_0)-\rho_1^{'}(\check{r}_l(\bb_0)/\sigma_0)\check{r}_l(\bb_0)/\sigma_0 \big],\\
 \frac{1}{b}\sum_{l=1}^b \rho_0^{'}(\check{r}_l(\bb_0)/\sigma_0)\check{r}_l(\bb_0)/\sigma_0, \frac{\sigma_0}{b\delta } \sum_{l=1}^b \rho_0(\check{r}_l(\bb_0)/\sigma_0) \Big).
\end{split}
\end{equation}
where $\check{r}_l(\bb_0)=\cy_l-\cbx_{[l]}^T \bb_0$. Then, consider the function $\mathbf{g}:\bR^{k_s\times k_s} \times \bR^{k_s \times k_s} \times \bR^{k_s} \times \bR^{k_s} \times \bR \times \bR \times \bR \rightarrow \bR^{k_s} \times \bR$ defined as follows:
\begin{equation}
\begin{split}
      \mathbf{g}(\bar{\mathbf{Z}}_1,\bar{\mathbf{Z}}_2,\bar{\mathbf{z}}_3,\bar{\mathbf{z}}_4,\bar{z}_5,\bar{z}_6,\bar{z}_7)\\=\begin{bmatrix}
    \big(\frac{\bar{z}_5}{\bar{z}_6}\times \bar{\mathbf{Z}}_1+\bar{\mathbf{Z}}_2\big)^{-1} \times \big(\frac{\bar{z}_5}{\bar{z}_6}\times \bar{\mathbf{z}}_3+\bar{\mathbf{z}}_4\big),
\\
\bar{z}_7 \end{bmatrix}.  
\end{split}
\end{equation}
which is a composition of differentiable functions, yielding a smooth function. Now, we can express $\bbf(\bth_0)=\mathbf{g}(\bar{\mathbf{Q}}_b(\bth_0))$, $\bth_0=\mathbf{g}(\boldsymbol{\mu}(\bth_0))$, $\bbf(\hbth_b)=\mathbf{g}(\bar{\mathbf{Q}}_b(\hbth_b))$ , $\bbf^{\star}(\hbth_b)=\mathbf{g}^{\star}(\bar{\mathbf{Q}}_{n,b}(\hbth_b))$ and $\bbf(\hbth_b)=\hbth_b$ (based on the fixed-point property) as smooth function of means and consequently
\begin{equation}
    \sqrt{b}[\bbf(\bth_0)-\bth_0]=\sqrt{b}[\mathbf{g}(\bar{\mathbf{Q}}_b(\bth_0))-\mathbf{g}(\boldsymbol{\mu}(\bth_0))].
\end{equation}
Based on Theorem 2.2 in \cite{bickel1981some}, we have 

\begin{equation}
\begin{split}
       \sqrt{b}[\bar{\mathbf{Q}}_b(\bth_0)-\boldsymbol{\mu}(\bth_0)] \sim \sqrt{b}[\bar{\mathbf{Q}}_b^{\star}(\bth_0)-\bar{\mathbf{Q}}_b(\bth_0)],\\
       \sqrt{b}[\bar{\mathbf{Q}}_b^{\star}(\bth_0)-\bar{\mathbf{Q}}_b(\bth_0)] \sim
       \sqrt{n}[\bar{\mathbf{Q}}_{n,b}^{\star}(\bth_0)-\bar{\mathbf{Q}}_b(\bth_0)]
\end{split}
\end{equation}
and since the estimator $\hbth_b$ is consistent, we can show that
\begin{equation}
 \sqrt{n}[\bar{\mathbf{Q}}_{n,b}^{\star}(\bth_0)-\bar{\mathbf{Q}}_b(\bth_0)] \sim
 \sqrt{n}[\bar{\mathbf{Q}}_{n,b}^{\star}(\hbth_b)-\bar{\mathbf{Q}}_b(\hbth_b)].
\end{equation}
Given $\mathbf{g}$ is a smooth function, the following holds using Lemma 8.10 given in \cite{bickel1981some},

\begin{equation}
    \begin{split}
       \sqrt{b}[\mathbf{g}(\bar{\mathbf{Q}}_b(\bth_0))-\mathbf{g}(\boldsymbol{\mu}(\bth_0))]  \\\sim \nabla \mathbf{g}(\boldsymbol{\mu}(\bth_0)) \sqrt{b}[\bar{\mathbf{Q}}_b(\bth_0)-\boldsymbol{\mu}(\bth_0)],\\
       \sqrt{n}[\mathbf{g}(\bar{\mathbf{Q}}_{n,b}^{\star}(\hbth_b))-\mathbf{g}(\bar{\mathbf{Q}}_b(\hbth_b)) \\
       \sim \nabla \mathbf{g}(\boldsymbol{\mu}(\hbth_b))\sqrt{n}[\bar{\mathbf{Q}}_{n,b}^{\star}(\hbth_b)-\bar{\mathbf{Q}}_b(\hbth_b)].
    \end{split}
\end{equation}

Therefore, we have
\begin{equation}
\begin{split}
    \sqrt{b}[\mathbf{g}(\bar{\mathbf{Q}}_b(\bth_0))-\mathbf{g}(\boldsymbol{\mu}(\bth_0))]  \\
\sim
 \sqrt{n}[\mathbf{g}(\bar{\mathbf{Q}}_{n,b}^{\star}(\hbth_b))-\mathbf{g}(\bar{\mathbf{Q}}_b(\hbth_b))]
\end{split}
\end{equation}
which basically proves
\begin{equation}
 \sqrt{b}[\bbf(\bth_0)-\bth_0]   \sim \sqrt{n}[\bbf^{\star}(\hbth_b)-\hbth_b]
\end{equation}
As discussed earlier, $[\mathbf{I}-\nabla \bbf(\hbth_b)]^{-1}$ is a consistent estimate of $[\mathbf{I}-\nabla \bbf(\bth_0)]^{-1}$. Therefore, we have
\begin{equation}
\begin{split}
    \sqrt{b}(\hbth_b-\bth_0)  \sim \sqrt{n}(\hbth_{n,b}^{R\star}-\hbth_b) \\
  \sqrt{b}(\hbth_b-\bth_0)  \sim \sqrt{n}(\hbth_n-\bth_0)  \\
  \sqrt{n}(\hbth_n-\bth_0) \sim \sqrt{n}(\hbth_{n,b}^{R\star}-\hbth_b)
\end{split}
\end{equation}
where they will converge to the same limiting distribution as $n$ and $b$ tend to infinity.

\section*{Proof of Theorem 3}
 To begin with the proof, it is assumed a certain proportion of observations in $\bY$ are contaminated by outliers that no longer comply with the linear regression model given in equation (\ref{eq:linear_model}). Basically, we will show that if there are at least $p$ non-outlying observations in a bootstrap sample, it is guaranteed the FRB $\hbb_n^{R\star}$ remains bounded. Hence, we will determine under what conditions, the FRB $\hbb_n^{R\star}$ becomes unbounded or equivalently the maximum bias goes to infinity. The FRB $\hbb_n^{R\star}$ is given by
\begin{equation}
    \hbb_n^{R\star}=\mathbf{M}_n(\hbb_n^{1\star}-\hbb_n)+\mathbf{d}_n(\hs_n^{1\star}-\hs_n)
\end{equation}
where $\mathbf{M}_n$ and $\mathbf{d}_n$ can be computed by using equation (\ref{eq:RSOBT_0}) by changing $b$ to $n$ and $\big(\cby,\tbX \big)$ to $\big(\by,\bX \big)$. It is easy to show that correction factors $\mathbf{M}_n$ and $\mathbf{d}_n$ depend on the original data set $\bY$ rather than bootstrap sample and stay bounded if the original $\tau$-estimator $\hbb_n$ does not break down. Next, we need to discuss how $\hbb_n^{1\star}$ and $\hs_n^{1\star}$ are influenced by bootstrapping. Recall from equation (\ref{eq:BLFRB_2}) that

\begin{equation}
    \hs_n^{1\star}=\frac{\hs_n}{n\delta}\sum_{l=1}^{n}\rho_0\big(\frac{y_l^{\star }-\bx_{[l]}^{\star T}\hbb_n}{\hs_n}\big)
\end{equation}
which implies $\hs_n^{1\star}$ remains bounded for any bootstrap sample due to boundedness of $\rho_0(\cdot)$. Subsequently, we will study under what conditions one-step bootstrap $\tau$-estimates $\hbb_n^{1\star}$ can break down. $\hbb_n^{1\star}$ is given by

\begin{equation}
    \hbb_n^{1\star}=\Big[ \sum_{l=1}^{n}  \tilde{\bx}_{[l]}^{\star} \tilde{\bx}_{[l]}^{\star T}  \Big]^{-1}\Big[ \sum_{l=1}^{n}  \tilde{y}_{l}^{\star} \tilde{\bx}_{[l]}^{\star}  \Big],
\end{equation}
 where $\tilde{\bx}_{[l]}^{\star}=\sqrt{\hat{w}_l^{\star}}\bx_{[l]}^{\star}$, $\tilde{y}_{l}^{\star}=\sqrt{\hat{w}_l^{\star}} y_{l}^{\star}$ and $\hat{w}_l^{\star}$ is given by
 \begin{equation} \label{eq:weight_boot}
    \begin{split}
          \hat{w}_l^{\star}=\frac{\hat{w}_{\tau}^{\star}\rho_0^{'}\big(\frac{\hat{r}_l^{\star}}{\hs_n}\big)+\rho_1^{'}\big(\frac{\hat{r}_l^{\star}}{\hs_n}\big)}{\hat{r}_l^{\star}},\\
          \hat{w}_{\tau}^{\star}=\frac{\sum_{l=1}^{n}\big[ 2\rho_1\big(\frac{\hat{r}_l^{\star}}{\hs_n}\big)-\rho_1^{'}\big(\frac{\hat{r}_l^{\star}}{\hs_n}\big)\frac{\hat{r}_l^{\star}}{\hs_n}\big]}{ \sum_{l=1}^{n}\rho_0^{'}\big(\frac{\hat{r}_l^{\star}}{\hs_n}\big)\frac{\hat{r}_l^{\star}}{\hs_n}}
          ,\\
          \hat{r}_l^{\star}=y_l^{\star}-\bx_{[l]}^{\star T}\hbb_n.
    \end{split}
 \end{equation}
 Thus, $\hbb_n^{1\star}$ can be expressed as the solution of a least-square problem with the observations $\big(\tilde{\by}^{\star},\tilde{\bX}^{\star}\big)$. It can be inferred from the equation (\ref{eq:weight_boot}) that the weights $\hat{w}_l^{\star}$ will remain bounded as $ \hat{r}_l^{\star}$ approaches infinity. Therefore, one needs to verify that as long as there are at least $p$ good, non-outlying observations within the bootstrap sample, the corresponding one-step bootstrap estimate $\hbb_n^{1\star}$ will remain bounded. In other words, it would suffice to show that contamination by outliers will influence one-step bootstrap estimate $\hbb_n^{1\star}$ by a finite amount whose value is independent of outliers. 
 
 The remaining of the proof follows exactly that of Theorem 2 in \cite{salibian2002bootrapping} with an exception $c_1$ is assumed to be greater than or equal to $c_0$ or equivalently $ c_1=\max(c_1, c_0)$ without loss of generality.


\section*{Proof of Theorem 4}

According to Theorem 3, the qunatile estimates $\hat{q}_t^{\star}$ obtained by FRB employing $\tau$-estimator can breakdown under two scenarios as follows:
\begin{itemize}
    \item If $\hbb_b$ is an unreliable estimate of $\bb_0$, which may be attributed to the higher proportion of outliers than the finite-sample breakdown point of the estimator in $\cbY$.
    \item If the number of bootstrap samples containing less than $p$ good, non-outlying  observations constitutes at least $t\%$ of the total number of bootstrap samples, $B$.
\end{itemize}
Unreliable implies the estimate does not remain bounded any longer. In regard to RSOB-T replicates, all the bootstrap qunatiles $\hat{q}_t^{*}$, $t \in \big(0,1\big)$,  will be driven above any bound if $\hbb_b$ is already unreliable. By contrast, we can show all the bootstrap quantile estimates $\hat{q}_t^{*}$ will remain bounded under the given assumptions of the theorem with high probability approaching to one in large-scale datasets, $n \rightarrow \infty$, as long as $\hbb_b$ is reliable, i.e. the proportion of outliers in $\cbY$ is less than the finite-sample breakdown point of the estimator. This implies all bootstrap samples formed according to RSOB-T scheme will contain at least $p$ good observations.

Following a similar approach given in Theorem 1 and the relationship \ref{eq:tau_M_relation} between $\tau$-scale and M-scale, it is easy to show that finite-sample breakdown of $\tau$-estimator is the same as the finite-sample breakdown point of S-estimator. Under the given assumptions, the finite-sample breakdown point of S-estimator is given by P. Rousseuw in theorem 1 of \cite{rousseeuw1984robust}. Then, the finite-sample breakdown point of $\tau$-estimator is equal to that of S-estimator as follows:

\begin{equation}
    \epsilon^{*} (\hbb_{b},\cbY)=\frac{\lfloor b/2 \rfloor-|\hat{\mathcal{S}}|+1 }{b}
\end{equation}
where the initial estimate of $\bb_0$ with high breakdown is an S-estimate of $\bb_0$. Considering $\hbb_b$ is a bounded and reliable estimate of $\bb_0$, there exists at least $h=b-\big(\lfloor b/2 \rfloor-|\hat{\mathcal{S}}|+1 \big)$ non-outlying observations. Using lemma 1 in \cite{basiri2015robust}, all observations within $\cbY$ will be drawn at least once in the bootstrap sample with high probability converging to 1 as $n \rightarrow \infty$ (Big Data). Given the assumption of general position and the fact that $h>|\hat{\mathcal{S}}|$ implies there exists at least more than $|\hat{\mathcal{S}}|$ non-outlying observations in the bootstrap sample, we can conclude all bootstrap quantiles $\hat{q}_t^{\star}$ are bounded and reliable with high probability converging to 1 as $n \rightarrow \infty$ (Big Data).\\

%\begin{thebibliography}{1}


%\end{thebibliography}



%\newpage



%\vfill


\end{document}